\documentclass{article}
\pdfoutput=1

\usepackage{PRIMEarxiv}

\usepackage{url}
\usepackage{algorithm}
\usepackage{algorithmic}
\usepackage[utf8]{inputenc} 
\usepackage[T1]{fontenc}    
\usepackage{hyperref}       
\usepackage{url}            
\usepackage{booktabs}       
\usepackage{amsfonts}       
\usepackage{nicefrac}       
\usepackage{microtype}      
\usepackage{graphicx} 
\usepackage{wrapfig}
\usepackage{float} %
\usepackage{subfigure} %
\usepackage[subfigure]{tocloft}
\usepackage{enumitem}
\usepackage{amsthm,amsmath,amssymb}
\usepackage{wrapfig,lipsum,booktabs}
\usepackage{multirow}
\usepackage{algorithm, algorithmic}
\usepackage{bm}
\usepackage{hyperref}
\usepackage{wasysym}
\usepackage{pifont}
\usepackage{makecell}
\usepackage{adjustbox} 

\usepackage{appendix}
\usepackage{minitoc}
\usepackage{tocvsec2}
\usepackage{titletoc} 
\usepackage{tocloft}

\usepackage[table,xcdraw]{xcolor}
\definecolor{MyPurple}{HTML}{A020F0}

\newcommand{\xv}[0]{\ensuremath{\boldsymbol{x}} }
\newcommand{\yv}[0]{{\boldsymbol{y}} }

\newcommand{\zv}[0]{\ensuremath{\boldsymbol{z}} }

\newcommand{\Av}[0]{{\bf{A}} }
\newcommand{\Bv}[0]{{\bf{B}} }
\newcommand{\Cv}[0]{{\bf{C}} }

\newcommand{\Ev}[0]{{\bf{E}} }
\newcommand{\Fv}[0]{{\bf{F}} }

\newcommand{\Iv}[0]{{\bf{I}} }

\newcommand{\Mv}[0]{{\bf{M}} }

\newcommand{\Rv}[0]{{\bf{R}} }
\newcommand{\Sv}[0]{{\bf{S}} }

\newcommand{\Sigmav}[0]{\ensuremath{\boldsymbol{\Sigma}}}

\newcommand{\muv}[0]{\ensuremath{\boldsymbol{\mu}} }

\theoremstyle{plain}

\theoremstyle{definition}
\newtheorem{definition}{Definition}

\theoremstyle{remark}

\def\rr{\textcolor{black}}

\newcommand{\rebuttal}{\color{black}}

\usepackage[utf8]{inputenc} 
\usepackage[T1]{fontenc}    
\usepackage{hyperref}       
\usepackage{url}            
\usepackage{booktabs}       
\usepackage{amsfonts}       
\usepackage{nicefrac}       
\usepackage{microtype}      
\usepackage{xcolor}         

\title{Resultant: Incremental Effectiveness on Likelihood \\for Unsupervised Out-of-Distribution Detection}

\author{%
    \textbf{Yewen Li}$^{1}$\quad 
    \textbf{Chaojie Wang}$^{1,2}$\quad
    \textbf{Xiaobo Xia}$^3$ \quad
    \textbf{Xu He}$^4$ \quad 
    \textbf{Ruyi An}$^1$ \quad \\
    \textbf{Dong Li}$^4$ \quad 
    \textbf{Tongliang Liu}$^{3}$ \quad 
    \textbf{Bo An}$^{1,2}$ \quad 
    \textbf{Xinrun Wang}$^{1,5}$\quad 
    \\
    $^{1}$Nanyang Technological University \quad $^{2}$Skywork AI, Singapore \\ $^{3}$ The University of Sydney\quad $^{4}$Noah’s Ark Lab, Huawei  \quad $^{5}$ Singapore Management University
}

\begin{document}

\maketitle

\begin{abstract}
Unsupervised out-of-distribution (U-OOD) detection is to identify OOD data samples with a detector trained solely on unlabeled in-distribution (ID) data.
The likelihood function estimated by a deep generative model (DGM) could be a natural detector, but its performance is limited in some popular ``hard'' benchmarks, such as FashionMNIST (ID) \textit{vs.} MNIST (OOD). 
Recent studies have developed various detectors based on DGMs to move beyond likelihood. 
However, despite their success on ``hard'' benchmarks, {most of them} struggle to consistently surpass or match the performance of likelihood on some ``non-hard'' cases, such as SVHN (ID) \textit{vs.} CIFAR10 (OOD) where likelihood could be a nearly perfect detector.
Therefore, we appeal for more attention to \textit{incremental effectiveness on likelihood}, \textit{i.e.}, whether a method could always surpass or at least match the performance of likelihood in U-OOD detection.
We first investigate the likelihood of variational DGMs and find its detection performance could be improved in two directions: \textbf{i)} alleviating latent distribution mismatch, and \textbf{ii)} calibrating the dataset entropy-mutual integration.
Then, we apply two techniques for each direction, specifically \emph{post-hoc prior} and \emph{dataset entropy-mutual calibration}.
{The final method},
named {\textit{Resultant}}, combines these two directions for better incremental effectiveness compared to either technique alone. 
Experimental results demonstrate that the {\textit{Resultant}} could be a new state-of-the-art U-OOD detector while maintaining incremental effectiveness on likelihood in a wide range of tasks.
\end{abstract}

\section{Introduction}
The detection of out-of-distribution (OOD) data, which is to identify data that differ from the in-distribution (ID) training set, is crucial for ensuring the reliability and safety of real-world applications \cite{goodfellow2014explaining, hendrycks2016baseline, nguyen2015deep, wei@open}. 
While the most commonly used OOD detection methods rely on supervised classifiers \cite{alemi2018uncertainty, unlabel_help, super2, huang2022harnessing, Category-Extensible, super1,  VRA,  Augment_Distributions, conjnorm}, which require labeled data, 
this paper focuses on designing an unsupervised out-of-distribution (U-OOD) detector.
\textbf{U-OOD detection} refers to the task of designing a detector, based solely on unlabeled training data, that can determine whether an input is ID or OOD \cite{unsuper_add1,  CADet, hvae_ood, Geometric, informative-vae, likelihood-ratio-jie, input_complexity, xiao,  EBM_OOD}.
This unsupervised approach is practical in real-world scenarios where labeling a vast amount of training data is unnecessary or too effort-intensive, such as 
identifying aberrant or potentially harmful outputs from foundation models \cite{Helpful_and_Harmless, Secrets_of_RLHF}.

The deep generative model (DGM) is trained to maximize the likelihood estimation $p_\theta(\xv)$ with only the in-distribution training data. Thus, it is promising to use the likelihood $p_\theta(\xv)$ to perform U-OOD detection, where the in-distribution testing data should be assigned a higher likelihood while the OOD data should be given a lower one.
However, using likelihood to perform U-OOD detection is reported to have limited performance in some ``hard'' benchmarks \cite{hard_datasets}, such as detecting SVHN as OOD with a DGM trained on CIFAR10, denoted as CIFAR10 (ID) / SVHN (OOD). 
Therefore, existing works propose various new score functions as the U-OOD detector, \textit{e.g.}, the log-likelihood ratio method \cite{hvae_ood, informative-vae} utilizes the consistency between different layers' latent distribution in a hierarchical VAE, and these methods do improve the performance in commonly used hard benchmarks.
However, in some cases where using likelihood as the detector could already achieve perfect performance, \textit{e.g.}, the likelihood could achieve nearly 100\% performance under the metric AUROC in the SVHN (ID) / CIFAR10 (OOD) case, these new score functions have been found to be worse than using the likelihood \cite{hvae_ood} (see our Table \ref{tab:main_tab}).
It indicates that the likelihood of a DGM may already possess some capability to detect OOD data. Therefore, we believe diagnosing the likelihood and applying corresponding incremental improvements directly to it still holds potential for U-OOD detection.

In this work, we call for more attention to the \textit{incremental effectiveness on likelihood}, which assesses whether a new detector could generally surpass or at least match the performance of using likelihood for U-OOD detection and could be assessed by an advantage function as defined in Section \ref{sec:incremental_effectiveness}. 
With a clear definition, we analyze several major approaches for U-OOD detection in Section \ref{sec:incremental_effectiveness} and Appendix \ref{app:sec_revisit}, and find out that they have no explicit  guarantee of incremental effectiveness on likelihood, which 
potentially leads to their worse experimental performance than likelihood on some ``non-hard'' benchmarks.
It motivates us to take a deep look into the likelihood for developing a new U-OOD detector satisfying incremental effectiveness.
Specifically, we take the variational DGMs, including the variational autoencoder (VAE) \cite{kingma2013auto} and the diffusion model \cite{diffusion_ood_2, DDPM, unify}, for example to 
demonstrate a whole procedure of it and then discuss the extension to other DGMs in Section \ref{sec:discuss_extenstion_dgms}.
Overall, the contribution of this work could be summarized as follows:
\begin{enumerate}[leftmargin=0.8cm]
    \item We first revisit the literature to assess the incremental effectiveness on likelihood, then break down the likelihood to identify two directions for enhancing U-OOD detection: \textbf{i)} alleviating latent distribution mismatch, and \textbf{ii)} calibrating the dataset entropy-mutual integration.
    \item We propose a \textit{post-hoc prior} (PHP) method replacing the prior to a learnable distribution approximating ID data's aggregated posterior for {direction I} and a \textit{dataset entropy-mutual calibration} (DEC) method based on a compressor for {direction II}.
    \item Finally, we propose a new detector, named \textit{Resultant} $\mathcal{S}_{R}(\xv)$, combining the above two directions to achieve better incremental effectiveness on likelihood for U-OOD detection than both PHP and DEC. Experimental results, especially commonly used hard benchmarks and their reverse verification, support our analysis and demonstrate not only the incremental effectiveness but also the state-of-the-art performance of our method for U-OOD detection.
\end{enumerate}

\vspace{-2mm}
\section{Background: Unsupervised Out-of-distribution Detection}
\label{sec:background}

\textbf{Problem statement.} 
While deploying a machine learning system,
it is possible to encounter inputs from unknown distributions that are semantically {(\textit{e.g.}, category)} and/or statistically {(\textit{e.g.}, data complexity)} different from the training data, 
and such inputs are 
referred to as OOD data \cite{choi2018waic, input_complexity}. 
Thus, the OOD detection task is to identify these OOD data, which could be seen as a binary classification task: determining whether an input {$\xv$} is
more likely ID or OOD. {\rebuttal An ID-OOD classifier $D(\xv)$} could be formalized as a level-set estimation:
\begin{equation}
    \centering
    \begin{aligned}
        {\rebuttal D(\xv)} =
        \begin{cases}
         \text{ID}, & \text{if}\quad \mathcal{S}(\xv) > \lambda,\\
        \text{OOD}, & \text{if}\quad \mathcal{S}(\xv) \leq \lambda,
        \end{cases}
    \end{aligned}
\end{equation}
where $\mathcal{S}(\xv)$ denotes the score function, \textit{i.e.}, \textbf{OOD detector}, and the threshold $\lambda$ is commonly chosen to make a high fraction (\textit{e.g.}, 95\%) of ID data correctly classified \cite{wei2022logitnorm}. 

\textbf{Setup.} 
{Denoting the input space with $\mathcal{X}$, an \textit{unlabeled} training dataset $\mathcal{D}_{\text{train}}=\{\xv_i\}_{i=1}^{N}$ containing of $N$ data points can be obtained by sampling \textit{i.i.d.} from a data distribution $\mathcal{P}_{\mathcal{X}}$.}
We {treat} the $\mathcal{P}_{\mathcal{X}}$ as ${p}_{{id}}$. With this \textit{unlabeled} training set, \textbf{U-OOD detection} is to design a score function $\mathcal{S}(\xv)$ that can determine whether an input is ID or OOD. This is different from supervised OOD detection, which typically leverages a classifier trained on labeled data {and primarily focuses on semantic difference} \cite{wei2021odnl, wei@open,  wei2022logitnorm}.
We provide a detailed discussion in {Appendix \ref{sec:app_differnce}}.

{\textbf{Related Work.}} 
The existing literature on U-OOD detection can be divided into two main categories: \textbf{i)} Exploring why likelihood sometimes assigns higher values to OOD data compared to ID data, attributed to factors like image properties (\textit{e.g.}, intensity and contrast \cite{de-bias-vae}) and model characteristics (\textit{e.g.}, model's curvature \cite{flow_cannot}). However, the ultimate reason for this paradox is still a challenging, unresolved topic. Our work still \textbf{cannot} address this issue but we tried to take a further step with a concentration on the model likelihood that may potentially inspire the following works for understanding this paradox.
\textbf{ii)} Developing new scoring methods, \textit{e.g.}, Log-likelihood Ratio \cite{hvae_ood, informative-vae} leverages consistency across different layers' latent distribution of a hierarchical VAE; Likelihood Regret \cite{xiao} requires retraining for individual test samples; Likelihood Ratio \cite{likelihood-ratio-jie} incorporates a background model; WAIC \cite{choi2018waic} utilizes the statistic of an ensemble of DGMs' likelihood, and Geometric \cite{Geometric} applies a two-step condition judgment based on geometric insights of likelihood peaks. Our research primarily contributes to this category, emphasizing incremental effectiveness on likelihood. We provide a more comprehensive \textbf{Related Work} in { Appendix \ref{sec:appen_related_work}}.


\section{Incremental Effectiveness on Likelihood for U-OOD detection}
\subsection{{Revisiting Existing Literature from a New Perspective}} \label{sec:incremental_effectiveness}
Following the common analysis procedure \cite{flow_cannot, VRA}, we could assess the U-OOD detection performance of a score function $\mathcal{S}(\xv)$ by a {\textit{performance gap} $\mathcal{G}$}:
\begin{align}
    \mathcal{G}(\mathcal{S}) = \mathbb{E}_{\xv\sim p_{{id}}(\xv)} [\mathcal{S}(\xv)] - \mathbb{E}_{\xv\sim p_{{ood}}(\xv)} [\mathcal{S}(\xv)],
    \label{eq:assess_performance}
\end{align}
where $p_{{id}}(\xv)$ and $p_{{ood}}(\xv)$ denote the distribution corresponding to the ID and OOD datasets, respectively. A larger gap $\mathcal{G}>0$ can usually lead to better OOD detection performance. 

For the likelihood, we would use log-likelihood $\log p_\theta(\xv)$ instead of $p_\theta(\xv)$ as the U-OOD detector in the following part, as the $\log p_\theta(\xv)$ is the direct output of most DGMs and has the same detection performance as using $p_\theta(\xv)$, and the performance gap on likelihood could be expressed as 
\begin{align}
    \mathcal{G}(\log p_\theta) = \mathbb{E}_{\xv\sim p_{{id}}(\xv)} [\log p_\theta(\xv)] - \mathbb{E}_{\xv\sim p_{{ood}}(\xv)} [\log p_\theta(\xv)].
    \label{eq:assess_performance_likelihood}
\end{align}
Now we give the definition of \textit{incremental effectiveness on likelihood} with the performance gap:
\begin{definition}[\textbf{Incremental effectiveness on likelihood for U-OOD Detection}]
     \textit{Assume we have a score function $\mathcal{S}(\xv)$ based on a DGM parameterized by $\theta$ and different from $\log p_\theta(\xv)$, then we could use an advantage function $\mathcal{A}(\mathcal{S})$ to measure the incremental effectiveness on likelihood as 
    \begin{align}
        \mathcal{A}(\mathcal{S}) = \mathcal{G}(\mathcal{S}) - \mathcal{G}(\log p_\theta),
    \end{align}
where $\mathcal{A}(\mathcal{S})\geq 0$ in an U-OOD detection task means the score function $\mathcal{S}(\xv)$ could achieve incremental effectiveness on likelihood in this task.}
\end{definition}

For existing DGM-based U-OOD detection methods, we take the log-likelihood ratio method $\mathcal{LLR}^{>k}$ \cite{hvae_ood, informative-vae} for example, which selects a specific layer $k$ at first and {the score} can be formulated as 
\begin{align}
    \mathcal{LLR}^{>k} &=D_{\text{KL}}[p_\theta(\zv_{\leq k}|\zv_{>k})q_\phi(\zv_{>k}|\xv)||p_\theta(\zv|\xv)] 
    = \log p_\theta(\xv) - \log p_\theta(\xv, k) \notag \\
    &= \log p_\theta(\xv) - \mathbb{E}_{p_\theta(\zv_{\leq k}|\zv_{>k})q_\phi(\zv_{>k}|\xv)}[\log \frac{p_\theta(\xv|\zv)p_\theta(\zv_{>k})}{q_\phi(\zv_{> k}|\xv)}]. 
\end{align}
Thus, 
its incremental effectiveness on likelihood, {denoted as $\mathcal{A}$}, could be expressed as
\begin{align}
    \mathcal{A}(\mathcal{LLR}^{>k})&= - [\mathbb{E}_{\xv\sim p_{{id}}} \log p_\theta(\xv, k) - \mathbb{E}_{\xv\sim p_{{ood}}} \log p_\theta(\xv, k)].
    \label{eq:incremental_llr}
\end{align}
Though $\mathcal{LLR}^{>k}$ could be a good enough U-OOD detector by intuitively utilizing the consistency between different levels' latent distributions, it shows no incremental effectiveness on likelihood, \textit{i.e.}, $\mathcal{A}(\mathcal{LLR}^{>k})\geq0$ cannot be explicitly satisfied all the time. This could be empirically supported by their worse reverse verification experimental performance than likelihood in Table \ref{tab:main_tab}.
\textbf{We provide a detailed analysis for $\mathcal{LLR}^{>k}$ and more U-OOD detection methods in Appendix \ref{app:sec_revisit}.}

\subsection{Identifying Directions towards Incremental Effectiveness on Likelihood}
Previous works have analyzed the likelihood of variational DGMs in different perspectives, \textit{e.g.}, ELBO Surgery \cite{elbo_surgery} provides several rewritten expressions of the likelihood estimation for the VAEs \cite{kingma2013auto}. We adopt a common expression that estimates $\log p_\theta(\xv)$ by a reconstruction likelihood term minus a KL term, defined as
\begin{align}
   \log p_\theta(\xv) := \mathbb{E}_{\zv\sim q_\phi(\zv|\xv)}[\log p_\theta(\xv|\zv)] - D_{\text{KL}}(q_\phi(\zv|\xv)||p(\zv)).
    \label{eq:elbo}
\end{align}
For hierarchical variational DGMs such as hierarchical VAEs \cite{ladder} and diffusion models \cite{DDPM}, the $\log p_\theta(\xv)$ could utilize a hierarchy of latent variables $\zv=\{\zv_1,...,\zv_T\}$ and could be rewritten as
\begin{align}
    \log p_\theta(\xv) = \mathbb{E}_{\zv\sim q_\phi(\zv|\xv)}[\log p_\theta(\xv|\zv)] -\sum\nolimits_{t=1}^T D_{\text{KL}}[q_\phi(\zv_t|\zv_{t+1},\xv)||p_\theta(\zv_t|\zv_{t+1})],
    \label{eq:hierarchical}
\end{align}
where $p_\theta(\zv_T|\zv_{T+1}):=p(\zv_T)$ and $q_\phi(\zv_T|\zv_{T+1},\xv):=q_\phi(\zv_T|\xv)$. Note that \{$\xv, \zv_1,...,\zv_T$\} {can be} also written as \{$\xv_0, \xv_1,...,\xv_T$\} in diffusion models \cite{DDPM, unify}. With these {definitions}, we could take a further step beyond ELBO Surgery \cite{elbo_surgery} by expectations over the data distribution to analyze the incremental effectiveness on likelihood.

A representative work, Entropy Issue \cite{entropic_issue}, breaks down the expectation of the likelihood into two terms: a KL term indicating the distance between the estimated data distribution $p_\theta(\xv)$ and the true data distribution $p(\xv)$, as well as an entropy term of the data distribution, expressed as
\begin{align}
    \mathbb{E}_{\xv\sim p}[\log p_\theta(\xv)] = -D_{\text{KL}}[p(\xv)||p_\theta(\xv)] - \mathcal{H}_p(\xv),
    \label{eq:general_decompose}
\end{align}
and underscore the entropy's impact on the U-OOD detection. 
Clearly, besides entropy, an accurate estimated likelihood $p_\theta(\xv)$ can also benefit detection. Yet, even advanced diffusion models have shown limited performance in this regard \cite{diffusion_ood_2, diffusion_ood_1}. Therefore, we propose a deeper examination of the likelihood estimates from \textbf{trained DGMs}, particularly by further decomposing $D_{\text{KL}}[p(\xv)||p_\theta(\xv)]$, and suggesting to enhance incremental effectiveness through direct regularization of the likelihood.

We firstly reformulate the terms of $\mathbb{E}_{\xv\sim p(\xv)}[\log p_\theta(\xv)]$ from a perspective of information theory \cite{cover1999elements}:
{\small
\begin{align}
    \mathbb{E}_{\xv{\sim}{p(\xv)}}[\mathbb{E}_{\zv{\sim}{q_\phi(\zv|\xv)}}\log{p_\theta(\xv|\zv)}] &=\mathbb{E}_{p(\xv)q_\phi(\zv|\xv)}[\log\frac{p_\theta(\zv|\xv)}{p(\zv)}p(\xv)]
    ={\rebuttal {{\rebuttal \hat{\mathcal{I}}}}}_{q,p}(\xv,\zv)-\mathcal{H}_p(\xv), 
     \\
    \mathbb{E}_{\xv{\sim}{p(\xv)}}[D_{\text{KL}}(q_{\phi}(\zv|\xv)||p(\zv))] =\mathbb{E}&_{p(\xv)q_{\phi}(\zv|\xv)}[\log\frac{q_\phi(\zv|\xv)}{q(\zv)}\frac{q(\zv)}{p(\zv)}]
    ={\rebuttal \hat{\mathcal{I}}}_q{(\xv,\zv)} + D_{\text{KL}}(q(\zv)||p(\zv)),
\end{align}
}

\vspace{-2mm}
where $q(\zv)=\mathbb{E}_{\xv\sim p(\xv)}q_\phi(\zv|\xv)$ denotes the aggregated posterior distribution of the latent variables $
\zv$,
and ${\rebuttal \hat{\mathcal{I}}}_{q,p}(\xv,\zv)$ and ${\rebuttal \hat{\mathcal{I}}}_{q}(\xv,\zv)$ 
are defined as
{\small
\begin{align}
    \hat{\mathcal{I}}_{q,p}(\xv,\zv) =& -\hat{\mathcal{H}}_{q,p}(\zv|\xv) + \hat{\mathcal{H}}_{q,p}(\zv) 
    = \mathbb{E}_{p(\xv)q_\phi(\zv|\xv)}[\log p_\theta(\zv|\xv)] - \mathbb{E}_{q(\zv)}[\log p(\zv)], 
    \label{eq:first_equation} \\
    {\rebuttal \hat{\mathcal{I}}}_{q}(\xv,\zv) &= -{\rebuttal \hat{\mathcal{H}}}_{q}(\zv|\xv) + {\rebuttal \hat{\mathcal{H}}}_{q}(\zv) 
    = \mathbb{E}_{p(\xv)q_\phi(\zv|\xv)}[\log q_\phi(\zv|\xv)] - \mathbb{E}_{q(\zv)}[\log q(\zv)].
    \label{eq:second_equation} 
\end{align}
}

\vspace{-2mm}
Therefore, the expectation of the $\log p_\theta(\xv)$ on the data distribution $p(\xv)$ could be rewritten as
{\small
\begin{align}\label{eq:expectation_logp}
        &\mathbb{E}_{\xv\sim p(\xv)} [\log p_\theta(\xv)]  = {\mathbb{E}_{\xv\sim p(\xv)} [\mathbb{E}_{\zv\sim q_\phi(\zv|\xv)}\log p_\theta(\xv|\zv)]}   - \mathbb{E}_{\xv\sim p(\xv)} [D_{\text{KL}}(q_\phi(\zv|\xv)||p(\zv))] \\ 
    =& - D_{\text{KL}}(q(\zv)||p(\zv)) - [\mathcal{H}_p(\xv)+  \hat{\mathcal{I}}_{q}(\xv,\zv) - \hat{\mathcal{I}}_{q,p}(\xv,\zv)] =  - D_{\text{KL}}(q(\zv)||p(\zv)) - \text{Ent-Mut}(\theta, \phi, p),\notag
\end{align}
}

\vspace{-2mm}
{where} we term the integration of dataset entropy and mutual information gap as \emph{dataset entropy-mutual integration}, denoted as $\text{Ent-Mut}(\theta, \phi, p)$,
which is a constant only related to data distribution and model parameters {when the parameters of DGMs are fixed.}
Thus, the gap $\mathcal{G}$ of $\log p_\theta(\xv)$ becomes 
{\small
\begin{align}
    \mathcal{G}(\log p_\theta) = [-D_{\text{KL}}(q_{{id}}(\zv)||p(\zv)) + D_{\text{KL}}(q_{{ood}}(\zv)||p(\zv))]  
    - [\text{Ent-Mut}(\theta, \phi, p_{id}) - \text{Ent-Mut}(\theta, \phi, p_{ood})].  \label{eq:root} 
\end{align}
}

Observing the above performance gap $\mathcal{G}(\log p_\theta)$, we find two directions to increase its magnitude:

\textbf{Direction I: Alleviating latent distribution mismatch.}
Several studies have argued that the aggregated posterior distribution of latent variables $q(\zv)$ cannot strictly equal the prior distribution $p(\zv)$ parameterized by a multivariate Gaussian $\mathcal{N}(\mathbf{0},\Iv)$ \cite{pz_not_proper2,  distribution_match, pz_not_proper1, trilema}. 
Besides, the non-informative prior $p(\zv)$ could overly cover the OOD data's latent distribution $q_{{ood}}(\zv)$ and lead to a small $D_{\text{KL}}(q_{{ood}}(\zv)||p(\zv))$ \cite{diffusion_ood_2}.
We also provide a justification for these points theoretically and experimentally in {Appendix \ref{app:sec_justify_latent}}.
Thus, the value of $[-D_{\text{KL}}(q_{{id}}(\zv)||p(\zv)) + D_{\text{KL}}(q_{{ood}}(\zv)||p(\zv))]$ could be further maximized by replacing the prior $p(\zv)$ in the KL term of Eq. \ref{eq:elbo} with a learnable distribution $\hat{q}_{{id}}(\zv)$ approximating the in-distribution data aggregated posterior distribution $q_{{id}}(\zv)$.

\textbf{Direction II: Calibrating the dataset entropy-mutual integration $\text{Ent-Mut}(\theta, \phi, p)$}.
Considering the dataset's statistics, such as the variance of pixel values, different datasets exhibit 
various levels of entropy. As an example, the FashionMNIST dataset should possess higher entropy compared to the MNIST dataset, { as it has a higher degree of variations}. Therefore, when the entropy of an observed ID dataset is too high, the value of $-\mathcal{H}_{p_{{id}}}(\xv) + \mathcal{H}_{p_{{ood}}}(\xv)$ could be small in magnitude.
The mutual information term ${\rebuttal \hat{\mathcal{I}}}_{q,p}(\xv, \zv) - {\rebuttal \hat{\mathcal{I}}}_{q}(\xv, \zv)$ {is derived from the insufficient optimization or model capacity} of model parameters $\theta$ and $\phi$ on characterizing the data distribution $p$ as the trained DGMs in practice are always not optimal \cite{bindai_2, bindai_1}. 
To avoid these terms' impact on hindering the U-OOD detection performance, we would develop an approximation method for this Ent-Mut ($\theta, \phi, p$) term to calibrate its impact in the following part.

Finally, we proposed a new detector, \textit{Resultant} $\mathcal{S}_R(\xv)$, to combine these two directions for better incremental effectiveness on likelihood.

\section{Method}

\begin{figure}[ht!]
    \centering
    \subfigure[Insight for PHP]{\includegraphics[width=0.45\textwidth]{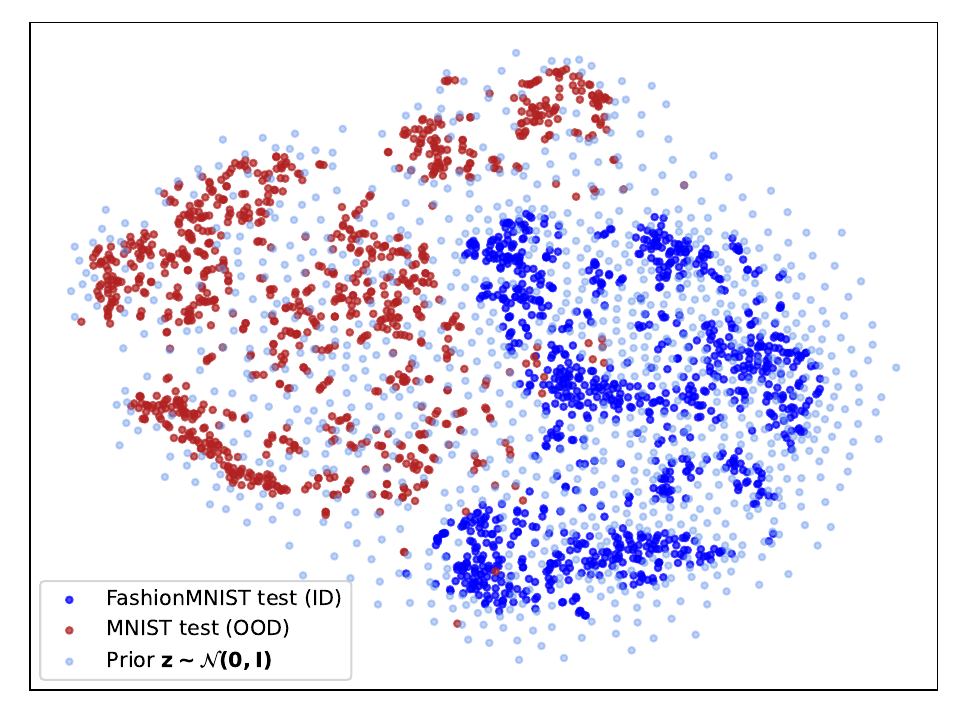}
    \label{fig:tsne}
    }
    \subfigure[Insight for DEC]{\includegraphics[width=0.45\textwidth]{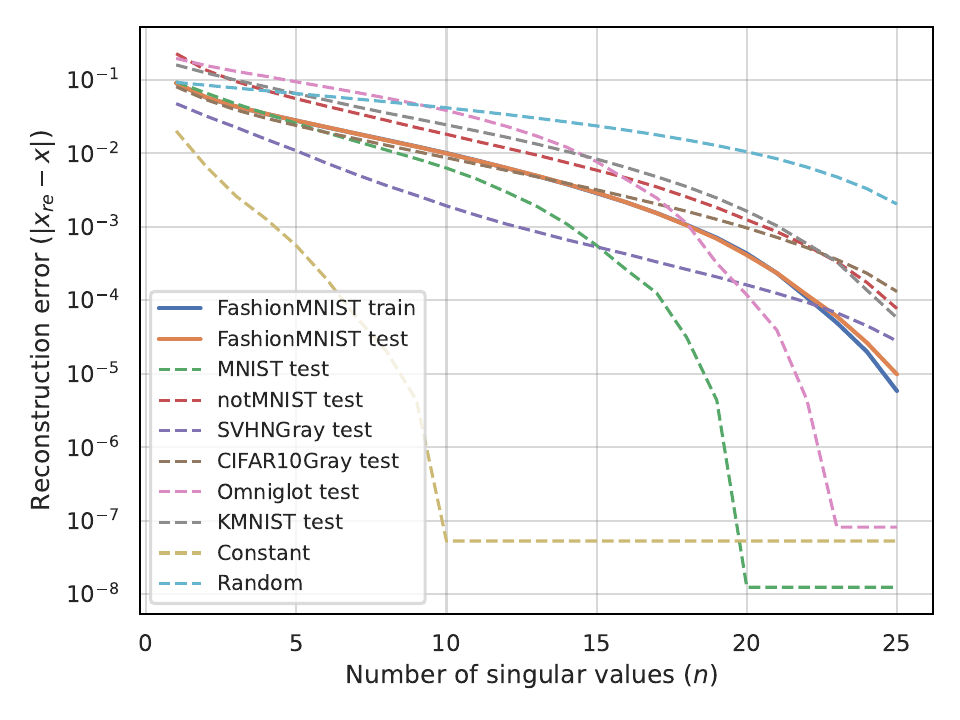}
    \label{fig:svd-fashionmnist}
    }
    \caption{Illustration of the insights of the proposed methods. \textbf{(a)}: The t-SNE visualization of latent representations on FashionMNIST(ID)/MNIST(OOD) dataset pair. \textbf{(b)} Visualization of the relationship between the number of singular values and the reconstruction error.}
\end{figure}

\subsection{Post-hoc Prior Method for Direction I}


To provide a more insightful view to investigate
the relationship between $q_{{id}}(\zv)$, $q_{{ood}}(\zv)$, and $p(\zv)$, we use t-SNE \cite{tsne} to visualize them in Fig.~\ref{fig:tsne}. We also provide an extra UMAP \cite{umap} visualization demonstrating a similar property for enhancing the result's reliability and their settings are in Appendix \ref{sec_re_umap}.
{As the {dark}-blue points (latent representations of FashionMNIST) are much more distinguishable from the red points (MNIST) than the light-blue points (latent $\zv$ sampled from $\mathcal{N}(\textbf{0},\Iv)$) from the red points, this indicates} that $p(\zv)$ cannot distinguish between the latent variables sampled from $q_{{id}}(\zv)$ and $q_{{ood}}(\zv)$, while $q_{{id}}(\zv)$ is clearly distinguishable from $q_{{ood}}(\zv)$.
A quantitative justification is also provided in {Fig. \ref{fig:multi_case} of Appendix \ref{app:sec_justify_latent}.}
This phenomenon could be explained from a perspective of manifold fitting, where an observed dataset actually lies on a unique low-dimensional data-generating manifold, \textit{i.e.}, the low-dimensional latent distribution $q(\zv)$, embedded in
high-dimensional ambient space \cite{manifold_overfit}.
Thus, ID and OOD data should be very different in this low-dimensional manifold if they share less data-generating manifold support like semantic similarities, leading to an extremely large $D_{\text{KL}}(q_{{ood}}(\zv)||q_{{id}}(\zv))$. But $p(\zv)$ could cause overfitting on them, making $D_{\text{KL}}(q_{{id}}(\zv)||p(\zv))$ larger and $D_{\text{KL}}(q_{{ood}}(\zv)||p(\zv))$ smaller than they should be.

Therefore, to gain incremental effectiveness, we can explicitly modify the prior distribution $p(\zv)$ in $\mathcal{G}(\log p_\theta)$ to force it to be closer to $q_{{id}}(\zv)$ and far from $q_{{ood}}(\zv)$,
\textit{i.e.}, pursuing $D_{\text{KL}}(q_{{id}}(\zv)||p(\zv))>D_{\text{KL}}(q_{{id}}(\zv)||q_{{id}}(\zv))$ and {$D_{\text{KL}}(q_{{ood}}(\zv)||p(\zv)) < D_{\text{KL}}(q_{{ood}}(\zv)||q_{{id}}(\zv))$}.
{This could be achieved by a two-step scheme by learning the $q_{id}(\zv)$ after training \cite{bindai_1, manifold_overfit}.}
Specifically, we could replace $p(\zv)$ in the KL-term of likelihood $\log p_\theta$ in Eq. \ref{eq:elbo} with a learnable distribution $\hat{q}_{{id}}(\zv)$ that can fit $q_{{id}}(\zv)$ well \textbf{after training} the DGMs, where the target value of $q_{{id}}(\zv)$ can be acquired by marginalizing 
$q_\phi(\zv|\xv)$ over the training set, \textit{i.e.}, $q_{{id}}(\zv)=\mathbb{E}_{\xv\sim p_{{id}}(\xv)}[q_\phi(\zv|\xv)]$.
Latent distribution matching have been analyzed thoroughly in previous studies\cite{bindai_2, bindai_1, distribution_match} but they have {\textbf{not}} been investigated for U-OOD detection, and we directly adopt an LSTM-based method proved efficiently to fit a latent distribution \cite{distribution_match}, \textit{i.e.},
\begin{equation}
\begin{small}
\begin{aligned}
\hat{q}_{{id}}(\zv) = \prod\nolimits_{t=1}^T q(\zv_t|\zv_{<t}), \text{ where } q(\zv_t|\zv_{<t}) = \mathcal{N}(\mu_i, \sigma_i^2).
\end{aligned}
\end{small}
\end{equation}
Thus, we could propose the \rr{\emph{post-hoc prior}} (PHP) method for Direction I, expressed as
\begin{equation}
\begin{small}
\begin{aligned}
    {\text{PHP}}(\xv) = \mathbb{E}_{\zv\sim q_\phi(\zv|\xv)}\log p_\theta(\xv|\zv)  - D_{\text{KL}}(q_\phi(\zv|\xv) || \hat{q}_{{id}}(\zv)),
    \label{eq:php}
\end{aligned}
\end{small}
\end{equation}
and its expectation on a data distribution is
\begin{equation}
\begin{small}
    \mathbb{E}_{\xv\sim p(\xv)}[\text{PHP}(\xv)] = - D_{\text{KL}}[q(\zv)||\hat{q}_{id}(\zv)] -\text{Ent-Mut}(\theta, \phi, p) ,
    \label{eq:Expectation_PHP}
\end{small}
\end{equation}

which could lead to better U-OOD detection performance since it could achieve incremental effectiveness on likelihood, \textit{i.e.},
{\small
\begin{align}
    &\mathcal{A}(\text{PHP}) = \mathcal{G}(\text{PHP}) - \mathcal{G}(\log p_\theta)  \\
    =& [-D_{\text{KL}}(q_{{id}}(\zv)||\hat{q}_{{id}}(\zv))+D_{\text{KL}}(q_{{ood}}(\zv)||\hat{q}_{{id}}(\zv))]  - [-D_{\text{KL}}(q_{{id}}(\zv)||p(\zv))+D_{\text{KL}}(q_{{ood}}(\zv)||p(\zv))] {\geq}0. \notag
\end{align}
}


\subsection{Dataset {Entropy-mutual} Calibration Method for Direction II}
\label{sec:dec}


Recall dataset entropy-mutual integration that includes entropy $\mathcal{H}_p(\xv)$ and mutual information gap $\hat{\mathcal{I}}_{q}(\xv,\zv) - \hat{\mathcal{I}}_{q,p}(\xv,\zv)$.
The former, known to significantly degenerate U-OOD detection performance \cite{entropic_issue}, can be empirically and roughly estimated using compressor-based methods \cite{input_complexity}. The latter, stemming from suboptimal model performance, is intractable but {expected to be minor as both terms are of similar scale.} This section focuses on proposing an approximation method for dataset entropy at first and then mitigating the impact of the mutual information gap in $\mathcal{G}$ through a reweighting scale.


We begin by assuming a dataset with finite data points is sampled from a unique underlying distribution $p$, which allows for infinite sampling of new examples sharing similar semantic and statistical properties. Simple datasets like MNIST, featuring only white digit foregrounds on black backgrounds, likely have lower dataset entropy than complex datasets like CIFAR10, which contain rich pictorial elements.  Data complexity can be empirically estimated using a compressor (e.g., a JPEG compressor) by the length of compressed bits \cite{input_complexity}, denoted as $\mathcal{C}(\xv)$. Therefore, if the $\mathcal{C}(\xv)$ is adjusted to match the scale of $\mathcal{H}_p(\xv)$, \textit{i.e.}, $\mathbb{E}_{p}[\mathcal{C}(\xv)]\cdot \text{scale} \approx \mathcal{H}_p(\xv)$, the dataset entropy's impact can be mitigated by adding it to the $\log p_\theta(\xv)$. 

To showcase the generalization ability of our proposed method using different compressor schemes, we utilized an SVD method to simulate compressors with varying degrees of compression loss. As illustrated in Fig.~\ref{fig:svd-fashionmnist}, an SVD compressor with a higher number of singular values, $n$, correlating with lower reconstruction errors, indicates a more effective compressor. For the $\mathcal{C}(\xv)$ computation, we start by selecting a representative compression ability, $n_{id}$, and its corresponding error, $\epsilon$, on the ID dataset. Notably, $\epsilon$ remains unchanged when $n_{id}$ exceeds a certain threshold $n_{id}^{max}$. $n_{id}$ is set to half this $n_{id}^{max}$ in our main experiments and other $n_{id}$ values are evaluated in an ablation study. For a test input $\xv$, SVD identifies the smallest $n_i$ within \([1, n_{id}^{max}]\) that achieves a lower error than $\epsilon$. A binary search could speed up this process and we set $n_i$ to $n_{id}^{max}$ if no smaller error is achievable. The complexity measure $ \mathcal{C}(\xv)$ is then defined as:
\begin{equation}
\begin{small}
\begin{aligned}
    \mathcal{C}(\xv) = {n_i}/n_{id}.
\end{aligned}   
\end{small}
\end{equation}
This is inspired and similar to the Input Complexity \cite{input_complexity}, yet removing the entropy term might {\textbf{not}} enhance effectiveness because OOD data with higher complexity than $n_{id}$ will get a larger $\mathcal{C}(\xv)$. Additionally, the compressor-based method may inaccurately estimate entropy, and the impact of the mutual information gap should also be considered.
Therefore, we firstly propose to further enhance $\mathcal{C}(\xv)$ with a following property:
\begin{equation}
\begin{small}
\begin{aligned}
    \mathbb{E}_{\xv\sim p_{{id}}(\xv)} [\mathcal{C}(\xv)]  {\geq}  \mathbb{E}_{\xv\sim p_{{ood}}(\xv)} [\mathcal{C}(\xv)],
    \label{eq:property_of_cx_1}
\end{aligned}   
\end{small}
\end{equation}
which could be simply satisfied by penalizing the test data sample that owns a larger complexity $n_i$ than the average complexity $n_{id}$ of the ID dataset, \textit{i.e.},
$n_i >n_{id}$, by assigning it lower $\mathcal{C}(\xv)<1$: 
\begin{equation}
\begin{small}
\begin{aligned}
    \mathcal{C}(\xv) =
    \begin{cases}
        n_i/n_{{id}}, & \text{if} \quad n_i < n_{{id}}, \\
        (n_{{id}} - (n_i - n_{{id}}))/{n_{{id}}}, & \text{if} \quad n_i \geq n_{{id}}.
        \label{eq:dec_property}
    \end{cases}
\end{aligned}   
\end{small}
\end{equation}
Note that alternative methods like the JPEG compressor can also implement $\mathcal{C}(\xv)$. Implementation details and comparisons of performance and computational efficiency are provided in {Appendix \ref{sec_re_compressor}}.

Secondly, we propose rescaling for 
$\mathcal{C}(\xv)$ by considering both dataset entropy and mutual information gap.  It could be rescaled by exploiting the PHP method, \textit{i.e.}, when $\hat{q}_{{id}}(\xv)\approx q_{id}(\xv)$, we have
\begin{equation}
\begin{small}
\begin{aligned}
    \mathbb{E}_{p_{{id}}}[\text{PHP}(\xv)]=-D_{\text{KL}}[q_{{id}}(\zv)||\hat{q}_{{id}}(\zv)] - \text{Ent-Mut}(\theta, \phi, p_{{id}}) \approx - \text{Ent-Mut}(\theta, \phi, p_{{id}}),
    \label{eq:max_direction1}
\end{aligned}   
\end{small}
\end{equation}
and we could define 
\begin{equation}
\begin{small}
\begin{aligned}
    \text{scale} = \mathbb{E}_{p_{{id}}}[-\text{PHP}(\xv)] / \mathbb{E}_{p_{{id}}}[\mathcal{C}(\xv)].
\end{aligned}   
\end{small}
\end{equation}
Therefore, the DEC method could be finally designed as
\begin{equation}
\begin{small}
\begin{aligned}
    \text{DEC}(\xv) = \log p_\theta(\xv) + \mathcal{C}(\xv)\cdot \text{scale}  = \log p_\theta(\xv) + \mathcal{C}(\xv)\cdot \mathbb{E}_{p_{{id}}}[-\text{PHP}(\xv)] / \mathbb{E}_{p_{{id}}}[\mathcal{C}(\xv)],
\end{aligned}   
\end{small}
\end{equation}
whose expectation of the data distribution could be approximated as
\begin{equation}
\begin{small}
\begin{aligned}
\mathbb{E}_{\xv\sim p(\xv)}[\text{DEC}(\xv)] \approx -D_{\text{KL}}[q(\zv)||p(\zv)] - \text{Ent-Mut}(\theta, \phi, p)\cdot (1 - \mathbb{E}_p[\mathcal{C}(\xv)]/\mathbb{E}_{p_{id}}[\mathcal{C}(\xv)]).
\label{eq:expectation_dec}
\end{aligned}   
\end{small}
\end{equation}
Its incremental effectiveness $\mathcal{A}(\text{DEC})$ becomes
{\small
\begin{align}
    \mathcal{A}(\text{DEC}) = \mathcal{G}(\text{DEC}) - \mathcal{G}(\log p_\theta)
    = (\mathbb{E}_{p_{id}}[\mathcal{C}(\xv)] - \mathbb{E}_{p_{ood}}[\mathcal{C}(\xv)])\cdot \mathbb{E}_{p_{id}}[-\text{PHP}(\xv)] / \mathbb{E}_{p_{id}}[\mathcal{C}(\xv)] {\geq}0. 
\end{align}
}

\subsection{Resultant: A New Detector Combining Two Directions}
Finally, we propose a new U-OOD detector combining the methods for two directions, named \textit{Resultant} $\mathcal{S}_{R}(\xv)$, defined by
\begin{equation}
\begin{small}
\begin{aligned}
    \mathcal{S}_{R}(\xv) =  \mathbb{E}_{q_{\phi}(\zv|\xv)} \left[ \log p_{\theta}(\xv|{\zv}) \right] \label{eq:avoid_loss} - D_{\text{KL}}[q_{\phi}(\zv|\xv)||\hat{q}_{{id}}(\zv)] + 
    {\mathcal{C}(\xv) \cdot \frac{\mathbb{E}_{p_{id}}[-\text{PHP}(\xv)]}{\mathbb{E}_{p_{id}}[\mathcal{C}(\xv)]}}.
\end{aligned}   
\end{small}
\end{equation}
As PHP only focuses on the latent distribution mismatch as in Eq. \ref{eq:Expectation_PHP} and DEC focuses on the Ent-Mut term as in Eq. \ref{eq:expectation_dec} if the reweighting scale satisfies Eq. \ref{eq:max_direction1}, there would be no conflicts between the PHP and DEC's contribution to incremental effectiveness on likelihood, leading to
\begin{equation}
\begin{small}
\begin{aligned}
    \mathcal{A}(\mathcal{S}_R) \geq  \mathcal{A}(\text{PHP}) \text{ and } \mathcal{A}(\mathcal{S}_R) \geq \mathcal{A}(\text{DEC}).
    \notag
\end{aligned}   
\end{small}
\end{equation}
Intuitively, as the PHP method mainly focuses on the semantic information and the DEC method focuses on the statistical information, the \textit{Resultant} detector $\mathcal{S}_{R}(\xv)$ should work in a wider range of scenarios in detecting semantic and/or statistic difference than either technique alone.

\section{Experiment}
\vspace{-1mm}
\subsection{Experimental Setup}
\label{sec:exp_setup}

\textbf{Baselines.} \label{sec:baseline}
We compare our method with DGM-based U-OOD detection methods and assess whether we can achieve state-of-the-art performance, and more importantly, achieve incremental effectiveness on likelihood.
For comparisons, we compare our method with a standard VAE \cite{kingma2013auto}, denoted as ``{Likelihood}'', which also serves as the foundation of all methods.
We also testify to the extension of our method to the diffusion models as detailed in { Appendix \ref{sec_re_diff}}.
Other baselines are introduced in Section \ref{sec:background}, including log-likelihood ratio methods ({HVK} \cite{hvae_ood} and {$\mathcal{LLR}^{ada}$} \cite{informative-vae}) and {Likelihood Regret} \cite{xiao}, {Likelihood Ratio} \cite{likelihood-ratio-jie}, {Input Complexity} with a PNG compressor \cite{input_complexity}, and {WAIC} \cite{choi2018waic}. 

\textbf{Implementation Details.} For implementation on the VAE, its latent variable's dimension is set as 200 for all experiments with the encoder and decoder parameterized by a 3-layer convolutional neural network, respectively. 
The reconstruction likelihood distribution is modeled by a discretized mixture of logistics \cite{salimans2017pixelcnn++}. For optimization, we adopt the Adam optimizer \cite{kingma2014adam} with a learning rate of 1e-3. We train all models in comparison by setting the batch size as 128 and the max epoch as 1000 following \cite{hvae_ood, informative-vae}. 
All experiments are performed on a PC with an NVIDIA A100 GPU and implemented with PyTorch \cite{paszke2019pytorch}. 
All results and error bars are calculated over 5 random runs.
More details of the encoder, decoder, PHP, and DEC can be seen in {Appendix \ref{sec:appen_exp_imples}}.

{{Due to page length limit}, we leave the common descriptions of the \textbf{Datasets} in Appendix \ref{sec:appen_exp_datasets} and \textbf{Evaluation Metrics} in Appendix \ref{sec:appen_eval_metrics}, which is following the previous works' procedure \cite{hvae_ood}.}

\vspace{-1mm}
\subsection{Comparison with U-OOD Detection Baselines}
We compare our methods (PHP, DEC, and Resultant) with the likelihood and some unsupervised methods on the five commonly acknowledged \textbf{``hard benchmarks''} \cite{hard_datasets} and their reverse directional verifications in Table \ref{tab:main_tab}.
As observed in Table \ref{tab:main_tab}, our Resultant could achieve state-of-the-art performance while satisfying incremental effectiveness on likelihood.
Though some methods could achieve general good performance in reverse verification experiments such as the Likelihood Ratio, they have no incremental effectiveness on likelihood.
To showcase our method's incremental effectiveness could be satisfied in various scenarios, we further compare Resultant with the likelihood in a wide range of datasets in Table \ref{tab:comparison_more_datasets}, and move more results on FashionMNIST (ID) and CelebA (ID) to {Appendix \ref{sec:app_re_avoid}.}
To ease the reading of experimental results, we put only VAE-based experiments on the main paper to serve as a proof of concept and move the \textbf{diffusion-based experiments} to {Appendix \ref{sec_re_diff}} to demonstrate the extension ability of the analysis and methods.
Experimental results strongly verify our analysis of the likelihood and demonstrate the incremental effectiveness of our methods.

\begin{table}[h!]
\centering
\caption{Comparison of U-OOD detection methods on \textbf{``Hard'' Benchmarks} and their corresponding reverse verification under the metric of AUROC $\uparrow$. The best performance is \textbf{bold}. 
{ ``Likelihood'' and ``Resultant'' (ours) are in \colorbox[RGB]{217,217,255}{purple} for comparison regarding the incremental effectiveness.}
}
\setlength\tabcolsep{7pt}
\renewcommand{\arraystretch}{1.2}
\resizebox{\linewidth}{!}{
\begin{tabular}{lccccc|ccccc}
\hline
                        & \multicolumn{5}{c|}{\textbf{``Hard'' Benchmarks}}                                                                                       & \multicolumn{5}{c}{\textbf{Reverse Verification}}                                                                                       \\ \hline
ID Dataset              & CIFAR-10             & FashionMNIST         & CelebA               & CelebA               & CIFAR-100             & SVHN                 & MNIST                & CIFAR-10             & CIFAT-100            & SVHN                 \\
OOD Dataset             & SVHN                 & MNIST                & CIFAR-10             & CIFAT-100            & SVHN                  & CIFAR-10             & FashionMNIST         & CelebA               & CelebA               & CIFAR-10O            \\ \hline
\rowcolor[HTML]{D9D9FF}
 Likelihood \cite{kingma2013auto}              & 24.9                 & 23.5                 & 27.8                 & 33.1                 & 7.68                  & \textbf{99.9}                 & \textbf{99.9}                 & 57.2                 & 58.2                 & \textbf{99.9}                 \\
HVK \cite{hvae_ood}                    & 89.1                 & 98.4                 & 40.1                 & 45.2                 & 52.3                  & 64.2                     &  80.5                    & 67.1                     & 69.3                     & 63.0                     \\
$\mathcal{LLR}^{ada}$ \cite{informative-vae}  & 94.2                 & 98.0                 & 58.0                 & 52.5                 & 86.7                  & 34.7                     & 83.0                     & \textbf{72.1}                     &  73.5                    & 36.0                     \\
Likelihood Ratio \cite{likelihood-ratio-jie}       &  29.8                    & 96.1                     & 67.8                     & 67.7                     & 24.6                      & 97.9                     &  \textbf{99.9}                    & 70.3                     & 64.7                     & 98.4                     \\
Likelihood Regret \cite{xiao}      & 85.5                     & 96.7                     & 70.5                     & 67.7                     & 34.9                      &  94.8                    &  99.8                    & 61.6                     & 52.8                     & 93.1                     \\
Complexity (PNG) \cite{input_complexity}       &75.7                      & 99.0                     & \textbf{75.9}                     &75.3                      &73.2                       & 99.3                     &{99.0}                      &50.7                      & 49.9                     & {98.9}                     \\
WAIC (5VAE) \cite{choi2018waic}              & 94.3                     & 86.6                     & 45.5                     &  45.7                    & 65.9               & 96.9                     & \textbf{99.9}                 & 54.5                     & 52.8                     &91.1                      \\
\textit{\textbf{-ours}} &                      &                      &                      &                      &                       &                      &                      &                      &                      &                      \\
PHP                     & 39.6                 & 89.7                 & 69.5                 & 68.9                 & 24.4                  & 99.9                 & 99.9                 & 70.9                 & 65.0                 & 99.9                 \\
DEC                     & 87.8                 & 34.1                 & 73.3                 & 73.7                 & 81.8                  & 99.9                 & 99.9                 & 69.0                 & 72.4                 & 99.9                 \\
\rowcolor[HTML]{D9D9FF}
 Resultant                   & \textbf{94.5}        & \textbf{99.2}        & {75.6}        & \textbf{75.5}        & \textbf{88.3}         & \textbf{99.9}        & \textbf{99.9}        & {71.2}        & \textbf{74.8}        & \textbf{99.9}        \\ \hline
\end{tabular}
}
\label{tab:main_tab}
\end{table}

\begin{table}[t]
\centering
\caption{The comparisons between ``Likelihood'' and our method ``Resultant'' on more datasets {(``Likelihood'' / ``Resultant'')}. \textbf{Bold} numbers are superior performance.}
\setlength\tabcolsep{20pt}
\renewcommand{\arraystretch}{1.3}
\scalebox{0.95}{
\resizebox{\linewidth}{!}{
\begin{tabular}{c|ccc|ccc}
\hline
\multirow{2}{*}{OOD} & \multicolumn{3}{c|}{CIFAR-10 (ID)}                                 & \multicolumn{3}{c}{CIFAR-100 (ID)}                                                            \\ \cline{2-7} 
                     & AUROC $\uparrow$     & AUPRC $\uparrow$     & FPR80 $\downarrow$   & AUROC $\uparrow$              & AUPRC $\uparrow$              & FPR80 $\downarrow$            \\ \hline
CelebA               & 57.2 / \textbf{71.2} & 54.5 / \textbf{72.1} & 69.0 / \textbf{54.4} & 58.2 / \textbf{74.8}          & 56.0 / \textbf{68.6}          & 65.8 / \textbf{35.7}          \\
SUN                  & 53.1 / \textbf{63.0} & 54.4 / \textbf{63.3} & 79.5 / \textbf{68.6} & 58.3 / \textbf{70.6}          & 55.6 / \textbf{62.6}          & 61.0 / \textbf{39.1}          \\
Places365            & 57.2 / \textbf{68.3} & 56.9 / \textbf{69.0} & 73.1 / \textbf{62.6} & 56.5 / \textbf{80.0}          & 55.5 / \textbf{74.0}          & 74.6 / \textbf{27.7}          \\
LFWPeople            & 64.1 / \textbf{67.7} & 59.7 / \textbf{68.8} & 59.4 / \textbf{54.4} & 63.9 / \textbf{81.5}          & 58.4 / \textbf{75.4}          & 61.3 / \textbf{26.3}          \\
Texture              & 37.8 / \textbf{81.8} & 40.9 / \textbf{62.4} & 82.2 / \textbf{64.3} & 52.7 / \textbf{75.0}          & 48.4 / \textbf{65.8}          & 66.1 / \textbf{31.5}          \\
Flowers102           & 67.6 / \textbf{76.8} & 64.6 / \textbf{78.0} & 57.9 / \textbf{46.6} & 80.5 / \textbf{92.3}          & 80.1 / \textbf{87.4}          & 23.9 / \textbf{9.20}           \\
GTSRB                & 39.5 / \textbf{53.0} & 41.7 / \textbf{49.8} & 86.6 / \textbf{73.6} & 58.7 / \textbf{75.3}          & 51.8 / \textbf{68.8}          & 49.4 / \textbf{35.3}          \\
Const                & 0.01 / \textbf{80.1} & 30.7 / \textbf{89.4} & 99.9 / \textbf{22.3} & 0.00 / \textbf{79.8}           & 0.00 / \textbf{80.8}           & 99.9 / \textbf{22.1}          \\
Random               & 71.8 / \textbf{99.3} & 82.9 / \textbf{99.5} & 85.7 / \textbf{0.00}  & \textbf{99.9} / \textbf{99.9} & \textbf{99.9} / \textbf{99.9} & \textbf{0.00} / \textbf{0.00} \\ \hline
\end{tabular}
}
}
\label{tab:comparison_more_datasets}
\end{table}

\vspace{-1mm}
\subsection{Ablation Study on Verifying the Post-hoc Prior Method}
\label{sec:ablation_php}

\begin{figure}[t]

    \centering
    \subfigure[Density plot of $\log p_\theta$]{\includegraphics[width=0.23\textwidth]{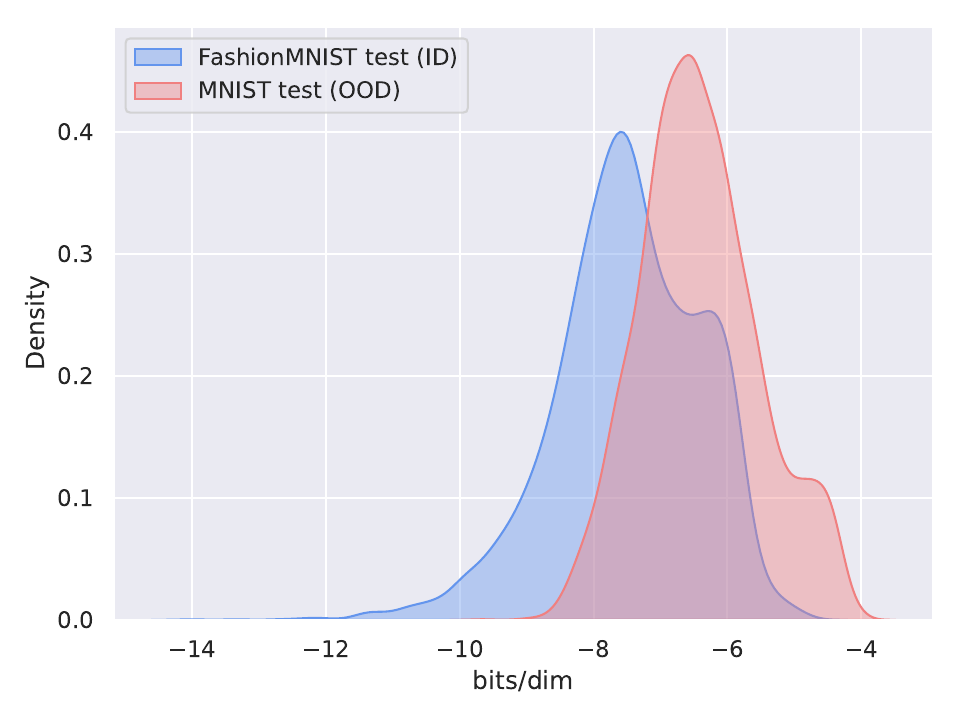}
    \label{fig:elbo_fm_m}}
    \subfigure[Density plot of PHP]{\includegraphics[width=0.23\textwidth]{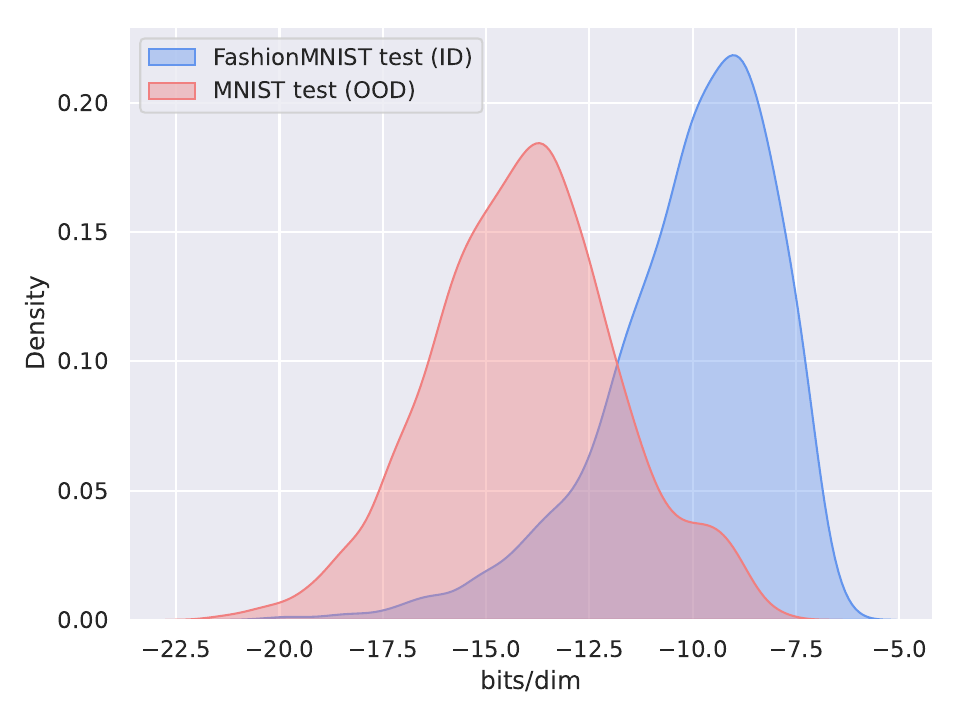}
    \label{fig:php_fm_m}}
    \subfigure[ROC curve of $\mathcal{LLR}$]{\includegraphics[width=0.23\textwidth]{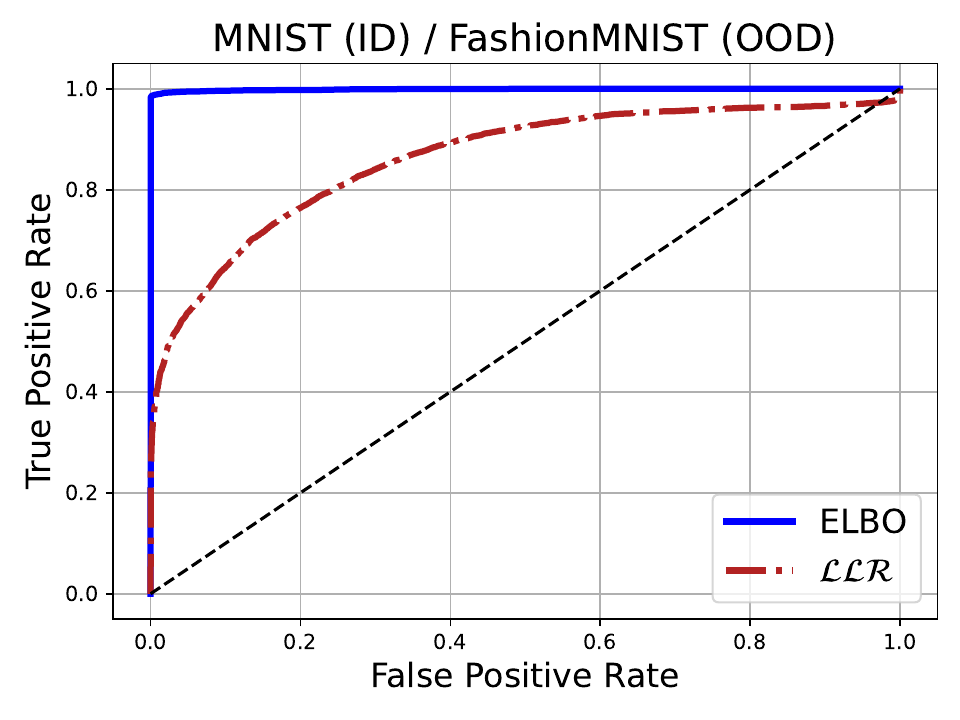}
    \label{fig:LLR_m_fm}}
    \subfigure[ROC curve of PHP]{\includegraphics[width=0.23\textwidth]{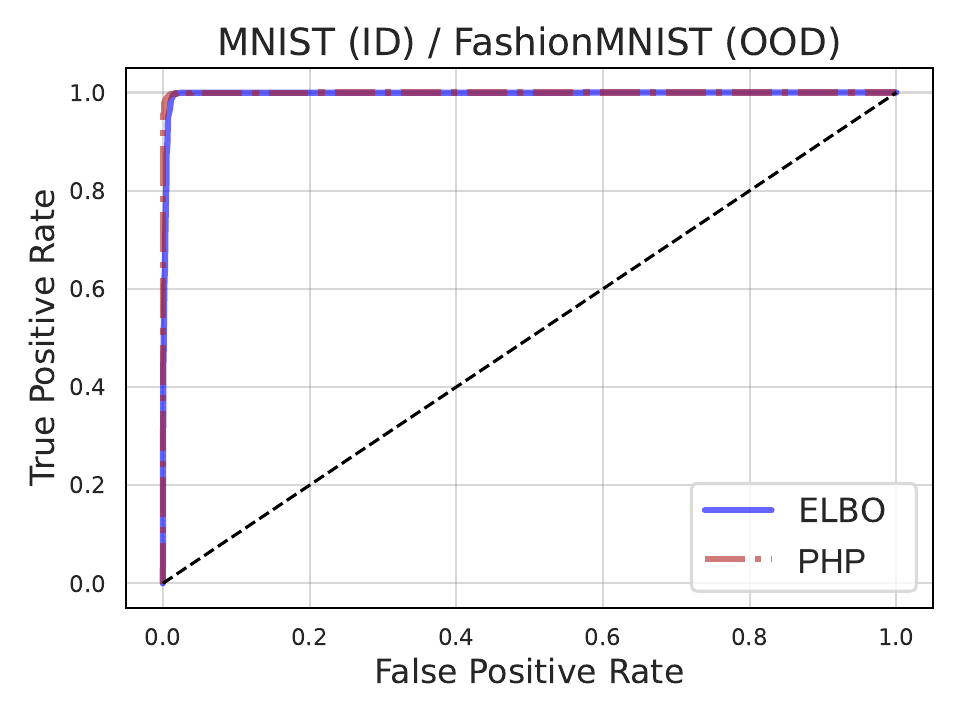}
    \label{fig:AVOID_m_fm}}
    \vspace{-3mm}
    \caption{Density plots and ROC curves. \textbf{(a):} Using the likelihood of a VAE trained on FashionMNIST leads to overestimation when detecting MNIST as OOD data, resulting in limited detection performance; \textbf{(b):} PHP could improve the detection performance; \textbf{(c):} SOTA method $\mathcal{LLR}^{ada}$ could degenerate the performance of likelihood; \textbf{(d):} PHP method satisfy the incremental effectiveness.
    }
    \label{fig:kl_exp}

\end{figure}

The comparison between the Post-hoc Prior (PHP) and the likelihood could be seen in Table \ref{tab:main_tab}, indicating its incremental effectiveness. {Moreover, we test the PHP method on additional datasets and present the results in Table \ref{tab:comparison_more_datasets_kl} and \ref{tab_cifar100_php} of Appendix \ref{sec:appen_ablation_prior}.   
A special experiment in Table \ref{tab_vflip} of Appendix \ref{sec_re_vflip} is designed to testify to the PHP method for detecting \textbf{vertically flipped OOD data}, where direction I dominates the performance improving direction of Resultant.}
These experimental results demonstrate that the PHP method's incremental effectiveness. 
For cases where the PHP method does not significantly improve detection, we have included a detailed discussion with visualizations in Fig.~\ref{fig:umap} of Appendix \ref{sec_re_umap}.
To provide a better understanding, we visualize the density plot of likelihood and PHP for the ``FashionMNIST(ID)/MNIST(OOD)'' dataset pair in Figs.~\ref{fig:elbo_fm_m} and \ref{fig:php_fm_m}, respectively.
Besides, we provide the ROC curve figures in Fig.~\ref{fig:LLR_m_fm} and Fig.~\ref{fig:AVOID_m_fm}, demonstrating that while likelihood could already perform well in detection, the $\mathcal{LLR}^{ada}$ method \cite{informative-vae} could negatively impact detection performance to some extent. On the contrary, our method can still maintain comparable performance since the PHP method is designed following the incremental effectiveness.

\vspace{-1mm}
\subsection{Ablation Study on Verifying Dataset Entropy-mutual Calibration Method}

\begin{figure}[h!]
\vspace{-3mm}
    \centering
    \subfigure[scaled ${\mathcal{C}}(\xv)$ of ID.]{\includegraphics[width=0.3\textwidth]{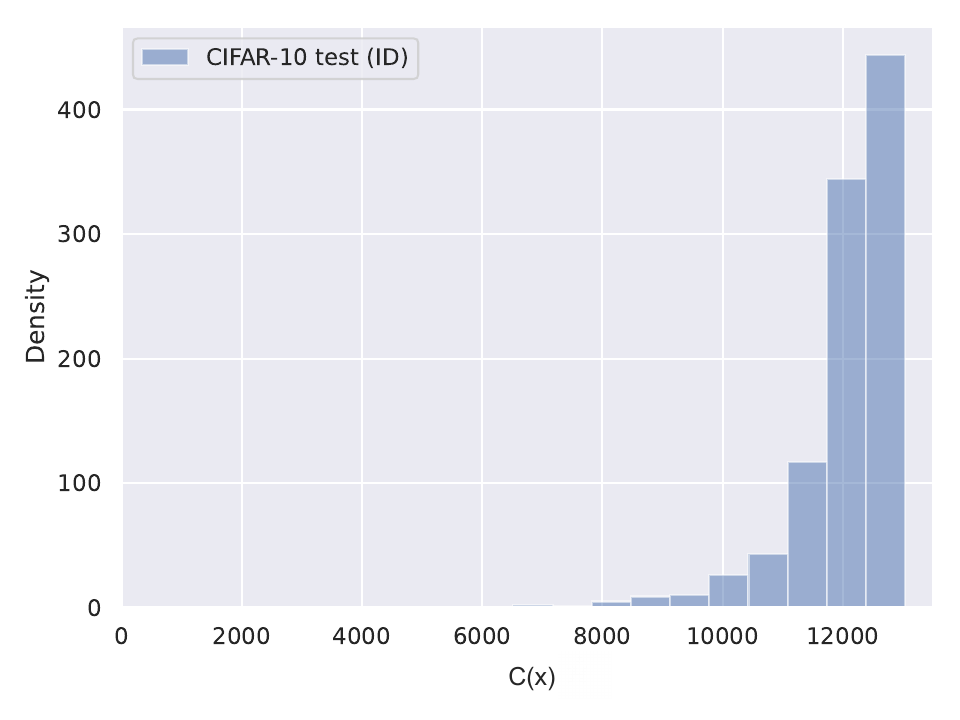}
    \label{fig:hx_id}}
    \subfigure[scaled ${\mathcal{C}}(\xv)$ of OOD.]{\includegraphics[width=0.3\textwidth]{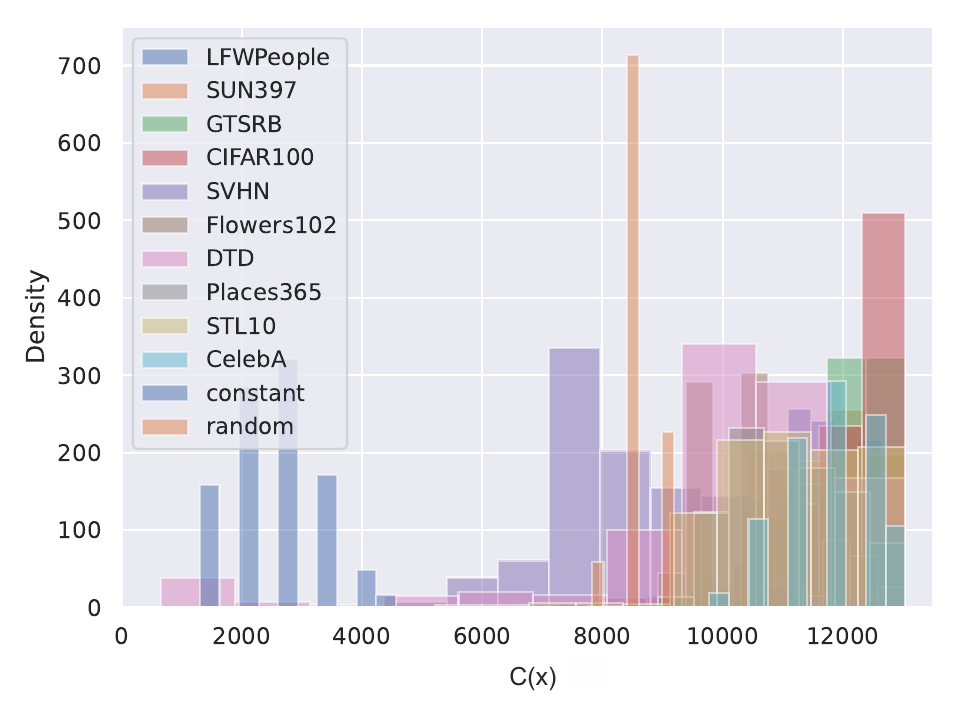}
    \label{fig:hx_ood}}
    \subfigure[Impact of $n_{{id}}$]{\includegraphics[width=0.3\textwidth]{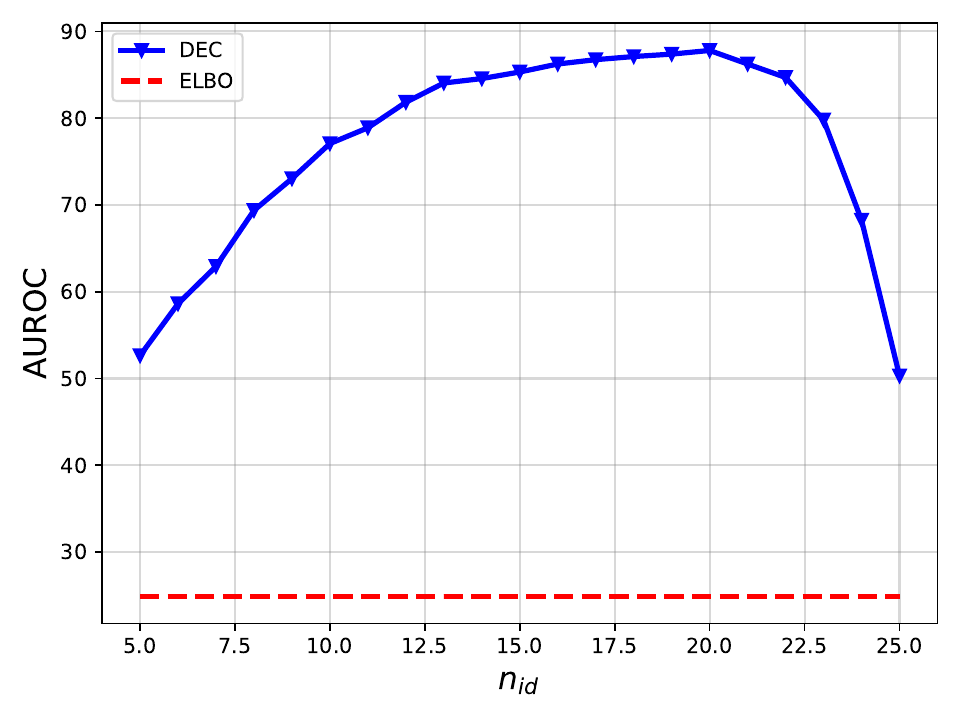}
    \label{fig:n_id_ablation}}
    \vspace{-2mm}
    \caption{\textbf{(a)} and \textbf{(b)} are respectively the visualizations of the scaled calculated entropy-mutual calibration ${\mathcal{C}}(\xv)$ of CIFAR-10 (ID) and other OOD datasets, where the ${\mathcal{C}}(\xv)$ of CIFAR-10 (ID) could achieve generally higher values. \textbf{(c)} is the OOD detection performance of dataset {entropy-mutual} calibration with different $n_{{id}}$ settings, which consistently outperforms likelihood.
    }
    \label{fig:ent_exp}
    \vspace{-5mm}
\end{figure}

We evaluate the performance of {Dataset Entropy-mutual Calibration} (DEC) method in Table \ref{tab:main_tab} and Tables \ref{tab:comparison_more_datasets_ent} and \ref{tab_cifar100_dec} of Appendix \ref{sec:appen_ablation_ent}. Our results show that it effectively improves the U-OOD detection performance.
A special task on detecting \textbf{noisy data} as OOD is designed to support the analysis of Direction II and DEC's effectiveness in this task, as shown in Tables \ref{tab_noisy_fm} and \ref{tab_noisy_cf} of Appendix \ref{sec_re_noisy}.
To better understand DEC, we visualize the calculated ${\mathcal{C}}(\xv)$ of CIFAR-10 (ID) in Fig.~\ref{fig:hx_id} and other OOD datasets in Fig.~\ref{fig:hx_ood} when $n_{{id}} = 20$. Our results show that the ${\mathcal{C}}(\xv)$ of CIFAR-10 (ID) achieves generally higher values than that of other datasets, which is the underlying reason for its effectiveness. Additionally, we investigate the impact of different $n_{{id}}$ on OOD detection performance in Fig.~\ref{fig:n_id_ablation}, where our results show that the performance {of DEC} is consistently better than the likelihood, supporting its incremental effectiveness.

\section{Discussion on Extension to other DGMs} \label{sec:discuss_extenstion_dgms}
Although our main paper focuses on variational deep generative models (DGMs) to illustrate a new method for achieving incremental effectiveness, this approach can also be applied to other DGMs, such as normalizing flow models (NFs) \cite{realnvp, glow}, auto-regressive models (ARs) \cite{pixelcnn,salimans2017pixelcnn++}, energy-based models (EBMs) \cite{lecun2006tutorial_ebm, train_ebm}, and score-based methods \cite{score-based}. For NFs, which have a structure similar to variational autoencoders (VAEs) in using a latent distribution for likelihood estimation, adjusting the latent distribution from a standard Gaussian to the aggregated posterior of an in-distribution (ID) dataset can enhance model performance by increasing the likelihood for ID data. As DEC is adaptable across different DGM types to modify the impact of entropy, it could also be applied to NFs and all other DGMs. 
For ARs, the hierarchical variational DGM has an intrinsic relationship to them \cite{DDPM}, as the likelihood in Eq. \ref{eq:hierarchical} could be rewritten as $\log p_\theta(\xv) = D_{\text{KL}}(q(\xv_T)||p(\xv_T)))+\mathbb{E}_q[\sum_{t\geq 1}D_{\text{KL}}(q(\xv_{t-1}|\xv_t)||p(\xv_{t-1}|\xv_t))] + \mathcal{H}_p(\xv_0)$ that is actually learning an AR $\log p_\theta(\xv)=\log [p_\theta(\xv_T)\prod\nolimits_{t=1}^{T}p_\theta(\xv_{t-1}|\xv_t)]$.
For EBM, which estimates likelihood by $p_\theta(\xv) = \frac{\exp(-E_\theta(\xv))}{Z_\theta(\xv)}$, has recently been proposed to be stably trained in a diffusion recovery likelihood \cite{diffusion_recovery_ebm, train_ebm} that bridges the EBM with the hierarchical variational DGMs.
For the score-based method, it unifies two powerful DGMs involving score matching with Langevin dynamics (SMLD) \cite{smld} and denoising diffusion probabilistic modeling (DMs) (\textit{i.e.}, the variational DGM as discussed above), and use the neural ordinary differential equations (ODE) solver \cite{score-based} for estimating the likelihood.
As the uncovered relationship between SMLD and hierarchical variational DGMs, our analysis and method hold the potential to be applied to it.

Among these DGMs, we would highly recommend implementation on VAEs, as it could be lightweight in computation, high-speed in inference, and powerful enough in U-OOD detection, making it suitable for devices with limited computational resources. However, we suggest exploring specific implementations with more DGMs in future work.

\vspace{-2mm}
\section{Conclusion}
\vspace{-2mm}
This work highlights the incremental effectiveness on likelihood in U-OOD detection
and develops a novel score function called ``Resultant'', which is effective in U-OOD detection. 
{This work may lead a research stream for improving U-OOD detection by developing more efficient and sophisticated methods aimed at optimizing the identified two directions. 

There are still some \textbf{limitations} such as the simplicity of the developed methods that may under-explore the full capabilities of DGMs on U-OOD detection.
Additionally, the extra model introduced by the PHP might be superfluous if future research can uncover the intrinsic properties of DGMs to approximate the ID data aggregated posterior distribution, although this remains a challenging task.
}





\bibliography{reference}
\bibliographystyle{plain}

\newpage
\appendix
\startcontents[appendices]
\printcontents[appendices]{}{1}{\section*{Appendix}\setcounter{tocdepth}{2}}

\newpage
\section{Background on OOD Detection} \label{sec:app_differnce}

To provide a clear distinction and avoid confusion between supervised and U-OOD detection, we delineate the key differences here, primarily focusing on their respective setups.

\textbf{Setup of \emph{unsupervised} OOD detection.} 
{Denoting the input space with $\mathcal{X}$, an \textit{unlabeled} training dataset $\mathcal{D}_{\text{train}}=\{\xv_i\}_{i=1}^{N}$ containing of $N$ data points can be obtained by sampling \textit{i.i.d.} from a data distribution $\mathcal{P}_{\mathcal{X}}$.}
Typically, we {treat} the $\mathcal{P}_{\mathcal{X}}$ as ${p}_{{id}}$, which represents the in-distribution (ID) \cite{hvae_ood, flow_cannot}. With this \textit{unlabeled} training set, U-OOD detection is to design a score function $\mathcal{S}(\xv)$ that can determine whether an input is ID or OOD.

\textbf{Setup of \emph{supervised} OOD detection.}
Compared with the setup of U-OOD detection, supervised one needs to additionally introduce a label space $\mathcal{Y}=\{1,...,k\}$ with $k$ classes, and the training set becomes $\mathcal{D}_{\text{train}}=\{(\xv_i, \yv_i)\}_{i=1}^{N}$. Then, it typically needs to train a classifier $f:\mathcal{X}\rightarrow \mathbb{R}^{k}$,  and OOD detection 
{can be} achieved based on the property of the classifier \cite{wei2021odnl, wei@open, wei2022logitnorm}.

\begin{figure}[h!]
    \centering
    \includegraphics[width=0.9\textwidth]{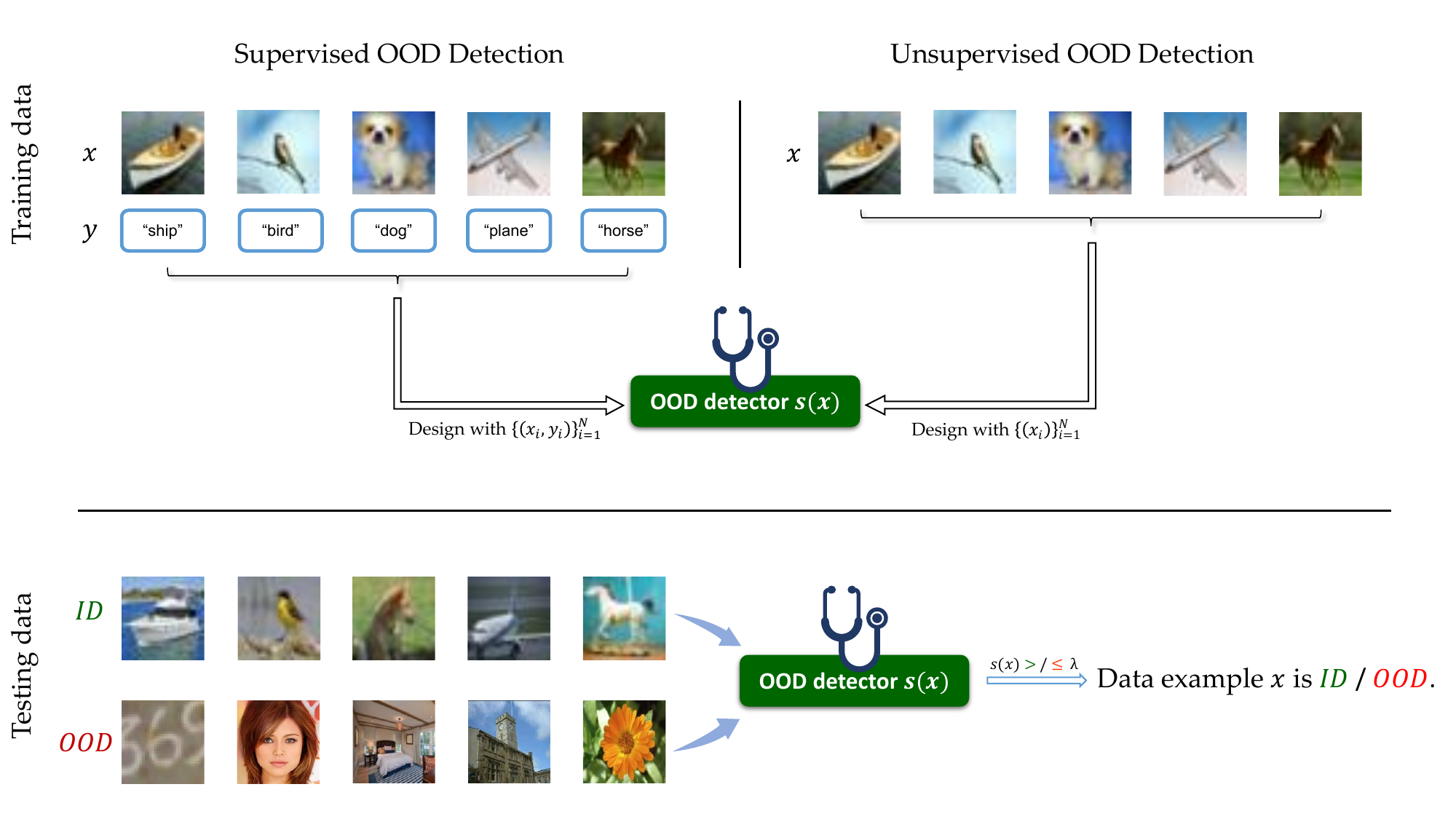}
    \caption{An illustration showcasing the difference between supervised and U-OOD detection.}
    \label{fig:super_unsuper}
\end{figure}

We illustrate the distinction between supervised and U-OOD detection in Fig.~\ref{fig:super_unsuper}.

{
We also present a discussion here about the methods for improving OOD detection performance in the supervised case (especially the classifier-based methods), which could be divided into two categories, and \textbf{draw parallels between the unsupervised and supervised cases}:
\begin{enumerate}
    \item Designing a score function based on the properties of a \textbf{trained} classifier, such as maximum softmax probability \cite{hendrycks2016baseline}, Mahalanobis distance-based score \cite{lee2018simple}, energy-based score \cite{DBLP:conf/aaai/MortezaL22}, and GradNorm score \cite{HuangGL21}. These methods operate on the premise that the statistical characteristics exhibited by the classifier when presented with an in-distribution (ID) data example are distinct from those observed when handling an out-of-distribution (OOD) data example. For unsupervised case, without the label to train a classifier, it is similar to designing a score function based on a trained deep generative model (DGM), such as HVK \cite{hvae_ood} that exploit the relationship between the posterior and prior latent distribution existed in a trained VAE to do OOD detection. More methods can be seen in Appendix \ref{sec:app_vae_related}. Our method also focuses on a trained DGM, which has the advantage of "plug-and-play".
    
    \item Introducing regularization techniques during the classifier's \textbf{training phase}, such as adding a fix to the cross-entropy loss \cite{wei2022logitnorm}, encouraging the classifier to give predictions with uniform distribution for OOD data \cite{hendrycks2018deep}, and shaping the log-likelihood by energy-based regularizer \cite{DBLP:conf/icml/Katz-SamuelsNNL22}. After training with the regularizer, the classifier exhibits differing statistics between ID and OOD data. There are also some similar works that modify the training objective of the DGM in unsupervised cases, \textit{e.g.}, the 
    $\mathcal{LLR}^{ada}$ \cite{informative-vae} adds a partial generation term into the ELBO during training a VAE that could explicitly enhance the semantic information's quality in the latent variables that could be helpful for OOD detection. More related works can be seen in Appendix \ref{sec:app_vae_related}. 
\end{enumerate}

}

\section{Related Work} \label{sec:appen_related_work}

\subsection{Deep Generative Models}
Deep Generative Models (DGMs) have been developed with the aim of modeling the true data distribution $p(\xv)$, leveraging deep neural networks to learn a generative process \cite{goodfellow2016deep}. These models span several types, mainly including the autoregressive model \cite{salimans2017pixelcnn++, pixelcnn}, flow model \cite{realnvp, glow}, generative adversarial network \cite{gan}, energy-based model \cite{lecun2006tutorial_ebm}, diffusion model \cite{DDPM}, and variational autoencoder (VAE) \cite{kingma2013auto}. Below, we briefly introduce each of these models:
The autoregressive model operates under the premise that a data sample $\xv$ is a sequential series, implying that the value of a pixel in an image is only dependent on the pixels preceding it. 
The flow model comes with an inherent requirement for the invertibility of the projection between $\xv$ and $\zv$, which imposes constraints on the implementation of its backbone. 
The generative adversarial network adopts an additional discriminator to implicitly learn the data distribution. Despite its power, it faces challenges such as unstable training and mode collapse \cite{trilema}.
The diffusion model could be formulated as a VAE as its training objective could also be derived from an evidence lower bound of the data distribution \cite{DDPM, unify}.
Among these models, 
{VAEs} and diffusion models stand out for their flexibility in implementation and comprehensive mode coverage \cite{trilema}. 
Once the DGM is adequately trained, the estimated likelihood function $p_\theta(\xv)$ should possess some capability for U-OOD detection.

{
\subsection{Analysis for Likelihood's Failure in OOD Detection}
\label{sec: app_survey_analysis}
Direct application of the likelihood as a U-OOD detector is challenging and can lead to a \textit{paradox}, as reported by previous studies. Specifically, the likelihood assigned to OOD data may be higher than that for ID data, resulting in poor U-OOD detection performance \cite{likelihood-ratio-jie}.
{This failure of deep generative models' likelihood in U-OOD detection remains a challenging and unresolved issue. This may be attributed to two primary factors: the characteristics of the data and the capabilities of the models.}

{For the analysis of the data characteristics, some works found the image properties, \textit{e.g.}, the background pixels \cite{likelihood-ratio-jie}, image intensity and contrast \cite{de-bias-vae}, and high-frequency information \cite{frl}, could lead to the failure. However, these findings are empirical and cannot be unified across different data types limiting their methods' generalizability. 
Thus, some works provide a more universal hypothesis for the occurrence of the {paradox}, \textit{i.e.}, the typical set hypothesis, where ID and OOD data have overlapping supports \cite{hard_datasets, choi2018waic, typical_1, dose}.  However, some proved that this hypothesis is not correct and appeal for more future works on analyzing the estimation error of the DGMs \cite{typical_is_wrong, typical_penalty}. 
}

{
For the analysis of the model capacity, 
a representative analysis \cite{entropic_issue} breaks down the expectation of the likelihood into two terms: a KL term indicating the distance between the estimated data distribution and the true data distribution and an entropy term of the data distribution for analyzing the failure of DGMs in U-OOD detection. This analysis is related to us, however, our analysis provides further insight into the the likelihood. Besides, perfectly
matching the data distribution via DGMs is too ideal and
even impossible in practice \cite{bindai_1, bindai_2}. This is also why we need to go deeper into the likelihood of the DGMs. 
\cite{flow_cannot} further find the paradox could arise from the intrinsic model curvature brought by the invertible architecture in flow models. \cite{flow_cannot_2} claim that normalizing flows learns latent representations based on local pixel correlations and not in-distribution data's specific semantic content, which leads to the failure. \cite{input_complexity} empirically find that the image complexity could be a reason for the paradox of flow and auto-regressive models, but this finding is not available in cases where the ID and OOD data share similar image complexity. \cite{anomaly_invertible_net} provide analysis for the deep invertible networks through hierarchies of distributions and features.
\cite{manifold_overfit} analyze the failure of DGMs as a manifold overfitting issue and appeals for projecting the high-dimension input data to a lower-dimension representation first and then fitting the lower-dimension representation, instead of fitting the likelihood in the high-dimension space directly. 
It could be directly related to our PHP method but the DEC method is still needed to further alleviate the paradox.}
Additionally, most of these analyses are for DGMs like flow and auto-regressive models and none of them provide a theoretical and comprehensive analysis for the variational DGMs like VAEs, where their training objective is a variational evidence lower bound bringing extra difficulties for analysis. Our work also serves to fill this gap in the analysis of variational DGMs.

However, the fundamental reasons behind the likelihood's failure in U-OOD detection remain challenging and unresolved. While our work does not resolve this issue, it aims to advance understanding by focusing on model likelihood, potentially inspiring future research to further explore and address these failures.

\subsection{DGM-based Unsupervised OOD Detection}\label{sec:app_vae_related}
Based on the necessity to modify the training of a DGM, these methods can be categorized into two groups.
\textbf{{\emph{i})}}
The first group includes methods that modify the \textbf{training} scheme. Hierarchical VAE expands the single-level VAE's layers to augment its representational capacity \cite{biva}, yet the improvements in performance are marginal, and the paradox persists. The adaptive log-likelihood ratio method, $\mathcal{LLR}^{ada}$, is also grounded in the hierarchical VAE and introduces a generative skip connection to propagate information to higher layers of latent variables \cite{informative-vae}. It utilizes the differences between each layer of latent variables for U-OOD detection, achieving state-of-the-art performance despite certain shortcomings on incremental effectiveness. The tilted VAE enforces the latent variable to exist within the sphere of a tilted Gaussian \cite{unsuper_add1}, thereby disrupting the efficient, widely adopted reparameterization based on the Gaussian. It should be noted that modifying the training of DGMs may be less practical as the proposed method cannot be directly applied to other DGMs trained with special setting like a two-step training setting \cite{manifold_overfit}. Thus, we focus on developing methods for trained DGMs.
\textbf{{\emph{ii})}}
The second group of methods attempts to utilize the properties of a \textbf{trained} DGM for OOD detection without modifying it. The likelihood-ratio method simulates the background using noise and employs the difference between the original and simulated background images for OOD detection \cite{likelihood-ratio-jie}. The likelihood-regret method finetunes the trained VAE with the test sample to observe changes in likelihood \cite{xiao}. The log-likelihood ratio method leverages the assumption that latent variables of lower layers capture low-level features of inputs while those of higher layers grasp semantic features \cite{hvae_ood}. The difference between these latent variables can then be used for OOD detection. WAIC utilizes empirical ensemble methods for OOD detection \cite{choi2018waic}. However, it should be emphasized that none of these methods provide explicit or empirical guarantees regarding incremental effectiveness on likelihood, which in some cases results in performance that is inferior to that of direct likelihood approaches.

\section{Revisit Existing Literature Regarding the Incremental Effectiveness on Likelihood}\label{app:sec_revisit}
We provide a comprehensive revisitation of the existing DGM-based U-OOD detection methods from the perspective of incremental effectiveness on likelihood, including \textbf{Log-likelihood Ratio} \cite{hvae_ood}, \textbf{Likelihood Regret} \cite{xiao}, \textbf{Likelihood Ratio} \cite{likelihood-ratio-jie}, \textbf{Input Complexity} \cite{input_complexity}, and \textbf{WAIC} \cite{choi2018waic}.

\subsection{Log-likelihood Ratio} 
\paragraph{Method description.} Log-likelihood Ratio methods \cite{hvae_ood, informative-vae} uses a score function with a selected specific layer $k$ for U-OOD detection, formulated as 
\begin{align}
    \mathcal{LLR}^{>k} &=D_{\text{KL}}[p_\theta(\zv_{\leq k}|\zv_{>k})q_\phi(\zv_{>k}|\xv)||p_\theta(\zv|\xv)] 
    = \log p_\theta(\xv) - \log p_\theta(\xv, k) \notag \\
    &= \log p_\theta(\xv) - \mathbb{E}_{p_\theta(\zv_{\leq k}|\zv_{>k})q_\phi(\zv_{>k}|\xv)}[\log \frac{p_\theta(\xv|\zv)p_\theta(\zv_{>k})}{q_\phi(\zv_{> k}|\xv)}]. 
\end{align}

\paragraph{Analysis of its incremental effectiveness on likelihood.}
Following the analysis for it in Section \ref{sec:incremental_effectiveness}, 
we first rewrite the $\log p_\theta(\xv, k)$ to a new formula expressed as
\begin{align}
    \log p_\theta(\xv, k) &=  \mathbb{E}_{p_\theta(\zv_{\leq k}|\zv_{>k})q_\phi(\zv_{>k}|\xv)}\log \frac{p_\theta(\xv|\zv)p_\theta(\zv_{>k})}{q_\phi(\zv_{> k}|\xv)}  \notag \\
    &= \mathbb{E}_{p_\theta(\zv_{\leq k}|\zv_{>k})q_\phi(\zv_{>k}|\xv)}\log p_\theta(\xv|\zv) - D_{\text{KL}}[q_\phi(\zv_{>k}|\xv)||p_\theta(\zv_{>k})].
\end{align}
Similar to our decompose and expectation for $\log p_\theta(\xv)$ on Eq. \ref{eq:expectation_logp},
we have
\begin{align}
    &\mathbb{E}_{\xv\sim p(\xv)}[\mathbb{E}_{\zv\sim p_\theta(\zv_{\leq k}|\zv_{>k})q_\phi(\zv_{>k}|\xv)}\log p_\theta(\xv|\zv)] \\
    =& \mathbb{E}_{p(\xv)p_\theta(\zv_{\leq k}|\zv_{>k})q_\phi(\zv_{>k}|\xv)}[\log \frac{p_\theta(\zv|\xv)}{p(\zv)}p(\xv)] = \hat{\mathcal{I}}_{q,p}(\xv,\zv,k) - \mathcal{H}_p(\xv), \notag
\end{align}
and
\begin{align}
    &\mathbb{E}_{\xv\sim p(\xv)}[D_{\text{KL}}((q_\phi(\zv_{>k})|\xv)||p_\theta(\zv_{>k}))] \\
    =& \mathbb{E}_{p(\xv)q_\phi(\zv_{>k}|\xv)}[\log \frac{q_\phi(\zv_{>k}|\xv)}{q(\zv_{>k})}\frac{q(\zv_{>k})}{p_\theta(\zv_{>k})}] = \hat{\mathcal{I}}_q(\xv, \zv, k) + D_{\text{KL}}(q(\zv_{>k})||p_\theta(\zv_{>k})).
\end{align}
Please note that we use $\hat{\mathcal{I}}_{q,p}(\xv,\zv,k)$ and $\hat{\mathcal{I}}_q(\xv, \zv, k)$ to keep a consistent expression for convenience to draw parallels with Eq. \ref{eq:expectation_logp}, which also remains unchanged after model parameters are fixed that could be fully expressed in details as
\begin{align}
    \hat{\mathcal{I}}_{q,p}(\xv,\zv,k)& = \mathbb{E}_{p(\xv)p_\theta(\zv_{\leq k}|\zv_{>k})q_\phi(\zv_{>k}|\xv)}[\log p_\theta(\zv|\xv)] - \mathbb{E}_{p_\theta(\zv_{\leq k}|\zv_{>k})q(\zv_{>k})}[\log p(\zv)], \\
    &\hat{\mathcal{I}}_q(\xv, \zv, k) = \mathbb{E}_{p(\xv)q_\phi(\zv_{>k}|\xv)}[\log q_\phi(\zv_{>k}|\xv)] - \mathbb{E}_{q_\phi(\zv_{>k})}[q(\zv_{>k})].
\end{align}

Therefore, the expectation of $\log p_\theta(\xv, k)$ on a data distribution $p$ could be expressed as
\begin{align}
    \mathbb{E}_{\xv\sim p}[\log p_\theta(\xv, k)] &= \mathbb{E}_{\xv\sim p}[\mathbb{E}_{p_\theta(\zv_{\leq k}|\zv_{>k})q_\phi(\zv_{>k}|\xv)}\log p_\theta(\xv|\zv)] - \mathbb{E}_{\xv\sim p}[D_{\text{KL}}(q_\phi(\zv_{>k}|\xv)||p_\theta(\zv_{>k}))] \notag \\
    &= -D_{\text{KL}}[q(\zv_{>k})||p(\zv_{>k})] - \text{Ent-Mut}(\theta, \phi, p, k),
\end{align}
where $\text{Ent-Mut}(\theta, \phi, p, k)$ denotes a constant only related to model parameters, data distribution, and layer index $k$, which is expressed as
\begin{align}
    \text{Ent-Mut}(\theta, \phi, p, k) = \mathcal{H}_p(\xv) + \hat{\mathcal{I}}_q(\xv, \zv, k) - \hat{\mathcal{I}}_{q,p}(\xv,\zv,k).
\end{align}

Finally, we could fully expand Eq. \ref{eq:incremental_llr} as 
{\small
\begin{align}
    &\mathcal{A}(\mathcal{LLR}^{>k}) = - [\mathbb{E}_{\xv\sim p_{id}}\log p_\theta(\xv, k) - \mathbb{E}_{\xv\sim p_{ood}}\log p_\theta(\xv, k)]  \label{eq:detail_incremental_llr}\\
    =& [D_{\text{KL}}(q_{id}(\zv_{>k})||p(\zv_{>k})) - D_{\text{KL}}(q_{ood}(\zv_{>k})||p(\zv_{>k})) ] + [\text{Ent-Mut}(\theta, \phi, p_{id}, k) - \text{Ent-Mut}(\theta, \phi, p_{ood}, k)]. \notag
\end{align}
}

Intuitively, when the ID dataset entropy is higher than the OOD dataset, $\mathcal{A}(\mathcal{LLR}^{>k})>0$ could be empirically guaranteed as the term $[\text{Ent-Mut}(\theta, \phi, p_{id}, k) - \text{Ent-Mut}(\theta, \phi, p_{ood}, k)]$ containing an entropy gap term could be empirically greater than 0, and the $D_{\text{KL}}(q_{id}(\zv_{>k})||p(\zv_{>k})) - D_{\text{KL}}(q_{ood}(\zv_{>k})||p(\zv_{>k}))$ term should be relatively small.
Thus, the U-OOD detection performance of log-likelihood ratio methods could achieve better performance than likelihood when the dataset entropy heavily limits its performance, \textit{e.g.}, the FashionMNIST (ID) / MNIST (OOD) and CIFAR10 (ID) / SVHN (OOD). 
However, on the contrary, it is highly possible to have $\mathcal{A}(\mathcal{LLR}^{>k})<0$ if the $D_{\text{KL}}(q_{id}(\zv_{>k})||p(\zv_{>k})) - D_{\text{KL}}(q_{ood}(\zv_{>k})||p(\zv_{>k}))$ cannot compensate the impact of the entropy gap in Ent-Mut terms.
This could empirically explain why the log-likelihood ratio methods could hurt the likelihood's performance in some reverse verification experiments in Table \ref{tab:main_tab}.

Since there is no explicit or empirical guarantee for $\mathcal{A}(\mathcal{LLR}^{>k})>0$ in various cases, our analysis indicates that the log-likelihood methods have no guarantee of incremental effectiveness.

\subsection{Likelihood Regret} 
\paragraph{Method description.} Different from other training-free methods, the likelihood regret \cite{xiao}, denoted as $LRe(\xv)$, needs to retrain certain parts of DGM, \textit{e.g.}, retrain the encoder, parameterized by $\phi$, of the VAE, on every single testing data for computing the score for detection, which could be defined as
\begin{align}
    LRe(\xv) = \log p_\theta(\xv; \phi) - \log p_\theta(\xv; \phi^*(\xv)),
\end{align}
where $\theta$ and $\phi$ are the parameters of a VAE trained on the whole training set $\mathcal{D}_{\text{train}}$ and $\phi^*(\xv)$ denotes the fine-tuned encoder's parameters on the single testing input data $\xv$, \textit{i.e.,}
\begin{align}
    \phi^*(\xv) = \arg\max_\phi \log p_\theta(\xv; \phi).
\end{align}
The intuition is that a data sample with a lower $LRe(\xv)$ could more possibly be an OOD data sample. 
Please kindly note that to make it consistent with other score functions by assigning a higher score for ID data, we actually multiply a "-" to the original definition $LRe(\xv)$ \cite{xiao}.

\paragraph{Analysis of its incremental effectiveness on likelihood.}
Similar to the above previous procedure, we could write its incremental effectiveness as
\begin{align}
    \mathcal{A}(LRe) &= \mathcal{G}(LRe) - \mathcal{G}(\log p_\theta) =  \mathcal{G}(\log p_\theta) - \mathcal{G}(\log p_{\theta,\phi^*}) - \mathcal{G}(\log p_\theta) \notag\\
    &=-\mathbb{E}_{\xv\sim p_{id}}\log p_\theta(\xv; \phi^*(\xv)) + \mathbb{E}_{\xv\sim p_{ood}}\log p_\theta(\xv; \phi^*(\xv)).
\end{align}
It could be very interesting to find that when the DGM exhibits a paradox when it assigns a higher likelihood to the OOD data than ID data, \textit{i.e.}, the "hard benchmarks", $\log p_\theta(\xv_{ood}; \phi^*(\xv))$ would be further maximized that makes the $\log p_\theta(\xv_{id}; \phi^*(\xv))$ harder to be greater than it with several iterations (typically a small number considering the computational efficiency) of fine-tuning, leading to LRe's incremental effectiveness $\mathcal{A}(LRe)>0$ in the "hard benchmarks".
However, when the DGM could already give a lower likelihood to OOD data and the decoder trained on the ID training set remains unchanged during fine-tuning, the $\mathcal{A}(LRe)$ can still be samller than 0. Therefore, the LR method still cannot guarantee incremental effectiveness on likelihood.

\subsection{Likelihood Ratio} 
\paragraph{Method description.} Likelihood Ratio (LRa) \cite{likelihood-ratio-jie} method introduces a background model parameterized by $\theta_b$ for capturing general background statistics, denoted as $\log p_{\theta_b}(\xv)$ and proposes using the following score function for U-OOD detection:
\begin{align}
    LRa(\xv) = \log p_\theta(\xv) - \log p_{\theta_b}(\xv).
\end{align}
Intuitively, $LRa(\xv)$ removes the background information in $\log p_\theta(\xv)$ and focuses more on the semantic foreground information, hoping to assign higher scores for data samples sharing similar semantic information with the ID training data.

\paragraph{Analysis of its incremental effectiveness on likelihood.} We write its advantage function $\mathcal{A}$ as follows:
\begin{align}
    \mathcal{A}(LRa) &= \mathcal{G}(LRa) - \mathcal{G}(\log p_\theta) = \mathcal{G}(\log p_\theta) - \mathcal{G}(\log p_{\theta_b}) - \mathcal{G}(\log p_\theta) \notag \\
    &= \mathbb{E}_{\xv\sim p_{ood}} \log p_{\theta_b}(\xv) - \mathbb{E}_{\xv\sim p_{id}} \log p_{\theta_b}(\xv).
\end{align}
If the DGM's likelihood's performance is limited due to that $\mathbb{E}_{\xv\sim p_{ood}}\log p_{\theta_b}(\xv)$ is overestimated all the time, \textit{i.e.}, higher than that of the ID data, the $\mathcal{A}(LRa) > 0$ could be satisfied. However, though it could explain its good performance in the "hard benchmarks" such as CIFAR-10 (ID) / SVHN (OOD), it may hurt the performance when switching the ID/OOD setting, \textit{i.e.,} $\mathcal{A}(LRa) < 0$ in SVHN (ID) / CIFAR (OOD).
Therefore, incremental effectiveness on likelihood is not explicitly guaranteed in this likelihood ratio method.

\subsection{Input Complexity} 
\paragraph{Method description.} The Input Complexity (IC) method \cite{input_complexity} is similar to our DEC method without penalization on the data with higher complexity than ID data. Formally, it introduces a compressor to calculate the compressed bits per dimension $L(\xv)$ and uses the following score function for U-OOD detection:
\begin{align}
    IC(\xv) = \log p_\theta(\xv) + L(\xv),
\end{align}
where a higher $IC(\xv)$ indicates the data sample $\xv$ has higher probability of being OOD data.

\paragraph{Analysis of its incremental effectiveness on likelihood.} The incremental effectiveness of it could be expressed as
\begin{align}
    \mathcal{A}(IC) &= \mathcal{G}(IC) - \mathcal{G}(\log p_\theta) = \mathcal{G}(\log p_\theta) + \mathcal{G}(L) - \mathcal{G}(\log p_\theta) \notag \\
    &= \mathbb{E}_{\xv\sim p_{id}} [L(\xv)] - \mathbb{E}_{\xv\sim p_{ood}} [L(\xv)].
\end{align}
In fact, as we discussed in Section \ref{sec:dec}, though this could potentially cancel out the impact of dataset entropy, it obviously cannot guarantee incremental effectiveness when ID data has a lower complexity than OOD data, \textit{i.e.}, $\mathbb{E}_{\xv\sim p_{id}} [L(\xv)] < \mathbb{E}_{\xv\sim p_{ood}} [L(\xv)]$ leads to $\mathcal{A}(IC)<0$. This observation inspired us to add penalization on the data that have higher complexity than the average complexity of the ID training set.

\subsection{WAIC} 
\paragraph{Method description.} WAIC \cite{choi2018waic} is a representative ensemble method that simply
but effectively uses the following statistic as a score function for U-OOD detection:
\begin{align}
    \text{WAIC}(\xv) = \mathbb{E}_{\theta}[\log p_\theta(\xv)] - \text{Var}_\theta[\log p_\theta(\xv)].
\end{align}
Intuitively, the unseen OOD data would get a large variance $\text{Var}_\theta[\log p_\theta(\xv)]$ when using an ensemble of DGMs' likelihood to assess it, leading to a lower score $\text{WAIC}(\xv)$.

\paragraph{Analysis of its incremental effectiveness on likelihood.} Different from other analyses, we need to treat the $\theta$ as a variable instead of a fixed vector and we rewrite $\log p_\theta$ of a random single DGM in that ensemble as $\log p_{{\theta}^*}$. Then, we could start to analyze WAIC's incremental effectiveness on likelihood, expressed as
\begin{align}
    &\mathcal{A}(\text{WAIC}) = \mathcal{G}(\text{WAIC}) - \mathcal{G}(\log p_{{\theta}^*}) \\
    =&\mathbb{E}_{\xv\sim p_{id}}[\mathbb{E}_{\theta}[\log p_\theta(\xv)] - \log p_{{\theta}^*}(\xv) - \text{Var}_\theta[\log p_\theta(\xv)]] \notag \\
    &\quad\quad\quad\quad\quad\quad\quad\quad\quad\quad- \mathbb{E}_{\xv\sim p_{ood}}[\mathbb{E}_{\theta}[\log p_\theta(\xv)] - \log p_{{\theta}^*}(\xv) - \text{Var}_\theta[\log p_\theta(\xv)]] \notag\\
    =& \mathbb{E}_{\xv\sim p_{ood}}[\text{Var}_\theta [\log p_\theta(\xv)]] - \mathbb{E}_{\xv\sim p_{id}}[\text{Var}_\theta [\log p_\theta(\xv)]]  \notag \\
    &+\mathbb{E}_{\xv\sim p_{id}}[\mathbb{E}_{\theta}[\log p_\theta(\xv)] - \log p_{{\theta}^*}(\xv)] -\mathbb{E}_{\xv\sim p_{ood}}[\mathbb{E}_{\theta}[\log p_\theta(\xv)] - \log p_{{\theta}^*}(\xv)]. 
\end{align}
For the first term, $\mathbb{E}_{\xv\sim p_{ood}}[\text{Var}_\theta [\log p_\theta(\xv)]] - \mathbb{E}_{\xv\sim p_{id}}[\text{Var}_\theta [\log p_\theta(\xv)]]>0$ is reasonable as the DGM would achieve more consistency on the training set than the unseen OOD data, though it is hard to theoretically prove.
For the remaining terms, if we could make an ensemble of infinite or a large number of DGMs and we let $\theta^*$ as a DGM that satisfies $\mathbb{E}_{\theta}[\log p_\theta(\xv)] = \log p_{{\theta}^*}(\xv)$, then the WAIC methods would empirically guarantee $\mathcal{A}(\text{WAIC})>0$.
However, considering the computation resource and efficiency, we can only ensemble a small number of DGMs like 5 randomly initialized DGMs.
As we have no prior knowledge of the performance of these randomly initialized DGMs in the ensemble, we have no guarantee of the relationship between $\mathbb{E}_{\theta}[\log p_\theta(\xv)]$ and $\log p_{{\theta}^*}(\xv)$.

\section{Jusitification on the Latent Distribution Mismatch} \label{app:sec_justify_latent}

\begin{figure*}[ht!]
    \centering
    \subfigure[Data distribution]{
    \centering
    \includegraphics[width=0.20\textwidth]{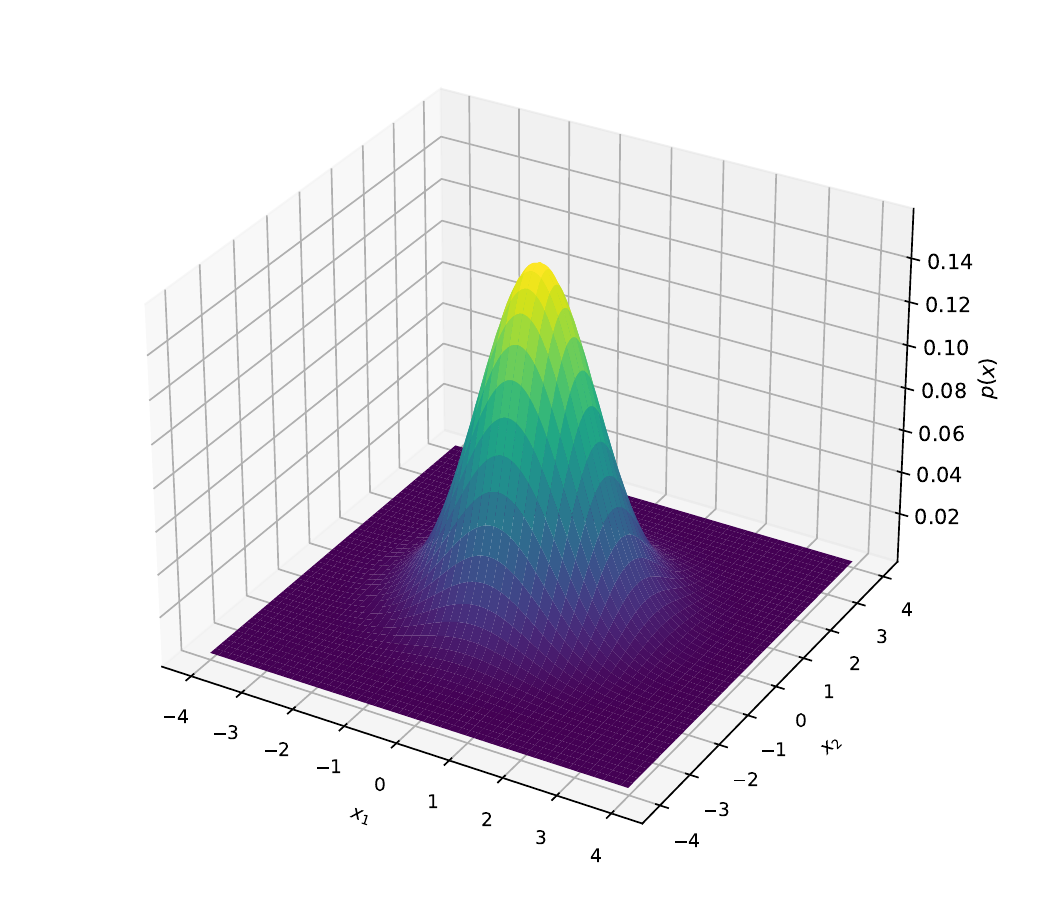}
    \label{fig:single-mode-px}}\ \ \ \ 
    \subfigure[Prior $p(\zv)$]{
    \centering
    \includegraphics[width=0.20\textwidth]{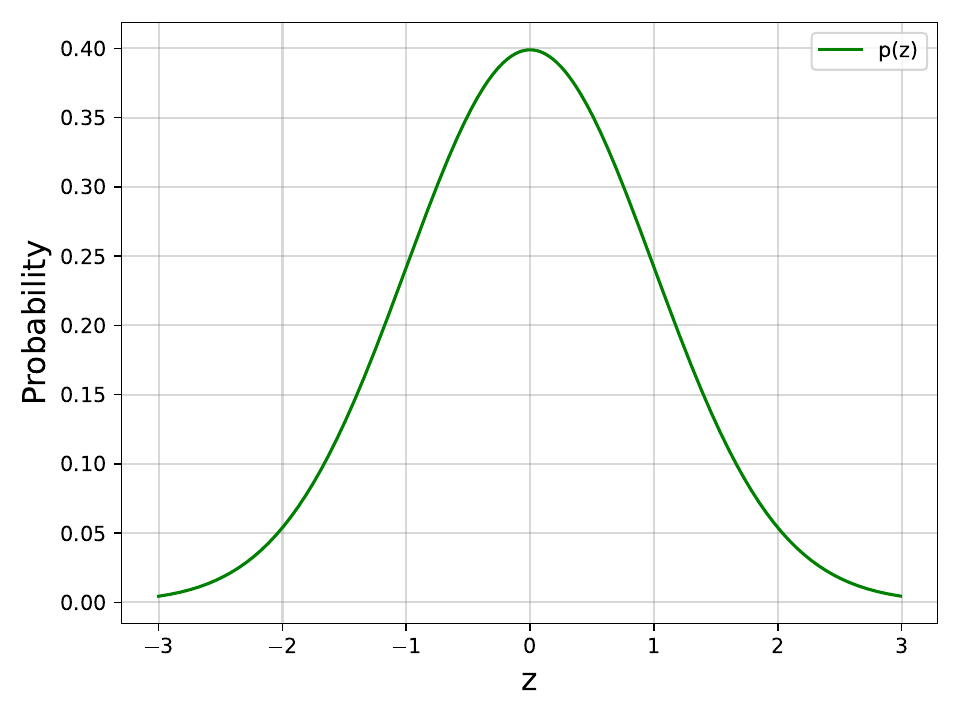}
    \label{fig:pz}}\ \ \ \ 
    \subfigure[Estimated {$p_\theta(\xv)$}]{
    \centering
    \includegraphics[width=0.20\textwidth]{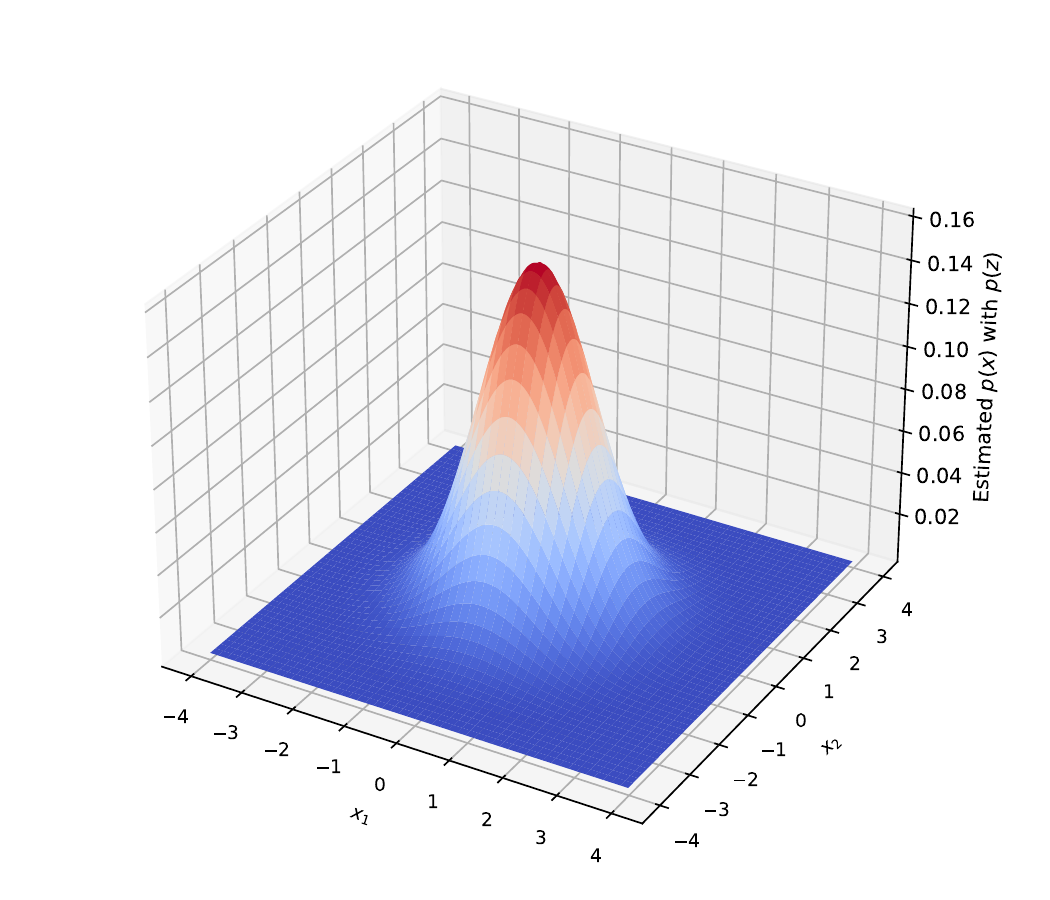}
    \label{fig:single-mode-ELBO}}\ \ \ \ 
    \subfigure[Posterior $q(\zv)$]{
    \centering
    \includegraphics[width=0.20\textwidth]{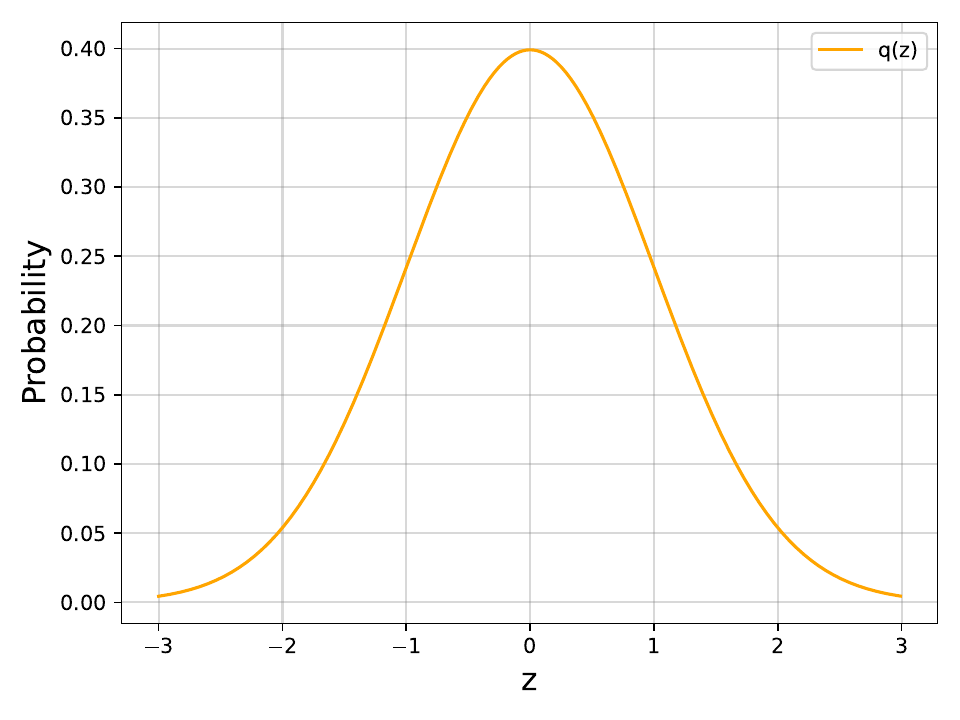}
    \label{fig:single-mode-qz}}\ \ \ \ 
    \subfigure[Data distribution]{
    \centering
    \includegraphics[width=0.20\textwidth]{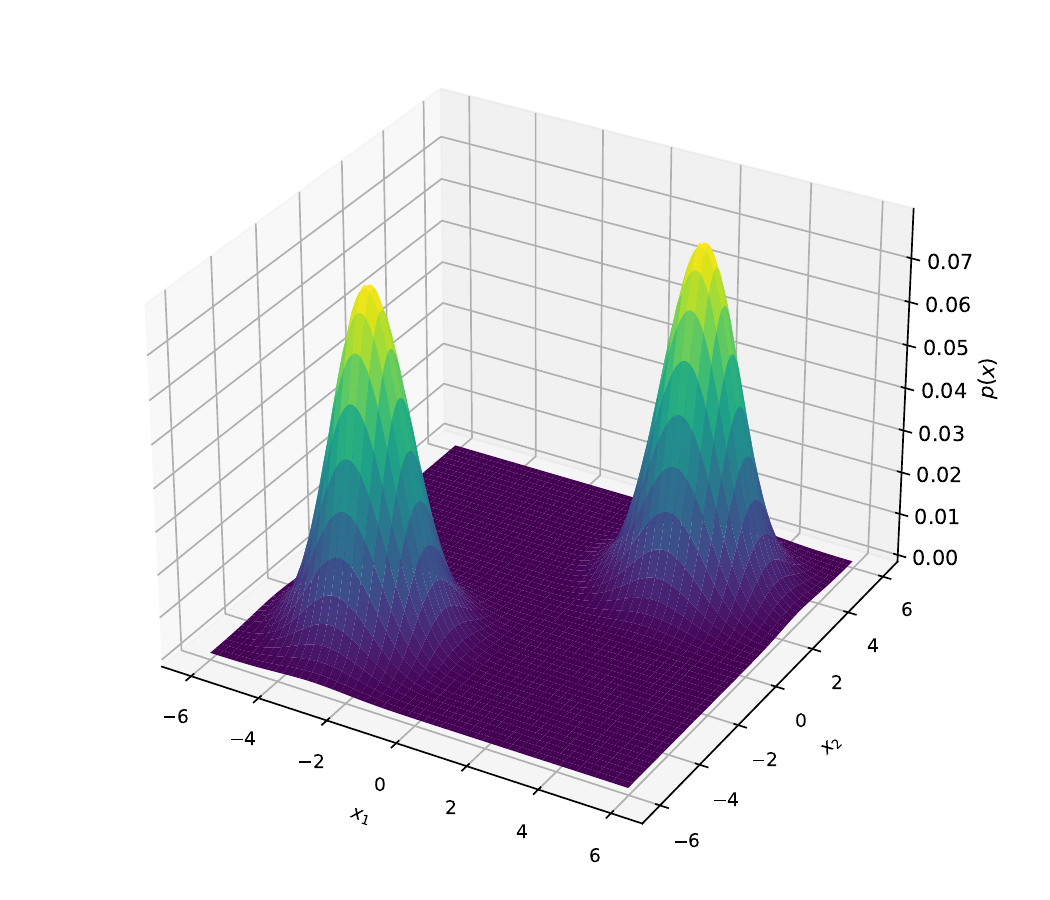}
    \label{fig:multi-mode-px}}\ \ \ \ 
    \subfigure[$p(\zv)$\&$q(\zv)$]{
    \centering
    \includegraphics[width=0.20\textwidth]{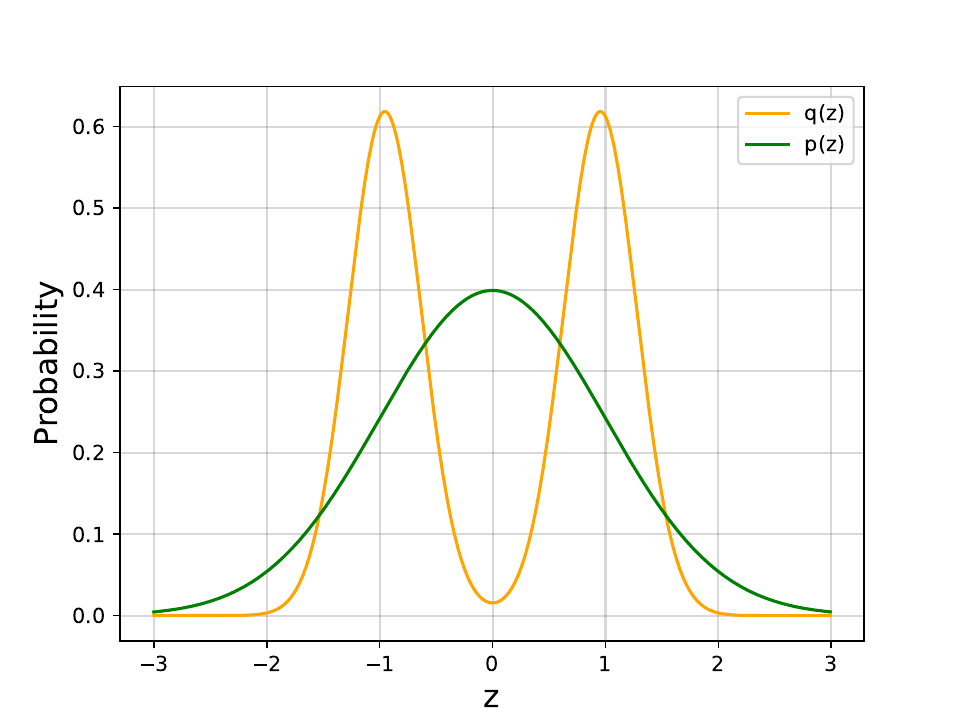}
    \label{fig:pz_qz}
    }\ \ \ \ 
    \subfigure[$\hat{p}_{\theta}(\xv)$ with $p(\zv)$]{
    \centering
    \includegraphics[width=0.20\textwidth]{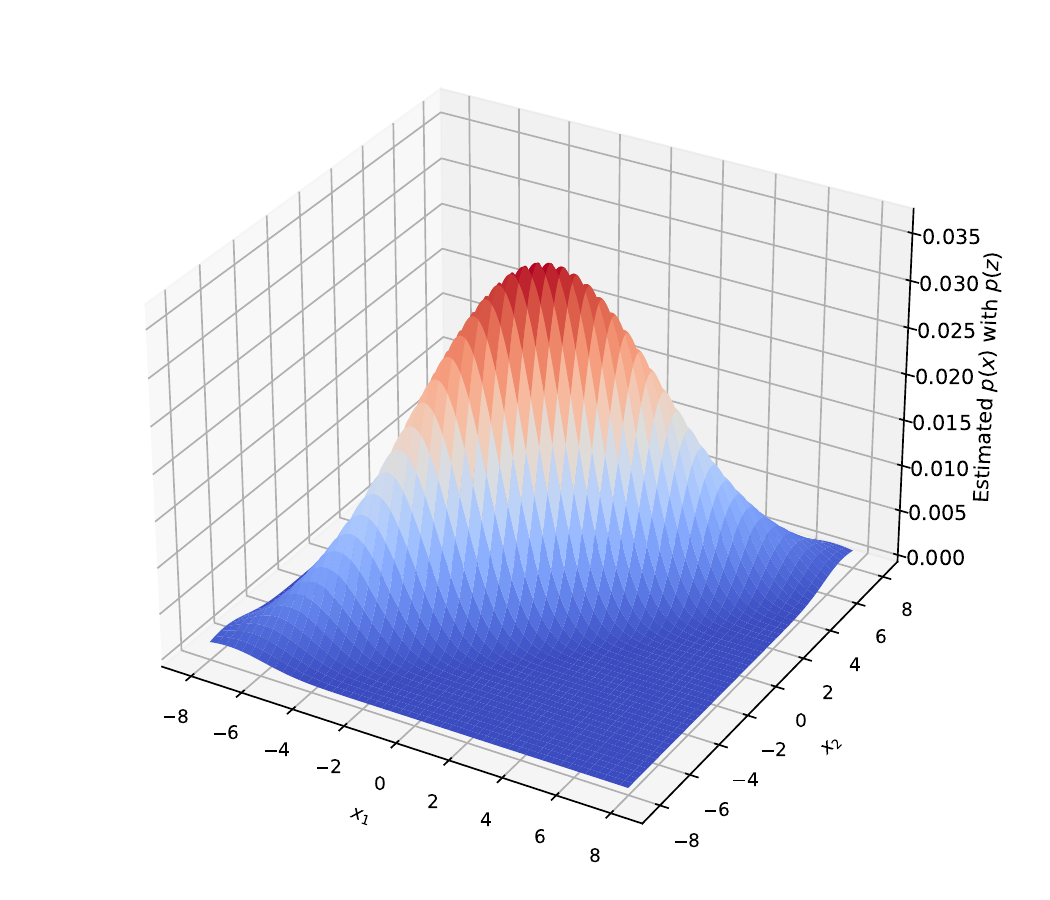}
    \label{fig:multi-mode-ELBO-pz}}\ \ \ \ 
    \subfigure[$\hat{p}_{\theta}(\xv)$ with $q(\zv)$]{
    \centering
    \includegraphics[width=0.20\textwidth]{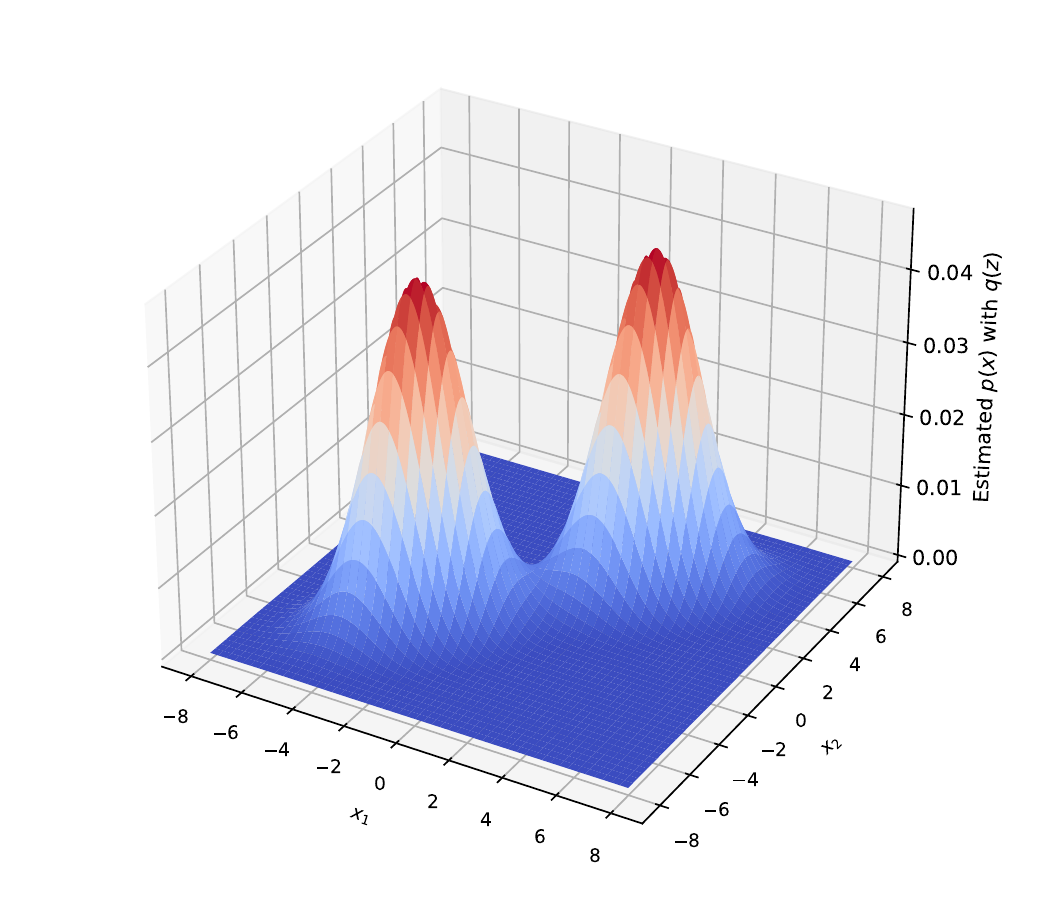}
    \label{fig:multi-mode-ELBO-qz}}\ \ \ \ 
    \caption{
    {(\textbf{a-d}): Visualization of modeling a \textbf{single-modal} 
 data distribution with a linear VAE; (\textbf{e-h}): Visualization of modeling a \textbf{multi-modal} 
 data distribution with a linear VAE.}}
    \label{fig:single_multi_case}
\end{figure*}

{Since the direction I, \textit{i.e.}, latent distribution mismatch, may be challenging to comprehend, we provide a further analysis in this section.} \textbf{For ease of reading, we present the conclusion first, followed by the detailed setup and derivation.}

\textbf{{When the design of prior is proper?}}
\label{sec:single-mode}
Assuming a dataset $\{\xv_i\}_{i=1}^{N}$ sampled \textit{i.i.d.} from  $p(\xv)=\mathcal{N}(\xv|\bf{0}, \Sigmav_x)$ as shown in Fig.~\ref{fig:single-mode-px}, 
and we construct a linear VAE to estimate $p(\xv)$, formulated as: 
\begin{align}
    p(\zv)&=\mathcal{N}(\zv|\mathbf{0}, \Iv), 
    q_{\phi}(\zv|\xv)     \label{eq:formulation} \\ \notag
    &=\mathcal{N}(\zv|\Av\xv+\Bv, \Cv),
    p_{\theta}(\xv|\zv) \\ \notag
    &=\mathcal{N}(\xv|\Ev\zv+\Fv, \sigma^2\Iv),
\end{align}
where all learnable parameters' optimal values can be obtained by the derivation in Appendix \ref{sec:appen_derivation_kl_single}.
As depicted in Fig.~\ref{fig:single-mode-ELBO},  we find that the linear VAE can accurately estimate the $p(\xv)$. Figs.~\ref{fig:pz} and \ref{fig:single-mode-qz} indicate that the design of the prior distribution is proper, where $q(\zv)$ equals $p(\zv)$.

\textbf{{When the design of prior is NOT proper?}}
Consider a more complex data distribution, \textit{e.g.}, {\rebuttal a mixture of Gaussians} as shown in Fig.~\ref{fig:multi-mode-px} (More details are in Appendix \ref{sec:appen_toy_details}), we could also get the optimal parameters of the same linear VAE in Eq.~\ref{eq:formulation}. 
After the derivation in Appendix  \ref{sec:appen_derivation_kl_multi}, Fig.~\ref{fig:pz_qz} illustrates that $q(\zv)$ is a multi-modal distribution instead of $p(\zv)=\mathcal{N}(\zv|\bf{0}, \Iv)$, \textit{i.e.}, the design of the prior is not proper, which leads to paradox as seen in Fig.~\ref{fig:multi-mode-ELBO-pz}. However, as analyzed in Direction I, we find that the paradox is mitigated when replacing $p(\zv)$ with $q(\zv)$ in the KL term of the ELBO, as shown in Fig.~\ref{fig:multi-mode-ELBO-qz}.

\textbf{More empirical studies {on non-linear VAEs} for the improper design of prior.}
For more practical cases,
we use non-linear deep VAEs to model $q_\phi(\zv|\xv)$ and $p_\theta(\xv|\zv)$ with $p(\zv)=\mathcal{N}(\bf{0}, \Iv)$ on the same multi-modal dataset in Fig.~\ref{fig:multi-mode-px} and image datasets. Implementation details are in \ref{sec:appen_imple_deep_vae}. For the low-dimensional multi-modal dataset, we observed that $q(\zv)$ still differs from $p(\zv)$, as shown in Fig.~\ref{fig:deep-vae-qz}. The likelihood still suffers from the paradox, especially in the region near $(0,0)$, as shown in Fig.~\ref{fig:deep-vae-px}. For the image datasets, please note that, if $q_{{id}}(\zv)$ is closer to $p(\zv)=\mathcal{N}(\bf{0}, \Iv)$, $\zv_{{id}} \sim q_{{id}}(\zv)$ should occupy the center of latent space $\mathcal{N}(\bf{0}, \Iv)$ and $\zv_{{id}} \sim q_{{id}}(\zv)$ should be pushed far from the center, leading to $p(\zv_{{id}})$ {to be larger} than $p(\zv_{{id}})$. Surprisingly, we find this expected phenomenon does not exist, as shown in Fig.~\ref{fig:p01-fmnist} and \ref{fig:p01-cifar}, where the experiments are on two dataset pairs, Fashion-MNIST(ID)/MNIST(OOD) and CIFAR-10(ID)/SVHN(OOD). This still suggests that 
the prior $p(\zv)$ is improper, even $q_{{id}}(\zv)$ for OOD data may be closer to $p(\zv)$ than $q_{{id}}(\zv)$. More ablation studies including the dataset size and model architecture could be seen in \ref{sec_app_aba_deep_vae}.
\begin{figure}[h!]
    \centering
    \subfigure[$q_{{id}}(\zv)$]{\includegraphics[width=0.21\textwidth]{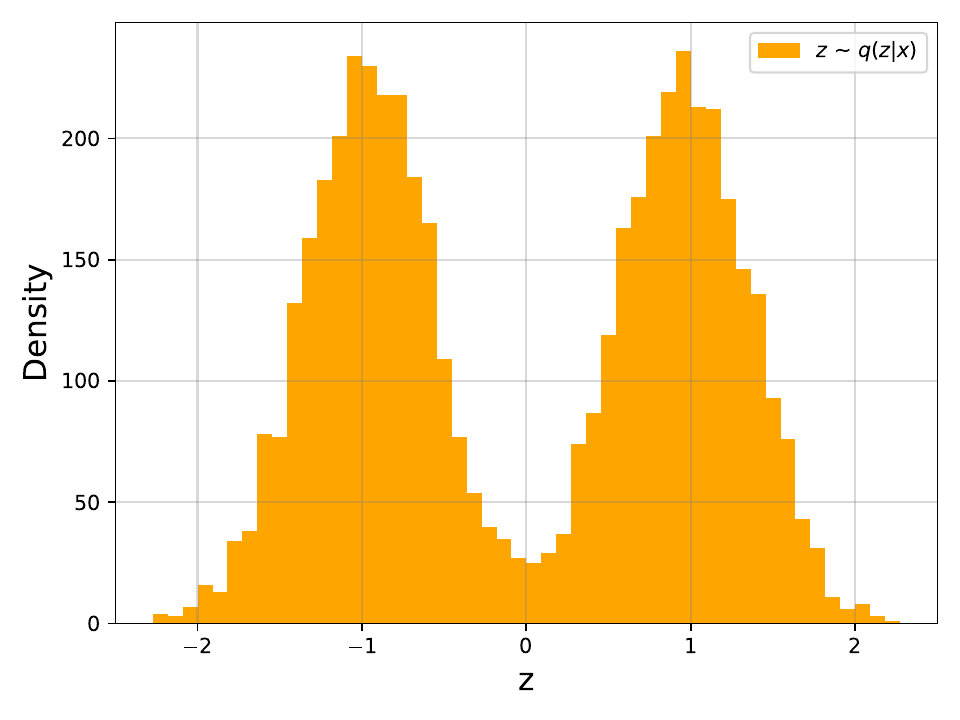}
    \label{fig:deep-vae-qz}}
    \subfigure[Estimated $p_\theta(\xv)$]{\includegraphics[width=0.21\textwidth]{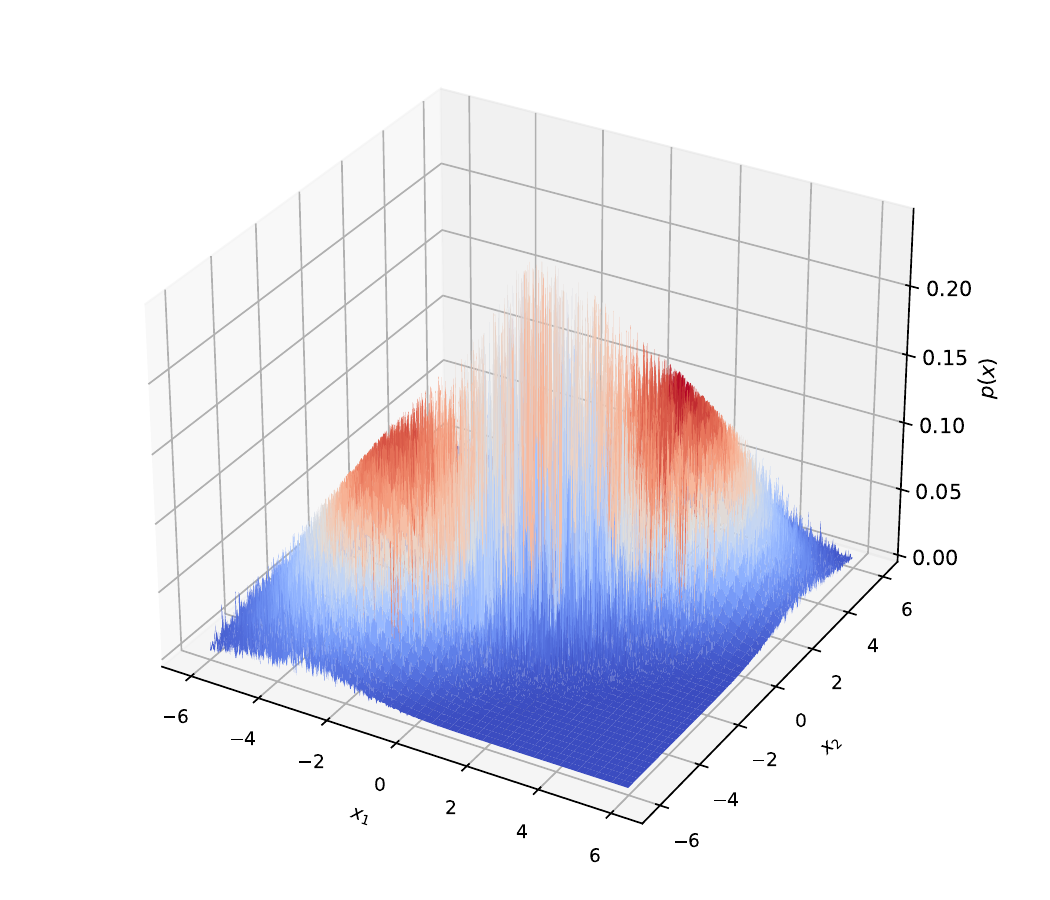}
    \label{fig:deep-vae-px}}
    \subfigure[FashionMNIST (ID)]{\includegraphics[width=0.21\textwidth]{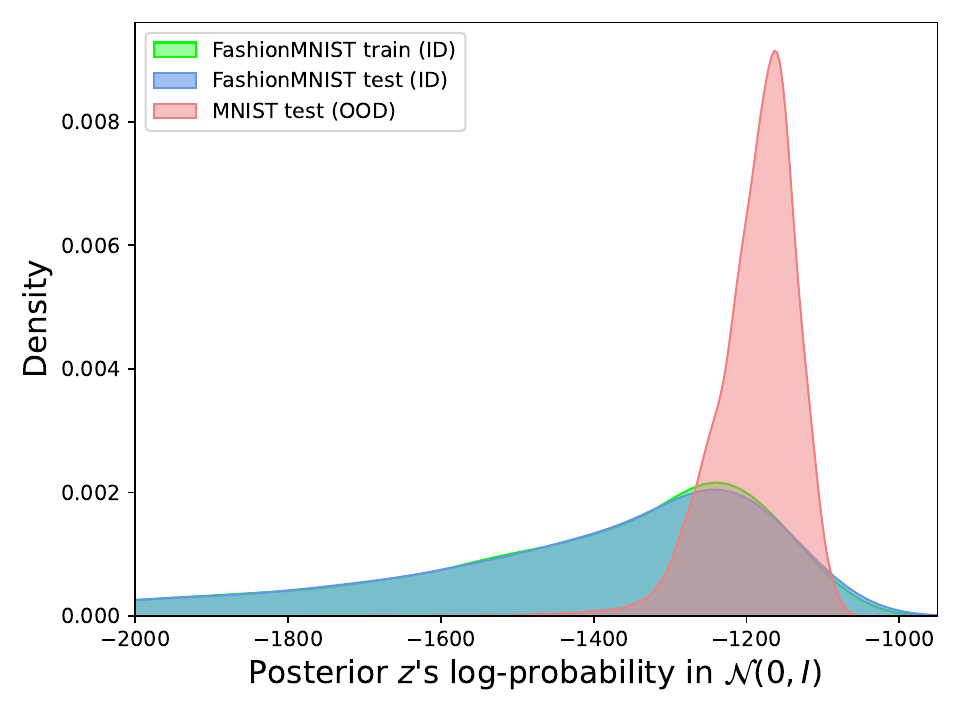}
    \label{fig:p01-fmnist}}
    \subfigure[CIFAR-10 (ID)]{\includegraphics[width=0.21\textwidth]{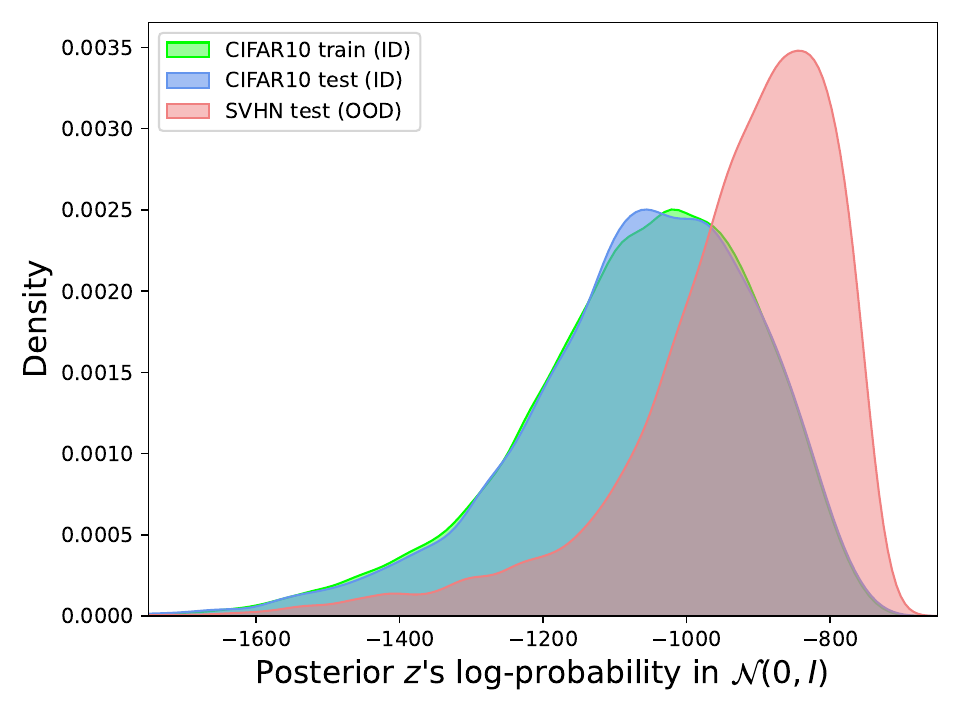}
    \label{fig:p01-cifar}}
    \vspace{-3mm}
    \caption{
    {\textbf{(a)} and \textbf{(b)}: Visualization of $q_{{id}}(\zv)$ and estimated $p_{\theta}(\xv)$ by $\text{ELBO}$ on the multi-modal 
 data distribution with a non-linear deep VAE;} \textbf{(c)} and \textbf{(d)}: Density plots of  the log-probability of posterior $\zv$ sampled from $q_{id/ood}(\zv)$ in prior $\mathcal{N}(\bf{0}, \Iv)$ on two dataset pairs. }
    \label{fig:multi_case}
\end{figure}

{\textbf{Brief summary.}} Through analyzing paradox scenarios from simple to complex, we could summarize as follows: {the prior distribution $p(\mathbf{\zv})=\mathcal{N}(\mathbf{0}, \mathbf{I})$ {\rebuttal may be} an improper choice for variational DGMs when modeling {a} complex data distribution $p({\xv})$}, leading to an overestimated $D_{\text{KL}}(q_{{id}}(\zv)||p(\zv))$ and further limiting the performance of likelihood in U-OOD detection.

\subsection{Toy Examples' Details} \label{sec:appen_toy_details}
\textbf{Single-modal case setup.} In this scenario, the data distribution is determined by a standard 2-dimensional Gaussian distribution $p(\xv) = \mathcal{N}(\xv| \bf{0}, \Sigmav_{\xv})$, where
\begin{align}
\Sigmav_{\xv} = \begin{bmatrix}
1 & 0 \\
0 & 1
\end{bmatrix}.
\end{align}

In order to simulate the dimension-reduction property of VAE, we designate the dimension of the latent variable as 1-dimensional; that is, the variance {$\Iv$ in $p(\zv)$ reduces to $1$}.
Under this configuration, we $i.i.d.$ sample $N=5000$ data points from the data distribution $p(\xv)$ to construct a training set. Each parameter's solutions are calculated analytically.

\textbf{Multi-modal case setup.} The data distribution is made by a mixture of two standard single-modal Gaussian distributions, \textit{i.e.},  $p(\xv)=\sum_{k=1}^{K}\pi_k \mathcal{N}(\xv|\muv_k, \Sigmav_k)$, where $K=2$, $\pi_k=1/2$ and  
\begin{align}
    \muv_1 = 
    \begin{bmatrix}
        3\\
        3
    \end{bmatrix}, 
    \muv_2 = 
    \begin{bmatrix}
        -3\\
        -3
    \end{bmatrix}, 
    \Sigmav_1 = 
    \begin{bmatrix}
        1 & 0 \\
        0 & 1
    \end{bmatrix}, 
    \Sigmav_2 = 
    \begin{bmatrix}
        1 & 0 \\
        0 & 1
    \end{bmatrix}.
\end{align}
The training set of this multi-modal case is built by $i.i.d.$ sampling from 5000 data points from each component Gaussian distribution $\mathcal{N}(\xv|\muv_k, \Sigmav_k)$, \textit{i.e.}, 10000 data points in total.

\subsection{Derivation for Single-modal Case}
\label{sec:appen_derivation_kl_single}
Assume we have a dataset containing $N$ data samples $\{\xv_1,\xv_2,...,\xv_N\}, x_i\in\mathbb{R}^d$, $d=2$, and we {already} know the groundtruth distribution of it, \textit{i.e.}, 
\begin{equation}
    p(\xv)=\mathcal{N}(\xv|\bf{0}, \Sigmav_x),
\end{equation}
where $\Sigmav_x = \Iv$.
We have a linear VAE model parameterized as:
\begin{align}
    p(\zv)&=\mathcal{N}(\zv|\bf{0}, \Iv) \\
    q_{\phi}(\zv|\xv)&=\mathcal{N}(\zv|\Av\xv+\Bv, \Cv)\\
    p_{\theta}(\xv|\zv)&=\mathcal{N}(\xv|\Ev\zv+\Fv, \sigma^2\Iv),
\end{align}
where $p(\zv)$ is the prior distribution, $\zv\in\mathbb{R}^q, q=1$, $q_\phi(\zv|\xv)$ is the approximated posterior distribution, and $p_\theta(\xv|\zv)$ is the approximated likelihood distribution. Directly employing the knowledge from probabilistic Principal Component Analysis (pPCA) \cite{ppca}, we could get the maximum likelihood estimation of $p_\theta(\xv|\zv)$:
\begin{align}
    \sigma^2_{\text{MLE}} &= \frac{1}{d-q}\sum_{j=q+1}^d \lambda_j \\ 
    \Ev_{\text{MLE}} &= \mathbf{U}_q\left(\boldsymbol{\Lambda}_q-\sigma^2_{\text{MLE}}\right)^{1 / 2} \mathbf{R}\\
    \Fv_{\text{MLE}} &= \bf{0} 
\end{align}
where $\lambda_{q+1},...,\lambda_d$ are the smallest eigenvalues of the sample covariance matrix $\Sv=\frac{1}{N}\sum_{n=1}^N \xv\xv^\top$, the $d\times q$ orthogonal matrix $\mathbf{U}_q$ 
is made by the $q$ dominant eigenvectors of $\Sv$, the diagonal matrix $\Lambda_q$ contains the corresponding $q$ largest eigenvalues, and $\Rv$ is an arbitary $q\times q$ orthogonal matrix. Note that, when $q=1$, we have $\Rv=\Iv$. 
After we get the parameters of $p_\theta(\xv|\zv)$, we could get the $p(\zv|\xv)$ by Bayes rule:
\begin{equation}
\begin{aligned}
    p(\zv|\xv) &=\frac{p_\theta(\xv|\zv)p(\zv)}{p(\xv)} \\
               &=\mathcal{N}(\zv|\Sigmav_x^{-1}\Ev_{\text{MLE}}^\top \xv,  \sigma_{\text{MLE}}^2\Sigmav_x^{-1}),
\end{aligned}
\end{equation}
where $\Sigmav_x=\Ev_{\text{MLE}}^\top\Ev_{\text{MLE}}+\sigma_{\text{MLE}}^2\Iv$. Thus, the maximum likelihood estimates of $q_\phi(\zv|\xv)$'s parameters are:
\begin{align}
    \Av_{\text{MLE}} &= \Sigmav_x^{-1}\Ev_{\text{MLE}}^{\top}\\
    \Bv_{\text{MLE}} &= \bf{0} \\
    \Cv_{\text{MLE}} &= \sigma_{\text{MLE}}^2\Sigmav_x^{-1}.
\end{align}
Although the maximum likelihood estimations are ascertained, it remains necessary to verify whether these estimations allow the ELBO to reach the global optimum. The derivation of ELBO is as follows:
\begin{equation}
\begin{aligned}
    \log p(\xv) &= \mathbb{E}_{q_\phi(\zv|\xv)}[\log p(\xv|\zv)] - D_{\text{KL}}(q_\phi(\zv|\xv)||p(\zv)) + D_{\text{KL}}(q_\phi(\zv|\xv)||p(\zv|\xv)) \\
    &=\text{ELBO}(\xv) +D_{\text{KL}}(q_\phi(\zv|\xv)||p(\zv|\xv)). 
    \label{eq:log_px}
\end{aligned}
\end{equation}
Given that $q_\phi(\zv|\xv)=\mathcal{N}(\zv|\Sigmav_{\xv}^{-1}\Ev_{\text{MLE}}^{\top}\xv, \sigma_{\text{MLE}}^2\Sigmav_{\xv}^{-1})= p(\zv|\xv)$, $D_{\text{KL}}(q_\phi(\zv|\xv)||p(\xv|\zv))$ becomes zero. Furthermore, any modifications to the parameters of $q_\phi$ would result in an increase of $D_{\text{KL}}(q_\phi(\zv|\xv)||p(\xv|\zv))$; in other words, it would result in a decrease of ELBO. Hence, the global optimum of the ELBO is attained when $\Av_{\text{MLE}}\sim \Ev_{\text{MLE}}, \sigma_{\text{MLE}}$ are implemented in the linear VAE. Moreover, in this situation, $\log p(\xv)$ equates to $\text{ELBO}$.

Finally, we could get the expression of the aggregated posterior distribution $q(\zv)$:
\begin{equation}
\begin{aligned}
    q(\zv) &= \int_{\xv} q_\phi(\zv|\xv)p(\xv) \\
    &=\int_{\xv} \mathcal{N}(\zv|\Sigmav_{\xv}^{-1}\Ev_{\text{MLE}}^{\top}\xv, \sigma_{\text{MLE}}^2\Sigmav_{\xv}^{-1}) \mathcal{N}(\xv|\bf{0}, \Sigmav_{\xv})  \\
    &=\int_{\xv} \mathcal{N}(\zv|\Iv^{-1}\Ev_{\text{MLE}}^{\top}\xv, \sigma_{\text{MLE}}^2\Iv^{-1}) \mathcal{N}(\xv|\bf{0}, \Iv) \\
    &=\int_{\xv} \mathcal{N}(\zv|\Ev_{\text{MLE}}^{\top}\xv, \sigma_{\text{MLE}}^2\Iv) \mathcal{N}(\xv|\bf{0}, \Iv)  \\
    &= \mathcal{N}(\bf{0}, \Ev_{\text{MLE}}^{\top}\Ev_{\text{MLE}}+\sigma_{\text{MLE}}^2\Iv) \\
    &= \mathcal{N}(\bf{0}, \Sigmav_{\xv}) \\
    &= \mathcal{N}(\bf{0}, \Iv) \\
    &= p(\zv).
\end{aligned}
\end{equation}

In summing up the single-modal case, our assertion is that $D_{\text{KL}}[q(\zv)||p(\zv)]=0$, indicating that the design of the prior distribution is appropriate and would not result in an \emph{overestimation} of VAE.

\subsection{Derivation for Multi-modal Case}
\label{sec:appen_derivation_kl_multi}
Assume we have a distribution $p(\xv)=\sum_{k=1}^K\pi_k\mathcal{N}(\xv|\muv_k, \Sigmav_k)$ and we build a dataset containing $K\times N$ data samples, which is made by sampling $N$ data samples from each $\mathcal{N}(\xv|\muv_k, \Sigmav_k)$. The parameterization setting of the $p(\zv)$, $q_\phi(\zv|\xv)$, and $p_\theta(\xv|\zv)$ is the same as the single-modal case.

{Deriving}
from the single-modal scenario, an analytical formulation of $D_{\text{KL}}(q_\phi(\zv|\xv)||p(\zv|\xv))$ is unattainable in the multi-modal case. Thus, it necessitates a derivation directly from the ELBO.
Due to the fact that the global optimum of the decoder's parameters in the ELBO coincides with the global maximum of the marginal likelihood of the observed data \cite{lucas2019don}, we firstly commence with the derivation of the maximum likelihood estimation of $p_\theta(\xv|\zv)$. Despite the feasibility of directly obtaining the maximum likelihood estimation of 
{the parameters in}
$p_\theta(\xv|\zv)$ by optimizing the integration $\hat{p}_\theta (\xv)= \int_{\zv} p_\theta(\xv|\zv)p(\zv)$ using {the} observed data, we propose an additional clarification connecting this integration and the estimated likelihood $\log p_\theta(\xv)$. For a clear notation, we term the estimated likelihood $\log p_\theta(\xv)$ as $\text{ELBO}(\xv)$ here.
With reference to the strictly tighter importance sampling on the $\text{ELBO}$ \cite{importance}, we can derive that
\begin{align}
    \text{ELBO}^{s}(\xv) &= \mathbb{E}_{q_\phi(\zv|\xv)}[\log \frac{1}{S}\sum_{s=1}^{S}\frac{p_\theta(\xv|\zv^{(s)})p(\zv^{(s)})}{q_\phi(\zv^{(s)}|\xv)}].
\end{align}
{Setting the number of instances}
$S=1$, $\text{ELBO}^{s}(\xv)$ equates to the regular $\text{ELBO}(\xv)$. As $S$ approaches $+\infty$, it follows that
\begin{equation}
\begin{aligned}
    \text{ELBO}^{s}(\xv) &= \mathbb{E}_{q_\phi(\zv|\xv)}[\log \mathbb{E}_{q_\phi(\zv|\xv)}\frac{p_\theta(\xv|\zv)p(\zv)}{q_\phi(\zv|\xv)}] \\
    &= \mathbb{E}_{q_\phi(\zv|\xv)}[\log \int_{\zv} {q_\phi(\zv|\xv)}\frac{p_\theta(\xv|\zv)p(\zv)}{q_\phi(\zv|\xv)}d\zv] \\
    &= \mathbb{E}_{q_\phi(\zv|\xv)}[\log \int_{\zv} {p_\theta(\xv|\zv)p(\zv)}d\zv] \\
    &=\log \int_{\zv} {p_\theta(\xv|\zv)p(\zv)}d\zv \\
    &=\log \hat{p}_\theta(\xv).
\end{aligned}
\end{equation}

{The expression of $\hat{p}_\theta(\xv)$}
is shown as:
\begin{equation}
\begin{aligned}
    \hat{p}_\theta(\xv) &= \int_{\zv} p_\theta(\xv|\zv)p(\zv) \\
                 &= \int_{\zv} \mathcal{N}(\xv|\bf{E}\zv+\bf{F}, \sigma^{2}\bf{I})\mathcal{N}(\zv|\bf{0},\bf{I})\\
                 &=\mathcal{N}(\xv|\Fv, \Ev\Ev^\top+\sigma^2\Iv).
\end{aligned}
\end{equation}

Then, the joint log-likelihood of the observed dataset $\{\xv^{(k)}_i\}_{i=1,k=1}^{N, K}$ can be formulated as: 
\begin{align}
        \mathcal{L} = {\sum_{k=1}^{K} \sum_{i=1}^{N}} \log \hat{p}_{{\theta}}(\xv_i^{(k)})  
                    = -\frac{KNd}{2}\log (2\pi) -\frac{KN}{2}\log \det(\Mv) - \frac{KN}{2}tr[\Mv^{-1}\Sv],
\end{align}
where $\Mv= \Ev\Ev^\top+\sigma^2\Iv$ and $\Sv=\frac{1}{KN}\sum_{k=1}^{K}\sum_{i=1}^{N}(\xv_i^{(k)}-\Fv)(\xv_i^{(k)}-\Fv)^\top$.

{Repeatly using}
the knowledge in pPCA again, we could get the maximum likelihood estimation of the parameters:
\begin{align}
    (\sigma^*)^2 &= \frac{1}{d-q}\sum_{j=q+1}^d \lambda_j \\ 
    \Ev^* &= \mathbf{U}_q\left(\boldsymbol{\Lambda}_q- (\sigma^*)^2\right)^{1 / 2} \mathbf{R}\\
    \Fv^* &= \bf{0},
\end{align}
where $\lambda_{q+1},...,\lambda_d$ are the smallest eigenvalues of the sample covariance matrix $\Sv=\frac{1}{N}\sum_{n=1}^N \xv\xv^\top$, the $d\times q$ orthogonal matrix $\mathbf{U}_q$ 
is made by the $q$ dominant eigenvectors of $\Sv$, the diagonal matrix $\Lambda_q$ contains the corresponding $q$ largest eigenvalues, and $\Rv$ is an arbitary $q\times q$ orthogonal matrix. Note that, when $q=1$, we have $\Rv=\Iv$. Actually, with the same $p(\zv)$ and a decoder $p_\theta(\xv|\zv)$ parameterized by the same linear network, the expression of the maximum likelihood estimation of the $p_\theta(\xv|\zv)$ in the multi-modal case is the same as the single-modal case.

In order to determine $q_\phi(\zv|\xv)$'s parameters, we can initiate the process by identifying the stationary points of $q_\phi(\zv|\xv)$ with respect to the ELBO. The ELBO can be analytically expressed as follows:
{\small
\begin{align}
        \text{ELBO}(\xv) =& \overbrace{\mathbb{E}_{q_\phi(\zv|\xv)}[\log p_\theta(\xv|\zv)]}^{L_1} - \overbrace{D_{\text{KL}}[q_\phi(\zv|\xv)||p(\zv)]}^{L_2}\\ \notag
            L_1 =& \mathbb{E}_{q_\phi(\zv|\xv)}[-\frac{(\Ev\zv-\xv)^\top(\Ev\zv-\xv)}{2\sigma^2}-\frac{d}{2}\log2\pi\sigma^2]\\ \notag
            =&\mathbb{E}_{q_\phi(\zv|\xv)}[\frac{-(\Ev\zv)^\top(\Ev\zv)+2\xv^\top\Ev\zv-\xv^\top\xv}{2\sigma^2}-\frac{d}{2}\log (2\pi\sigma^2)] \\ \notag
            =&\frac{1}{2\sigma^2}[-\text{Tr}(\Ev\Cv\Ev^\top)-(\Ev\Av\xv+\Ev\Bv)^\top(\Ev\Av\xv+\Ev\Bv)+2\xv^\top(\Ev\Av\xv+\Ev\Bv)-\xv^\top\xv] \\ 
            &- \frac{d}{2}\log (2\pi\sigma^2) \\
            L_2 =& \frac{1}{2}[-\log \det(\Cv)+(\Av\xv+\Bv)^\top(\Av\xv+\Bv)+\text{Tr}(\Cv)-q]
\end{align}
}

For a dataset consisting of $KN$ data samples, the stationary points with respect to the ELBO can be obtained through the following expressions:
\begin{align}
    \dfrac{\partial(\sum^{KN} \text{ELBO}(\xv))}{\partial\mathbf{\Av}}=&KN[-\Av\Sv-\Bv\Bar{\xv}^\top-\frac{1}{\sigma^2}(\Ev^\top\Ev\Av\Sv)-\frac{1}{\sigma^2}(\Ev^\top\Ev\Bv\Bar{\xv}^\top-\Ev^\top\Sv)]=\bf{0}\\
    \dfrac{\partial(\sum^{KN} \text{ELBO}(\xv))}{\partial\mathbf{\Bv}}=&KN[-\Av\Bar{\xv}-\frac{1}{\sigma^2}\Ev^\top\Ev\Av\Bar{\xv}+\frac{1}{\sigma^2}\Ev^\top\Bar{\xv}-(\Iv+\frac{\Ev^\top\Ev}{\sigma^2})\Bv] = \bf{0}\\
    \dfrac{\partial(\sum^{KN} \text{ELBO}(\xv))}{\partial\mathbf{\Cv}}=&\dfrac{KN}{2}((\mathbf{C}^{-1})^\top-\mathbf{I}-\dfrac{1}{\sigma^2}(\mathbf{E}^\top\mathbf{E}))=\bf{0} ,
\end{align}
where $\Sv=\frac{1}{KN}\sum^{KN}\xv\xv^\top$ and $\Bar{\xv}=\frac{1}{KN}\sum^{KN}\xv$.
Upon further investigation, we have discovered that the stationary points of $\Av$, $\Bv$, and $\Cv$ solely depend on the parameters $\Ev$ and $\sigma$. In mathematical terms, they can be expressed as:
\begin{align}
    \Av^* =& \frac{(\Iv+\frac{1}{\sigma^2}\Ev^\top\Ev)^{-1}}{\sigma^2}\Ev^\top\\
    \Bv^* =& \bf{0}\\
    \Cv^* =& ((\Iv+\frac{1}{\sigma^2}\Ev^\top\Ev)^\top)^{-1}.
\end{align}

Finally, we can derive the expression of $q(\zv)$ in this multi-modal case as follows:
\begin{equation}
\begin{aligned}
    q(\zv) &= \int_{\xv} q_\phi(\zv|\xv)p(\xv) \\
    &= \int_{\xv} \mathcal{N}(\zv|\Av^* \xv, \Cv^*)\sum_{k=1}^K\pi_k\mathcal{N}(\xv|\muv_k, \Sigmav_k)\\
    &=\sum_{k=1}^{K}\pi_k\int_{\xv} \mathcal{N}(\zv|\Av^* \xv, \Cv^*)\mathcal{N}(\xv|\muv_k, \Sigmav_k)\\
    &=\sum_{k=1}^{K}\pi_k\mathcal{N}(\zv|\Av^* \muv_k, \Av^*\Sigmav_k(\Av^*)^\top+\Cv^*)\\
    &\neq p(\zv).
\end{aligned}
\end{equation}
In conclusion, we observe that $D_{\text{KL}}[q(\zv)||p(\zv)]\neq 0$, indicating that the design of the prior distribution $p(\zv)$ is not appropriate in this multi-modal case and may result in \emph{overestimation} issue of VAE.

\subsection{Implementation Details of Deep VAE}
\label{sec:appen_imple_deep_vae}
The non-linear deep VAE's encoder is implemented as a 3-layer MLP, which takes the 2D data points as inputs. The encoder consists of two linear layers with a hidden dimension of 10 and LeakyReLU activation function \cite{LeakyReLU}. The output layer, with a dimension of 2, does not have an activation function and provides the values for $\mu_z$ and $\log \sigma^2_z$ for each dimension of the latent variable.

For the decoder, it takes the sampled latent variable $\zv$ through reparameterization and feeds it into two linear layers with a hidden dimension of 10 and LeakyReLU activation function. The final output is obtained by a linear layer without activation function, with a dimension of 4. The reconstruction likelihood is modeled as a Gaussian distribution, where the first two dimensions represent $\muv_{\xv}$ (the mean of the reconstruction likelihood) and the remaining dimension represents $\log \sigma_{\xv}^2$ (the log variance of the reconstruction likelihood).

The deep VAE is trained using the Adam optimizer \cite{kingma2014adam} with a learning rate of 1e-5. The training set consists of a total of 10,000 data points.

\subsection{Abalation Study on Dataset Size and Model Architecture for Deep VAE}
\label{sec_app_aba_deep_vae}
{
We also investigated the influence of dataset size (amount of training data) and model architecture (number of neural network layers) on the OOD detection performance of ELBO, using both the synthesized 2D multi-modal dataset and realistic image datasets ("FashionMNIST(ID) / MNIST(OOD)" and "CIFAR-10(ID) / SVHN(OOD)"). Our findings are illustrated in Fig.~\ref{fig:app_more_multi_case} and Table \ref{tab:size_capacity_image}. For the 2D multi-modal dataset, we sampled a data volume 10 times greater than its inherent distribution p(x) than the original configuration seen in Fig.~\ref{fig:multi_case}(a-b) of the main paper, increasing from 10,000 to 100,000 training samples. The VAE for this experiment utilized a 10-layer MLP as opposed to the original 3-layer MLP. Notably, the results from Fig.~\ref{fig:app_more_multi_case}(a) highlight that the is still not equal to p(z) = N (0, I) and Fig.~\ref{fig:app_more_multi_case} (b) indicates the persistence of the \emph{overestimation} problem in the non-linear deep VAE. 
For the practical image datasets, we varied the dataset size and model architecture (number of CNN layers) to investigate their effects on ELBO's OOD detection performance. However, results show that \textbf{increasing the amount of data and the number of CNN layers does not yield significant improvements.}
}

\begin{figure*}[h!]
    \centering
    \subfigure[$q_{{id}}(\zv)$]{\includegraphics[width=0.43\textwidth]{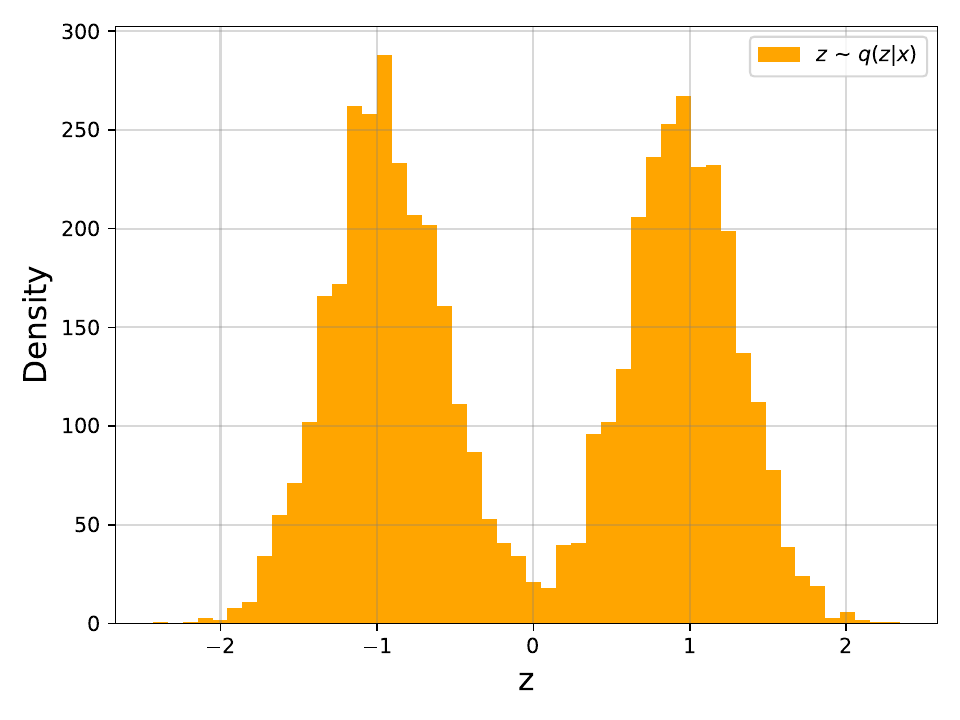}
    }
    \subfigure[Estimated $p_\theta(\xv)$]{\includegraphics[width=0.45\textwidth]{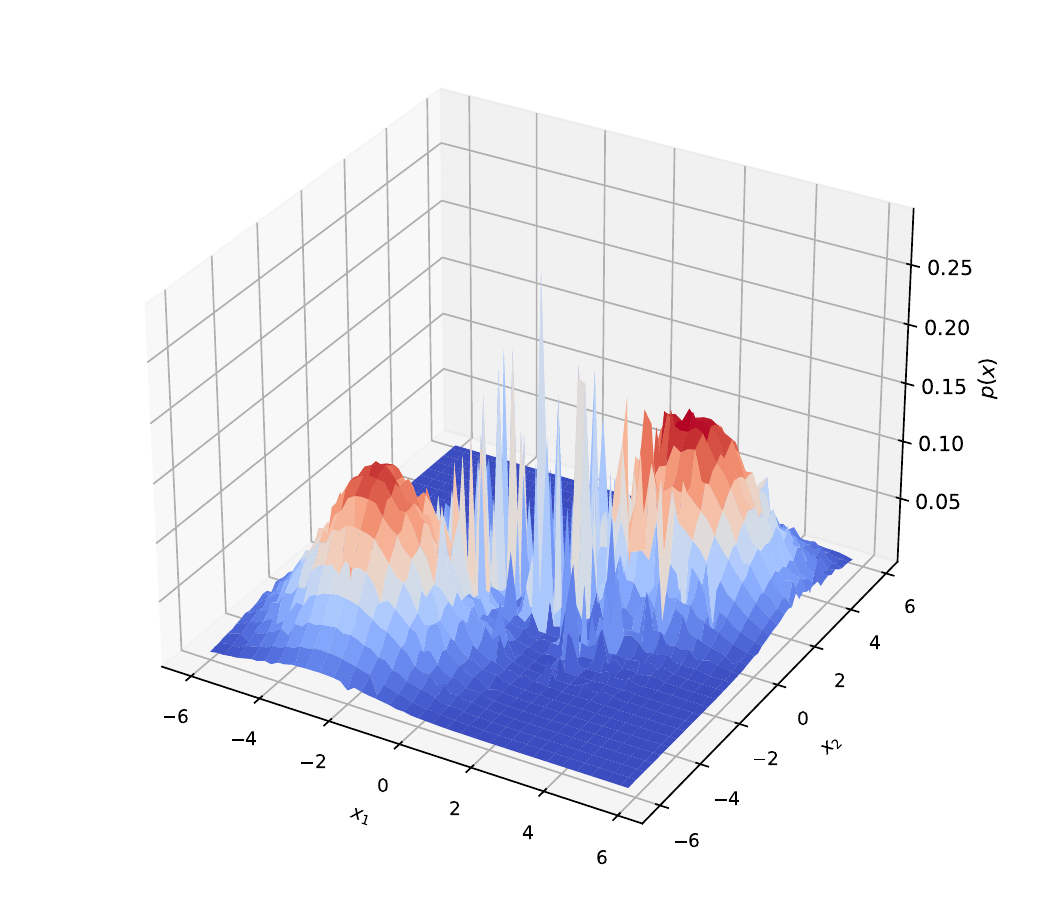}
    }
    \caption{
    {Visualization of $q_{{id}}(\zv)$ and estimated $p_\theta(\xv)$ by $\text{ELBO}$ on a synthesized 2D multi-modal dataset. The data amount here is 10 times larger than in Fig. \ref{fig:multi-mode-px}, increasing from 10,000 to 100,000 samples. The VAE used is a non-linear deep one based on a 10-layer MLP, in contrast to the 3-layer MLP used in Fig. \ref{fig:multi-mode-ELBO-pz}. Results indicate that the $q_{id}(\zv)$ is still not equal to $p(\zv)=\mathcal{N}(0,\Iv)$ and the paradox still exists near the region of (0, 0).}}
    \label{fig:app_more_multi_case}
\end{figure*}

\begin{table}[h!]
\centering
\caption{Ablation study examining the effects of dataset size (data amount) and model capacity (number of convolutional neural network (CNN) layers) on the OOD detection performance of ELBO. Results indicate that increasing the amount of data and the number of CNN layers does not yield significant improvements.}
\setlength\tabcolsep{6.5pt}
\renewcommand{\arraystretch}{1.3}
\scalebox{0.9}{
\begin{tabular}{cccccccccccc}
\hline
 \multicolumn{6}{c}{FashionMNIST(ID) / MNIST(OOD)}       &              \multicolumn{6}{c}{CIFAR-10(ID) / SVHN(OOD)} \\ \hline
                & \multicolumn{5}{c}{Num. of Layers}                      &             & \multicolumn{5}{c}{Num. of Layers}           \\ \hline
Data Amount     & 3    & 6    & 9    & 12   & \multicolumn{1}{c|}{15}   & Data Amount & 3      & 6      & 9      & 12     & 15     \\ \hline
10000           & 9.45 & 14.0 & 13.2 & 14.2 & \multicolumn{1}{c|}{14.6} & 10000       &14.4    &12.8    &16.9   &20.5    &20.3        \\
30000           & 16.3 & 14.5 & 15.3 & 14.5 & \multicolumn{1}{c|}{15.8} & 30000       &24.6    & 25.3   &25.9   &24.4    &23.9        \\
60000           & 23.5 & 25.1 & 23.0 & 20.3 & \multicolumn{1}{c|}{19.8} & 50000       &24.9   & 22.6   & 23.5   & 28.1   &24.0   \\ \hline
\end{tabular}}
\label{tab:size_capacity_image}
\end{table}

\section{Details of Experimental Setup} \label{sec:detailed_setup}
\subsection{Description of all Datasets} \label{sec:appen_exp_datasets}

In accordance with the existing literature 
\cite{flow_cannot, hard_datasets, dose}
we evaluate our method against previous works using { commonly acknowledged ``\textbf{harder}'' dataset pairs \cite{dose, hard_datasets}: 
FashionMNIST (ID) $\rightarrow$ MNIST (OOD), CIFARs (ID) $\rightarrow$ SVHN (OOD), and CelebA (ID) $\rightarrow$ CIFARs (OOD). 
The suffixes ``ID'' and ``OOD'' represent in-distribution and out-of-distribution datasets, respectively. 

To more comprehensively assess the generalization capabilities of these methods, we incorporate additional OOD datasets as follows.
Notably, datasets featuring the suffix ``-G'' (\textit{e.g.}, ``CIFAR-10-G'') have been converted to grayscale, resulting in a single-channel format.


For grayscale image datasets, we utilize the following datasets: FashionMNIST \cite{fashionmnist}, MNIST \cite{mnist}, KMNIST \cite{kmnist}, notMNIST \cite{notmnist}, Omniglot \cite{omniglot}, and several grayscale datasets transformed from RGB datasets.
\textbf{FashionMNIST} is a dataset consisting of 60,000 grayscale images of Zalando's article pictures for training, and 10,000 images for testing. Each image is 28x28 pixels and belongs to one of the 10 classes.
\textbf{MNIST} is a widely used dataset containing 70,000 grayscale images of handwritten digits. It consists of a training set of 60,000 images and a test set of 10,000 images. Each image is 28x28 pixels.
\textbf{KMNIST} is derived from the Kuzushiji Dataset and serves as a drop-in replacement for the MNIST dataset. It includes 70,000 grayscale images, each with a resolution of 28x28 pixels.
\textbf{notMNIST} is a dataset composed of 547,838 grayscale images of glyphs extracted from publicly available fonts. The images are 28x28 pixels in size and cover letters A to J from various fonts.
\textbf{Omniglot} contains 32,460 grayscale images of 1623 different handwritten characters from 50 distinct alphabets. Each image has a resolution of 28x28 pixels.
Additionally, we have transformed several RGB datasets into grayscale versions, including CIFAR-10-G, CIFAR-100-G, SVHN-G, CelebA-G, and SUN-G.


For RGB datasets, we utilize the following datasets: CIFAR-10/CIFAR-100 \cite{cifar10}, SVHN \cite{svhn}, CelebA \cite{celeba}, Places365 \cite{places365}, Flower102 \cite{flower102}, LFWPeople \cite{lfwpeople}, SUN \cite{SUN}, GTSRB \cite{GTSRB}, and Texture \cite{dtd} datasets.
\textbf{CIFAR-10} and \textbf{CIFAR-100} are datasets consisting of 32x32 color images. CIFAR-10 contains 50,000 training images and 10,000 testing images, with 10 different classes. CIFAR-100 has the same number of images but includes 100 classes.
\textbf{SVHN} is a dataset obtained from Google Street View images, primarily used for recognizing digits and numbers in natural scene images.
\textbf{CelebA} is a large-scale face attributes dataset containing over 200,000 celebrity images, each annotated with 40 attribute labels.
\textbf{Places365} is a dataset that includes 1.8 million training images from 365 scene categories. The validation set contains 50 images per category, and the testing set contains 900 images per category.
\textbf{Flower102} is an image classification dataset consisting of 102 flower categories, with each class containing between 40 and 258 images. The selected flowers are commonly found in the United Kingdom.
\textbf{LFWPeople} contains more than 13,000 images of faces collected from the web, making it a popular dataset for face-related tasks.
\textbf{SUN} is a large-scale scene recognition dataset, covering a wide range of scenes from abbey to zoo.
\textbf{GTSRB} is a dataset specifically developed for the task of German traffic sign recognition.
\textbf{Texture} is an evolving collection of textured images in various real-world settings.
All images from these datasets are resized to the dimensions of 32x32x3 before being used as input for the models.

\subsection{Evaluation and Metrics} \label{sec:appen_eval_metrics}
We adhere to the previous evaluation procedure \cite{hvae_ood, informative-vae}, where all methods are trained using the training split of the ID dataset, and their OOD detection performance is assessed on both the testing split of the ID dataset and the OOD dataset.
In line with previous works \cite{hendrycks2016baseline, alemi2018uncertainty, hendrycks2018deep}, we employ evaluation metrics including the area
under the receiver operating characteristic curve (AUROC $\uparrow$), the area under the precision-recall curve (AUPRC $\uparrow$), and the false positive rate at 80\% true positive rate (FPR80 $\downarrow$). The arrows indicate the direction of improvement for each metric.

\subsection{Implementation Details}\label{sec:appen_exp_imples}
The VAE's latent variable's dimension is set as 200 for all experiments with the encoder and decoder parameterized by a 3-layer convolutional neural network, respectively. 
The reconstruction likelihood distribution is modeled by a discretized mixture of logistics \cite{salimans2017pixelcnn++}.
For optimization, we adopt the Adam optimizer \cite{kingma2014adam} with a learning rate of 1e-3. 
We train all models in comparison by setting the batch size as 128 and the max epoch as 1000 following \cite{hvae_ood, informative-vae}. All experiments are performed on a PC with an NVIDIA A100 GPU and implemented with PyTorch \cite{paszke2019pytorch}. 

The encoder of the VAE is implemented as a 3-layer convolutional network with kernel numbers of 32, 64, and 128, and strides of 1, 2, and 2, respectively. The ReLU \cite{CNN} activation function is applied. The output layer consists of a linear layer that outputs the mean and log-variance of the latent variables, with a dimension of 200.

On the other hand, the decoder takes the reparameterized latent variables as input and utilizes a 3-layer transposed convolutional network. The network has kernel numbers of 128, 64, and 32, and strides of 2, 2, and 1, respectively. The ReLU activation function is used. Finally, the output layer is parameterized by a convolutional layer that models the distribution as a discretized mixture of logistics.

In the PHP method, an LSTM is employed as the backbone \cite{lstm}. The hidden size of the LSTM is set to 64, and the outputted hidden state is fed into a 3-layer linear network. The hidden sizes of the linear layers are 64, 32, and 2, respectively. The ReLU activation function is applied to the first two layers. The optimizer used for learning the $q(\zv)$ distribution is Adam, and the learning rate is set to 1e-4.

\subsection{Algorithm for the Dataset Entropy-mutual Calibration Method}\label{sec:appen_method}
We provide a pseudo code here for calculating the $\mathcal{C}(\xv)$ of a testing sample $\xv$ in Algorithm \ref{alg:my_algo}. Note that, the maximum number of singular values $N$ should be larger than $n_{\text{id}}$. In our experiments, we set $N={n_{id}^{max}}$ and $n_{id}={n_{id}^{max}}/2$. The implementation can easily employ a \textbf{binary search} scheme for enhanced computational efficiency. 

\begin{algorithm}[h!]
\caption{Dataset {entropy-mutual} calibration $\mathcal{C}(\xv)$ algorithm}\label{alg:my_algo}
\begin{algorithmic}
\STATE {\textbf{Input:} Hyperparameter $n_{\text{id}}$ and its corresponding reconstruction error $\epsilon=\mathbb{E}_{\xv\sim p_{\text{id}}}|\xv_{\text{recon}} - \xv|$}, maximum number of singular values $N$, a testing sample $\xv$.
\STATE {\textbf{Ouput:} $\mathcal{C}(\xv)$.}
\STATE Do SVD for the testing sample $\xv$;
\FOR {$n_i=1$ to $N$}
    \STATE Calculate reconstruction error $\epsilon_{i}$ using $n_i$ singular values;
    \IF {$\epsilon_{i}\leq \epsilon$}
        \STATE \textbf{break};
    \ENDIF 
\ENDFOR
\IF {$n_i<n_{\text{id}}$}
    \STATE Calculate $\mathcal{C}(\xv) = {n_i}/{n_{\text{id}}}$;
\ELSE
    \STATE Calculate $\mathcal{C}(\xv) = (n_{\text{id}}-(n_i - n_{\text{id}}))/{n_{\text{id}}}$;
\ENDIF
\STATE \textbf{return} $\mathcal{C}(\xv)$
\end{algorithmic}
\end{algorithm}

{
\section{Experiments on the Diffusion Model} \label{sec_re_diff}

\begin{table}[ht!]
\centering
\caption{{\rebuttal The comparisons to the likelihood of a Diffusion and a 10-layer latent diffusion model (``Shallow Diffusion'') with our method post-hoc prior (denoted as ``PHP''), dataset {entropy-mutual} calibration (denoted as ``DEC''), and ``Resultant'' based on a  10-layer latent diffusion model.}}
\setlength\tabcolsep{10pt}
\renewcommand{\arraystretch}{1.2}
\scalebox{0.75}{
\begin{tabular}{lccc|lccc}
\hline
\multicolumn{4}{c|}{\textbf{{FashionMNIST(ID) / MNIST(OOD)}}}                                                 & \multicolumn{4}{c}{\textbf{CIFAR-10(ID) / SVHN(OOD)}}                                                    \\ \hline
Method                   & AUROC$\uparrow$      & AUPRC$\uparrow$    & FPR80$\downarrow$     & Method                  & AUROC$\uparrow$      & AUPRC$\uparrow$    & FPR80$\downarrow$    \\ \hline
Diffusion  &18.0 &34.1 &97.4 &Diffusion &19.3 &35.0 &97.2 \\
Shallow Diffusion             & 11.5                & 34.6                &  98.6                 & Shallow Diffusion         & 4.11        & 31.14         &  99.8                 \\
PHP     & 80.9                 & 83.5                & 29.5                  & PHP    & 43.0                 & 46.0                & 86.2               \\
DEC     & 34.7                 & 39.7                 & 90.6             &DEC     &  88.9               &  90.1             &  12.8               \\
Resultant    & 85.9                 & 87.5                & 19.5            & Resultant   & 95.0        & 95.1       & 4.11       \\ \hline
\end{tabular}
}
\label{tab_diffusion}
\end{table}

\begin{table}[t!]
\centering
\caption{{\rebuttal The reverse verification comparisons to the likelihood of a Diffusion and a 10-layer latent diffusion model (``Shallow Diffusion'') with our method post-hoc prior (denoted as ``PHP''), dataset {entropy-mutual} calibration (denoted as ``DEC''), and ``Resultant'' based on a  10-layer latent diffusion model.}}
\setlength\tabcolsep{10pt}
\renewcommand{\arraystretch}{1.2}
\scalebox{0.75}{
\begin{tabular}{lccc|lccc}
\hline
\multicolumn{4}{c|}{\textbf{{MNIST(ID) / FashonMNIST(OOD)}}}                                                 & \multicolumn{4}{c}{\textbf{SVHN(ID) / CIFAR-10(OOD)}}                                                    \\ \hline
Method                   & AUROC$\uparrow$      & AUPRC$\uparrow$    & FPR80$\downarrow$     & Method                  & AUROC$\uparrow$      & AUPRC$\uparrow$    & FPR80$\downarrow$    \\ \hline
Diffusion  &99.9 &99.9 &0.00 &Diffusion &99.9 &99.9 &0.00 \\
Shallow Diffusion             & 99.9                & 99.9               &  0.00                 & Shallow Diffusion         & 99.9        & 99.9         &  0.00                  \\
PHP     & 99.9                 & 99.9                & 0.00                  & PHP    & 99.9                 & 99.9                & 0.00               \\
DEC     & 99.9                 & 99.9                 & 0.00             &DEC     &  99.9               &  99.9             &  0.00               \\
Resultant    & 99.9                 & 99.9                & 0.00            & Resultant   & 99.9        & 99.9       & 0.00       \\ \hline
\end{tabular}
}
\label{tab_diffusion_reverse}
\end{table}

As the diffusion model is a more powerful and complex variational DGM, whose training objective is also derived from a variational evidence lower bound (ELBO) as an estimation of the data marginal log-likelihood \cite{DDPM}. We testify our method with replace the original backbone (VAE) to a diffusion model. 
We first directly testify the OOD detection performance of the likelihood of a trained diffusion model with 1000 time steps \cite{DDPM}, \textit{i.e.}, $\{\xv_t\}_{t=0}^{T=1000}$ where $\xv_0$ is the input data $\xv$ and $\xv_1,...,\xv_T$ could be seen as the latent variables. As Table \ref{tab_diffusion} shows, directly applying the likelihood of a diffusion model could still suffer from the paradox issue in the "hard" benchmarks.
Therefore, it would be interesting to see whether our method could also improve the performance of a diffusion model.

To apply our Resultant method to diffusion models, we need to fit its aggregated posterior distribution $q_{id}(\zv)$ at first. Let us recall the estimated likelihood of a $T$ time step diffusion model, expressed as 
{\small
\begin{align}
    \log p_\theta(\xv)= \mathbb{E}_{q(\xv_1|\xv_0)}\log p_\theta(\xv_0|\xv_1) - \sum_{t=2}^{T}D_{\text{KL}}[q(\xv_{t-1}|\xv_t,\xv_0)||p_\theta(\xv_{t-1}|\xv_t)] - D_{\text{KL}}[q(\xv_T|\xv_0)||p(\xv_T)],
\end{align}
}

where $q(\xv_1, \xv_2, ..., \xv_T)$ is decomposed to a product of $T$ terms $q(\xv_{t-1}|\xv_t)$ for every time step, \textit{i.e.},
{\small
\begin{align}
    q(\xv_1, \xv_2, ..., \xv_T) = q(\xv_T)\prod_{t=2}^{T} q(\xv_{t-1}|\xv_t).
\end{align}
}

To fit the $q(\xv_1, \xv_2, ..., \xv_T)$ for applying PHP method, we need $T$ individual LSTMs to fit every step's $q(\xv_{t-1}|\xv_t)$.
However, as the dimension of $\xv_t$ is the same as the input data, it could be difficult or even impossible to fit the $q(\xv_{t-1}|\xv_t)$ well. 

Recognizing this difficulty, we use a variant of the diffusion model, \textit{i.e.}, latent diffusion model.
For latent diffusion models \cite{latent_diffusion}, to improve the computational efficiency, the $\xv_1$ is replaced by a low-dimension latent variable $\zv_1$ encoded by a auto-encoder, then the following $\xv_2,...,\xv_T$ is replaced by $\zv_2, ..., \zv_T$. Therefore, the ELBO of the latent diffusion model is expressed by
{\small
\begin{align}
    \text{ELBO}(\xv)= \mathbb{E}_{q_\phi(\zv_1|\xv_0)}\log p_\theta(\xv_0|\zv_1) - \sum_{t=2}^{T}D_{\text{KL}}[q(\zv_{t-1}|\zv_t,\xv_0)||p_\theta(\zv_{t-1}|\zv_t)] - D_{\text{KL}}[q(\zv_T|\xv_0)||p(\zv_T)],
\end{align}
}

and the latent $q(\zv_1, \zv_2, ..., \zv_T)$ is decomposed to a product of $T$ terms $q(\zv_{t-1}|\zv_t)$ for every time step,  \textit{i.e.},
{\small
\begin{align}
    q(\zv_1, \zv_2, ..., \zv_T) = q(\zv_T)\prod_{t=2}^{T} q(\zv_{t-1}|\zv_t).
\end{align}
}

To fit the $q(\zv_1, \zv_2, ..., \zv_T)$ for applying PHP method, we need $T$ individual LSTMs $\hat{q}_{id}(\zv_{t-1}|\zv_{t})$ to fit every step's $q(\zv_{t-1}|\zv_t)$ on the training set and then change the ELBO to 
{\small
\begin{align}
    \text{PHP}(\xv)= \log p_\theta(\xv_0|\zv_1) - \sum_{t=2}^{T}D_{\text{KL}}[q(\zv_{t-1}|\zv_t,\xv_0)||\hat{q}_{id}(\zv_{t-1}|\zv_t)] - D_{\text{KL}}[q(\zv_T|\xv_0)||\hat{q}_{id}(\zv_T)].
\end{align}
}

However, the $T$ could still be too large, \textit{e.g.}, when $T=1000$, we still need to train 1000 individual models to fit $q(\zv_{t-1}|\zv_t)$, 
which we think is very low-efficiency in computation.
Thus, we consider a shallow latent diffusion model, termed ``Shallow-Diffusion'', where the $T$ is set as 10. Besides, we let it directly maximize the evidence lower bound as its training objective with a trainable encoder, which could also be interpreted as a hierarchical VAE \cite{unify}.

As results show in Table \ref{tab_diffusion} and Table \ref{tab_diffusion_reverse}, our method could improve the U-OOD detection performance and achieve incremental effectiveness on likelihood. 
}

\section{More Experimental Results on Verifying the Post-hoc Prior}
\label{sec:appen_ablation_prior}
\subsection{Why does PHP not work well in some cases?}\label{sec_re_umap}
We provide a further analysis for the PHP method of when it would not work well, specifically why it works well in FashionMNIST (ID) / MNIST (0OD) dataset pair but not work well on CIFAR-10 (ID) / SVHN (OOD) dataset pair.  
Additionally, we replace t-SNE with UMAP (Uniform Manifold Approximation and Projection) \cite{umap} for visualizing the latent representations. 
{ We also provide the \textbf{implementation details for t-SNE and UMAP} here:
for t-SNE, we import the "TSNE()" function from "sklearn.manifold" python library and set all the hyper-parameters as default; for UMAP, we import the "UMAP()" function from the "UMAP" python library with its default hyper-parameters setting. Then, we use their "fit\_transform()" function directly on the extracted latent codes. 
}

As the results shown in Fig.~\ref{fig:umap}, the latent variable's aggregated posterior distribution of FashionMNIST $q_{fashion}(\zv)$ is pretty distinguishable from that of MNIST $q_{mnist}(\zv)$ but the aggregated posterior distribution of CIFAR-10 $q_{cifar}(\zv)$ has some overlapping with SVHN $q_{svhn}(\zv)$, which may due to the shared low-level features across CIFAR-10 and SVHN datasets. 
Thus, the $D_{\text{KL}}[q_{svhn}(\zv)||\hat{q}_{cifar}(\zv)]$ could be smaller than $D_{\text{KL}}[q_{mnist}(\zv)||\hat{q}_{fashion}(\zv)]$, which indicates the reason for why PHP works well in FashionMNIST (ID) / MNIST (OOD) but not that well in CIFAR-10 (ID) / SVHN (OOD).
This also inspires us that encoding more dataset-specific semantic information into the latent variables could further improve the performance of the PHP method.

\begin{figure*}[h!]
    \centering
    \subfigure[FashionMNIST(ID)/MNIST(OOD)]{\includegraphics[width=0.45\textwidth]{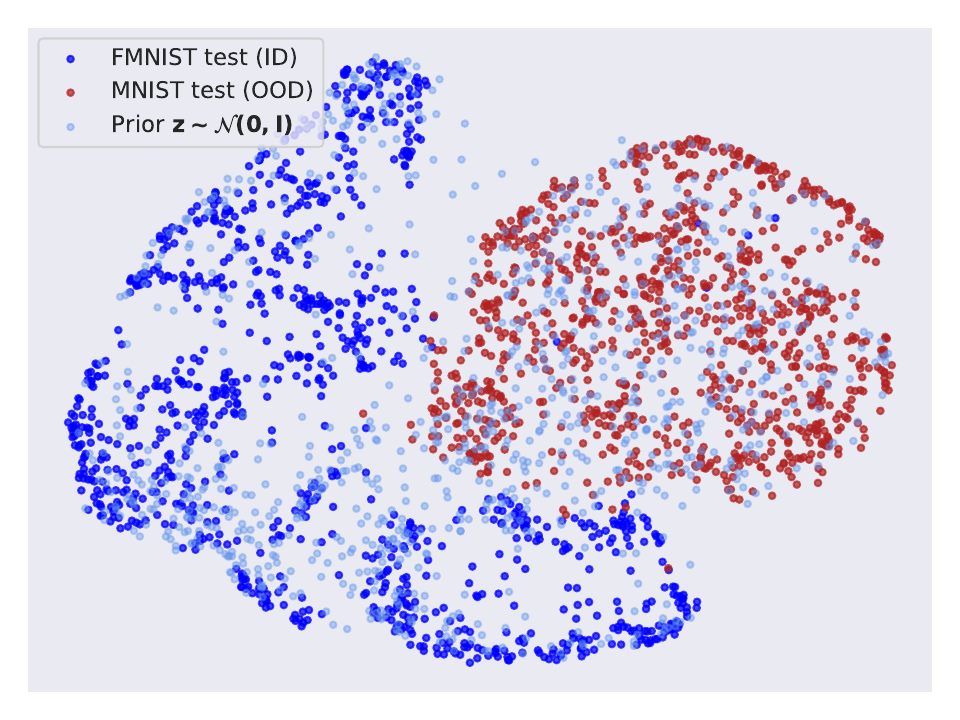}
    \label{fig:umap_fm}
    }
    \subfigure[CIFAR-10(ID)/SVHN(OOD)]{\includegraphics[width=0.45\textwidth]{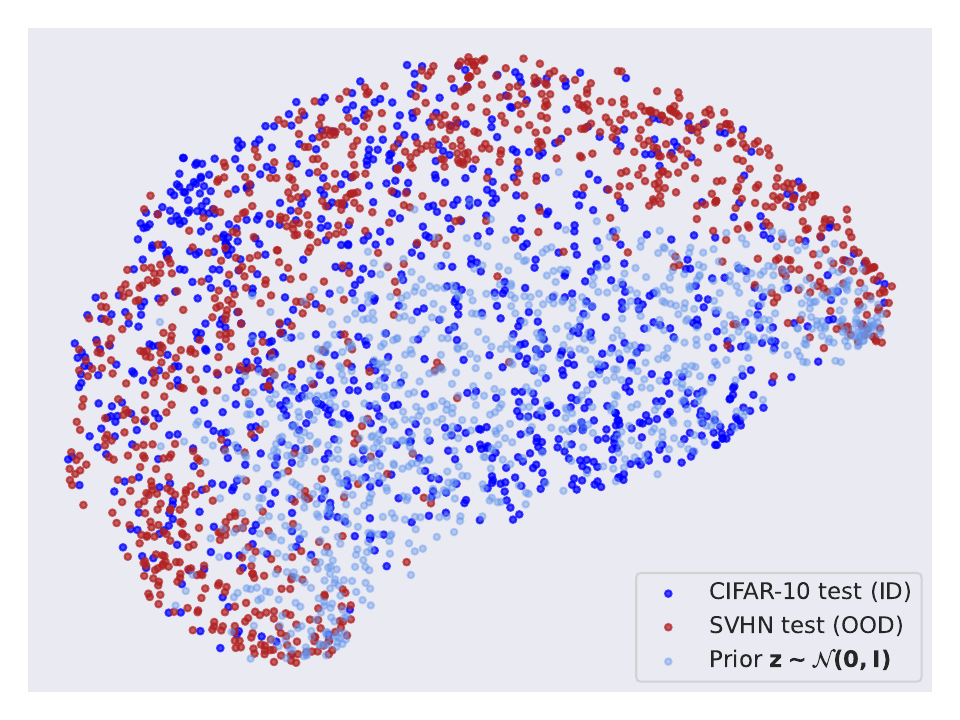}
    \label{fig:umap_cf}
    }
    \caption{
    {{\rebuttal The UMAP visualization of latent representations on FashionMNIST(ID) / MNIST(OOD) 
 and CIFAR-10(ID) / SVHN(OOD) dataset pairs.}}}
    \label{fig:umap}
\end{figure*}

\subsection{Evaluation of PHP Method on Vertically Flipped OOD Data}\label{sec_re_vflip}
We have added experiments on detecting vertically flipped data as OOD data in Table \ref{tab_vflip}. Since the flipped dataset maintains the same entropy as the original dataset, direction II has a diminished effect in improving incremental effectiveness. Consequently, the DEC method, which relies on image compressors, becomes ineffective. Therefore, the Resultant method must primarily depend on the PHP method to detect the vertically flipped data. As shown in Table \ref{tab_vflip}, the PHP method continues to perform effectively even in this extreme scenario.

\begin{table}[h!]
\centering
\caption{Comparison with likelihood on detecting vertically flipped (``VFlip'') data as OOD. }
\setlength\tabcolsep{13pt}
\renewcommand{\arraystretch}{1.3}
\scalebox{0.74}{
\begin{tabular}{lccccc}
\hline
\multicolumn{6}{c}{AUROC $\uparrow$ in detecting VFlip data as OOD.}               \\ \hline
\multicolumn{1}{l|}{Method}      & CelebA & CIFAR-10 & SVHN & FashionMNIST & MNIST \\ \hline
\multicolumn{1}{l|}{Likelihood \cite{kingma2013auto}}        & 74.2   & 49.5     & 50.4 & 69.5         & 82.7  \\
\multicolumn{1}{l|}{Resultant(=PHP)} & 85.7   & 53.7     & 52.7 & 86.2         & 84.9  \\ \hline
\end{tabular}
}
\label{tab_vflip}
\end{table}

\subsection{More Datasets for Evaluating the PHP method}
We {evaluate the effectiveness of}
the PHP method on additional datasets as shown in Table \ref{tab:comparison_more_datasets_kl} and Table \ref{tab_cifar100_php}.
\begin{table}[h!]
\centering
\caption{The comparisons of the U-OOD detection performance of our ``DEC'' and likelihood on more datasets. The new score function only has \textbf{post-hoc prior} part.}
\setlength\tabcolsep{9pt}
\renewcommand{\arraystretch}{1.3}
\scalebox{0.7}{
\begin{tabular}{c|ccc|c|ccc}
\hline
ID   & \multicolumn{3}{c|}{FashionMNIST}                        & ID   & \multicolumn{3}{c}{CIFAR-10}                            \\ \hline
OOD  & AUROC $\uparrow$  & AUPRC $\uparrow$ & FPR80 $\downarrow$ & OOD  & AUROC $\uparrow$  & AUPRC $\uparrow$ & PFR80 $\downarrow$ \\ \hline
             & \multicolumn{3}{c|}{Likelihood / PHP \textbf{(ours)}} &              & \multicolumn{3}{c}{Likelihood / PHP \textbf{(ours)}} \\ \hline
KMNIST       &60.03 / \textbf{72.98}   &54.60 / \textbf{69.34}    &61.6 / \textbf{48.1}     & CelebA       &57.27 / \textbf{70.91}        &54.51 / \textbf{72.16}     &69.03 / \textbf{52.95}                \\
Omniglot     &99.86 / \textbf{99.90}    &99.89 / \textbf{99.89}   &0.00 / \textbf{0.00}     & CIFAR-100    &52.91 / \textbf{55.00}        &51.15 / \textbf{54.01}     &77.42 / \textbf{70.23} \\
notMNIST     &94.12 / \textbf{94.39}    &94.09 / \textbf{94.35}   &8.29 / \textbf{7.79}     & Places365    &57.24 / \textbf{57.36}        &56.96 / \textbf{56.55}     &73.13 / \textbf{52.95}           \\
CIFAR-10-G    &98.01 / \textbf{98.84}    &98.24 / \textbf{99.13}   &1.20 / \textbf{0.30}     & LFWPeople    &64.15 / \textbf{64.57}        &59.71 / \textbf{65.20}     &59.44 / \textbf{64.74}                 \\
CIFAR-100-G   &98.49 / \textbf{98.50}    &97.49 / \textbf{97.50}    &1.00 / \textbf{0.90}    & SUN          &53.14 / \textbf{53.27}        &54.48 / \textbf{54.67}     &79.52 / \textbf{78.12}         \\ 
SVHN-G       &95.61 / \textbf{96.00}    &96.20 / \textbf{97.13}    &3.00 / \textbf{0.60}    & Texture          &37.86 / \textbf{43.38}        &40.93 / \textbf{43.99}     &82.22 / \textbf{80.12}               \\ 
CelebA-G     &97.33 / \textbf{97.71}    &94.71 / \textbf{95.62}    &3.00 / \textbf{2.20}    & Flowers102   &67.68 / \textbf{67.76}        &64.68 / \textbf{64.75}     &57.94 / \textbf{57.63}          \\ 
SUN-G        &99.16 / \textbf{99.26}    &99.39 / \textbf{99.40}    &0.00 / \textbf{0.00}    &GTSRB         &39.50 / \textbf{52.62}        &41.73 / \textbf{50.81}     &86.61 / \textbf{75.12}                   \\ 

Const        &94.94 / \textbf{95.08}    &97.27 / \textbf{97.35}    &1.80 / \textbf{0.00}    & Const        &0.001 / \textbf{15.70}        &30.71 / \textbf{30.78}     &100.0 / \textbf{86.62}                  \\
Random       &99.80 / \textbf{99.81}    &99.90 / \textbf{99.90}    &0.00 / \textbf{0.00}    & Random       &71.81 / \textbf{72.52}        & 82.89 / \textbf{83.42}     &85.71 / \textbf{85.00}                \\ \hline
\end{tabular}
}
\label{tab:comparison_more_datasets_kl}
\end{table}

\begin{table}[h!]
\centering
\caption{{\rebuttal Comparison of U-OOD detection performance between our PHP method and likelihood with models trained on CIFAR-100 (ID). Bold are the best results.}}
\setlength\tabcolsep{14pt}
\renewcommand{\arraystretch}{1.5}
\scalebox{0.65}{
\begin{tabular}{c|ccc}
\hline
ID                    & \multicolumn{3}{c}{CIFAR-100}                            \\ \hline
OOD                   & AUROC $\uparrow$ & AUPRC $\uparrow$ & FPR80 $\downarrow$ \\ \hline
\multicolumn{1}{l|}{} & \multicolumn{3}{c}{\textbf{Likelihood / PHP(ours)}}            \\ \hline
CelebA                & 58.2 / \textbf{65.0} & 56.0 / \textbf{64.9} & 65.8 / \textbf{65.9}                   \\
Places365             & 56.5 / \textbf{69.5}            & 55.5 / \textbf{67.2}            & 74.6 / \textbf{52.0}                  \\
LFWPeople             & 63.9 / \textbf{74.3}            & 58.4 / \textbf{72.4}            & 61.3 / \textbf{46.6}                  \\
SUN                   & 58.3 / \textbf{60.0}            & 55.6 / \textbf{57.9}            & 61.0 / \textbf{60.6}                  \\
Texture               & 52.7 / \textbf{55.4}            & 48.4 / \textbf{51.5}            & 66.1 / \textbf{64.4}                   \\
Flowers102            & 80.5 / \textbf{84.3}            & 80.1 / \textbf{80.8}            & 23.9 / \textbf{23.8}                  \\
GTSRB                 & 58.7 / \textbf{67.5}            & 51.8 / \textbf{59.6}            & 49.4 / \textbf{49.1}                 \\
Const                 & 0.00 / \textbf{2.40}            & 0.00 / \textbf{31.8}            & 100. / \textbf{100.}                  \\
Random                & 100. / \textbf{100.}            & 100. / \textbf{100.}            & 0.00 / \textbf{0.00}                 \\ \hline
\end{tabular}}
\label{tab_cifar100_php}
\end{table}

\section{More Experimental Results on Verifying the Dataset {Entropy-mutual} Calibration} \label{sec:appen_ablation_ent}

\subsection{Different Implementation of the DEC Method and Computational Efficiency} \label{sec_re_compressor}
We kindly emphasize that the SVD is only one of the choices for implementing the DEC method, which is simply implemented to support our analysis and the method's effectiveness.
Other compressors like JPEG and PNG compressors could also be directly applied to implement the DEC method. 
In addition to the originally implemented DEC method based on SVD, termed "DEC(SVD)", we implement three different variants of the DEC method for comparison: DEC(SVD-bs), DEC(JPEG), and DEC(PNG).
``DEC(SVD-bs)'' means a binary search version of the ``DEC(SVD)'', where the $n_i$ is determined by binary searching from 1 to $2*n_{id}$.
``DEC(JPEG)'' and ``DEC(PNG)'' are implementing DEC with different compressors, JPEG compressor and PNG compressor. They could provide a value $bit(\xv)$ indicating the complexity of an image $\xv$, then we could get the average $bit(\xv)$ of the training set, termed ``$bit_{{id}}$''. Thus, we could replace the $n_i$ with $bit(\xv)$ to implement the $\mathcal{C}(\xv)$ DEC method as follows:
\begin{align}
    \mathcal{C}(\xv) =
    \begin{cases}
        (bit(\xv)/bit_{{id}}), & \text{if} \quad bit(\xv) < bit_{{id}}, \\
        (bit_{{id}} - (bit(\xv) - bit_{{id}}))/bit_{{id}}, & \text{if} \quad bit(\xv) \geq bit_{{id}},
        \label{eq:dec_property1}
    \end{cases}
\end{align}
which could avoid the low-efficiency searching of $n_i$ in the DEC(SVD) method.

We compare their performance in Table \ref{tab_compressor}.
Actually, we find that the relationship between the data type and compressor plays an important role in its performance, \textit{e.g}, the ratio of the average bit, \textit{bits}(FashionMNIST)/\textit{bits}(MNIST), compressed by JEPG is only 1.12, which leads to its poor performance; in contrast, the ratio of the average bit, \textit{bits}(FashionMNIST)/\textit{bits}(MNIST), compressed by PNG is 1.92, which is much larger and leads to its best performance on this dataset pair.

\begin{table}[h!]
\centering
\caption{{\rebuttal U-OOD detection performance of DEC method with different compressors.}}
\setlength\tabcolsep{10pt}
\renewcommand{\arraystretch}{1.2}
\scalebox{0.9}{
\begin{tabular}{lccc|lccc}
\hline
\multicolumn{4}{c|}{\textbf{{FashionMNIST(ID)/MNIST(OOD)}}}                                                 & \multicolumn{4}{c}{\textbf{CIFAR-10(ID)/SVHN(OOD)}}                                                    \\ \hline
Method                   & AUROC$\uparrow$      & AUPRC$\uparrow$    & FPR80$\downarrow$     & Method                  & AUROC$\uparrow$      & AUPRC$\uparrow$    & FPR80$\downarrow$    \\ \hline
SVD            & 34.1                & 40.7                & 92.5                  & SVD          & 87.8                & 89,9                 & 17.8                 \\
JPEG                      & 31.1       & 36.6        & 93.0                   & JPEG                   & 90.4                 &88.6                  & 12.5                 \\
PNG    &  99.9                & 99.9                & 0.00        & PNG   & 80.5       &75.4         &  26.4     \\ \hline
\end{tabular}
}
\label{tab_compressor}
\end{table}

For the computational efficiency, we conduct comparisons from two perspectives: inference time and training time, where results are shown in Table \ref{tab:computation_infer_time} and Table \ref{tab:computation_training_time}, respectively.
First, we testify to the computational efficiency regarding the inference by recording the average inference time in scoring a data point on an A100 GPU. The time is averaged by sampling 100 data points from each of all the used gray-scale datasets or RGB nature datasets. 
We also provide a discussion for the ensemble method WAIC \cite{choi2018waic}, it relies on an ensemble of $N$ generative models $\{\log p_{\theta_i}(\xv)\}_{i=1}^N$ to detect the OOD data with a score function:
\begin{align}
    \mathcal{S}^{waic}(\xv) = \mathbb{E}_{i}(\log p_{\theta_i}(\xv)) - \text{Var}_i(\log p_{\theta_i}(\xv)).
    \label{eq:ensemble_score}
\end{align}
Though the time cost in scoring with Eq. \ref{eq:ensemble_score} could be ignored, its computational efficiency is much lower than our methods since it needs to train $N$ independent models and run inference for all the models to get the score.
To achieve better performance, the ensemble number $N$ should be large enough that leads to worse computational efficiency, \textit{e.g.}, when $N=5$, we need 5 times the computation resource of the likelihood method to get such 5 models and run the inference.


As the results shown in Table \ref{tab:computation_infer_time}, our method Resultant with SVD implementing the DEC method is only approximately 1.5x computation time than the likelihood, which is still fast. 
For the training efficiency shown in Table \ref{tab:computation_training_time}, the PHP method still does not need much computation resources.
while the computational efficiency of the WAIC largely depends on the computation resource.

\begin{table}[h!]
\centering
\caption{{\rebuttal Computational efficiency under the metric of average wall clock inference time for a data example on an NVIDIA A100 GPU. The inference time for ensemble methods means it could cost the least time if all the models are doing inference at the same time with sufficient computational resources like multiple GPUs but will cost $N$ times longer than it if only one model of the ensemble could be inferred at one time due to limited computational resources.}}
\begin{tabular}{lcc}
\hline
\multicolumn{3}{c}{\textbf{Avg. inference time (ms) for a data example }}              \\ \hline
\multicolumn{1}{l|}{Method}     & Gray-scale data & RGB nature data \\ \hline
\multicolumn{1}{l|}{Likelihood}       & 9.33            & 9.58            \\
\multicolumn{1}{l|}{Ensemble of $N$ VAEs}  & (9.33, $N\times9.33$]            & (9.58, $N\times9.58$] \\
\multicolumn{1}{l|}{PHP}        & 10.6            & 11.0            \\
\multicolumn{1}{l|}{DEC(SVD)}   & 13.3            & 14.4            \\
\multicolumn{1}{l|}{DEC(SVD-bs)}   &11.0             &11.6             \\
\multicolumn{1}{l|}{DEC(JPEG)}  & 10.0            & 11.4            \\
\multicolumn{1}{l|}{DEC(PNG)}   & 10.1            & 10.7            \\
\multicolumn{1}{l|}{Resultant(SVD)} & 14.3            & 14.6            \\ \hline
\end{tabular}
\label{tab:computation_infer_time}
\end{table}

\begin{table}[h!]
\centering
\caption{{\rebuttal Computational efficiency under the metric of average wall clock training time for different methods on an NVIDIA A100 GPU. The training time for ensemble methods means it could cost the least time if all the models are trained at the same time with sufficient computational resources like multiple GPUs but will cost $N$ times longer than it if only one model of the ensemble could be trained at one time due to limited computational resources.}}
\begin{tabular}{ccc}
\hline
\multicolumn{3}{c}{\textbf{Avg. training time of a model}}                      \\ \hline
\multicolumn{1}{l|}{}                     & Gray-scale data & RGB nature data \\ \hline
\multicolumn{1}{c|}{Likelihood}                 & $\approx 5h$                &  $\approx 7h$                 \\
\multicolumn{1}{c|}{PHP}                  & $\approx 0.5h$                 & $\approx 0.5h$                 \\
\multicolumn{1}{c|}{Ensemble of $N$ VAEs} &  $\approx 5h \sim N\times 5h$                  &  $\approx 7h \sim N\times 7h$                \\ \hline
\end{tabular}
\label{tab:computation_training_time}
\end{table}

\rebuttal{
\subsection{Evaluation of DEC Method on Noisy OOD Data} \label{sec_re_noisy}
It would be interesting to add different scales of random noise to the original in-distribution data and see what would happen in detecting the noisy data as OOD data.
Intuitively, adding random noise to the data could largely increase the entropy of the distribution of in-distribution data, which could lead to better U-OOD detection performance, indicating the analysis for direction II.

We demonstrate the changing process when we gradually add noise to the data with scale $\alpha$ from 0 to 1, expressed as
\begin{align}
    \xv_{noisy} \leftarrow \xv + \alpha \times \epsilon, \epsilon \sim \mathcal{N}(0,\mathbf{I}).
\end{align}

As Table \ref{tab_noisy_fm} and Table \ref{tab_noisy_cf} show, along with the scale of noise increases, the U-OOD detection performance of all methods becomes better, supporting our analysis of the direction II for incremental effectiveness on likelihood.
To be more specific, $\mathcal{H}_{p_{noisy}}(\xv_{noisy}) > \mathcal{H}_{p_{ID}}(\xv_{id})$ contributes to $\text{Ent-Mut}(\theta, \phi, p_{noisy}) > \text{Ent-Mut}(\theta, \phi, p_{id}$), and further leads to the gap  $\mathcal{G}>0$ in Eq. \ref{eq:root}.

}
\begin{table}[h!]
\centering
\caption{{\rebuttal U-OOD detection performance under the metric AUROC $\uparrow$ on the FashionMNIST(ID)/Noisy-FashionMNIST(OOD) dataset pair in different noisy scales. ``Comp.'' denotes the ratio of average bits between ID and OOD datasets, \textit{i.e.}, $bits(\text{Fashion})/bits(\text{Noisy-Fashion})$, where the bits are measured by JPEG compressor.}}
\setlength\tabcolsep{10pt}
\renewcommand{\arraystretch}{1.2}
\scalebox{0.9}{
\begin{tabular}{cccccc}
\hline
\multicolumn{6}{c}{FashionMNIST(ID) / Noisy-FashionMNIST(OOD)}             \\ \hline
\multicolumn{1}{c|}{noise scale} & Likelihood & PHP & DEC & Resultant & Comp. \\ \hline
\multicolumn{1}{c|}{0.0}         &50.0 & 50.0    & 50.0    &50.0   & 1.00   \\
\multicolumn{1}{c|}{0.1}         &92.0      &93.8     &99.4     &99.7   &1.26    \\
\multicolumn{1}{c|}{0.2}         &97.7      &98.2    &99.5     &100.   &1.38   \\
\multicolumn{1}{c|}{0.3}         &99.4     & 99.5    & 99.5    & 100.  &1.46    \\
\multicolumn{1}{c|}{0.4}         & 99.9     &99.9     &99.9     &100.   &1.53    \\
\multicolumn{1}{c|}{0.5}         &99.9      &99.9     &99.9    &100.   &1.58    \\
\multicolumn{1}{c|}{0.6}         &100.      &100.    &100.     &100.   &1.63    \\
\multicolumn{1}{c|}{0.7}         &100.      &100.    &100.     &100.   &1.65    \\
\multicolumn{1}{c|}{0.8}         &100.      &100.    &100.     &100.  & 1.67   \\
\multicolumn{1}{c|}{0.9}         &100.      &100.    &100.     &100.   & 1.68   \\
\multicolumn{1}{c|}{1.0}          &100.      &100.    &100.     &100.   & 1.69   \\ \hline
\end{tabular}}
\label{tab_noisy_fm}
\end{table}

\begin{table}[h!]
\centering
\caption{{\rebuttal U-OOD detection performance under the metric AUROC $\uparrow$ on the CIFAR(ID)/Noisy-CIFAR-10(OOD) dataset pair in different noisy scales. ``Comp.'' denotes the ratio of average bits between ID and OOD datasets, \textit{i.e.}, $bits(\text{CIFAR10})/bits(\text{Noisy-CIFAR10})$, where the bits are measured by JPEG compressor.}}
\setlength\tabcolsep{10pt}
\renewcommand{\arraystretch}{1.2}
\scalebox{0.9}{
\begin{tabular}{cccccc}
\hline
\multicolumn{6}{c}{CIFAR10(ID) / Noisy-CIFAR10(OOD)}             \\ \hline
\multicolumn{1}{c|}{noise scale} & Likelihood & PHP & DEC & Resultant & Comp.\\ \hline
\multicolumn{1}{c|}{0.0}         &50.0      & 50.0    & 50.0    &  50.0   & 1.00 \\
\multicolumn{1}{c|}{0.1}         &94.3     &94.7     &98.1     &98.5     & 1.26 \\
\multicolumn{1}{c|}{0.2}         &99.2     &99.4     &99.9     & 99.9   &1.39   \\
\multicolumn{1}{c|}{0.3}         &99.9     &99.9     &99.9     &99.9    &1.48  \\
\multicolumn{1}{c|}{0.4}         &100.     &100.     &100.     &100.    &1.53   \\
\multicolumn{1}{c|}{0.5}         &100.     &100.     &100.     &100.   & 1.57   \\
\multicolumn{1}{c|}{0.6}         &100.     &100.     &100.     &100.  &1.59    \\
\multicolumn{1}{c|}{0.7}         &100.     &100.     &100.     &100.  &1.61  \\
\multicolumn{1}{c|}{0.8}         &100.     &100.     &100.     &100.  &1.63  \\
\multicolumn{1}{c|}{0.9}         &100.     &100.     &100.     &100.  &1.64 \\
\multicolumn{1}{c|}{1.0}         &100.     &100.     &100.     &100.  & 1.65    \\ \hline
\end{tabular}}
\label{tab_noisy_cf}
\end{table}

\subsection{More Datasets for Evaluating the DEC method}
We
{evaluate the effectiveness of}
the DEC method on additional datasets as shown in Table \ref{tab:comparison_more_datasets_ent} and Table \ref{tab_cifar100_dec} with models trained on FashionMNIST, CIFAR-10, and CIFAR-100, respectively.
These experimental results strongly supports the incremental effectiveness on likelihood.

\begin{table}[h!]
\centering
\caption{ The comparisons of the U-OOD detection performance of our method on more datasets.
The new score function only has the dataset entropy-mutual calibration part. \textbf{Bold} numbers are superior results.}
\setlength\tabcolsep{9pt}
\renewcommand{\arraystretch}{1.3}
\scalebox{0.7}{
\begin{tabular}{c|ccc|c|ccc}
\hline
ID   & \multicolumn{3}{c|}{FashionMNIST}                        & ID   & \multicolumn{3}{c}{CIFAR-10}                            \\ \hline
OOD  & AUROC $\uparrow$  & AUPRC $\uparrow$ & FPR80 $\downarrow$ & OOD  & AUROC $\uparrow$  & AUPRC $\uparrow$ & PFR80 $\downarrow$ \\ \hline
             & \multicolumn{3}{c|}{Likelihood / DEC \textbf{(ours)}} &              & \multicolumn{3}{c}{Likelihood / DEC \textbf{(ours)}} \\ \hline
KMNIST       &60.03 / \textbf{60.54}    &54.60 / \textbf{55.18}    &61.6 / \textbf{60.3}    & CelebA       &57.27 / \textbf{69.00}        &54.51 / \textbf{61.83}     &69.03 / \textbf{50.93}               \\
Omniglot     &99.86 / \textbf{99.91}    &99.89 / \textbf{99.94}    &0.00 / \textbf{0.00}    & CIFAR-100    &52.91 / \textbf{54.69}        &51.15 / \textbf{52.98}     &77.42 / \textbf{73.23}\\
notMNIST     &94.12 / \textbf{94.50}    &94.09 / \textbf{93.61}    &8.29 / \textbf{6.89}    & Places365    &57.24 / \textbf{68.14}        &56.96 / \textbf{65.16}     &73.13 / \textbf{64.26}           \\
CIFAR-10-G    &98.01 / \textbf{99.31}    &98.24 / \textbf{99.25}    &1.20 / \textbf{0.40}    & LFWPeople    &64.15 / \textbf{67.84}        &59.71 / \textbf{60.28}     &59.44 / \textbf{54.75}                 \\
CIFAR-100-G   &98.49 / \textbf{98.81}    &97.49 / \textbf{98.05}    &1.00 / \textbf{0.90}    & SUN          &53.14 / \textbf{60.55}        &54.48 / \textbf{60.67}     &79.52 / \textbf{68.75}         \\ 
SVHN-G       &95.61 / \textbf{97.06}    &96.20 / \textbf{97.92}    &3.00 / \textbf{0.00}    &  Texture          &37.86 / \textbf{70.36}        &40.93 / \textbf{60.02}     &82.22 / \textbf{64.16}                       \\ 
CelebA-G     &97.33 / \textbf{97.69}    &94.71 / \textbf{95.94}    &3.00 / \textbf{2.10}    & Flowers102   &67.68 / \textbf{75.59}        &64.68 / \textbf{77.84}     &57.94 / \textbf{46.48}                   \\ 
SUN-G        &99.16 / \textbf{99.58}    &99.39 / \textbf{99.67}    &0.00 / \textbf{0.00}    & GTSRB        &39.50 / \textbf{48.35}        &41.73 / \textbf{45.59}     &86.61 / \textbf{73.83}                   \\ 
Const        &94.94 / \textbf{99.31}    &97.27 / \textbf{99.25}    &1.80 / \textbf{0.40}    & Const        &0.001 / \textbf{76.20}        &30.71 / \textbf{83.27}     &100.0 / \textbf{58.04}                  \\
Random       &99.80 / \textbf{100.0}    &99.90 / \textbf{100.0}    &0.00 / \textbf{0.00}    & Random       &71.81 / \textbf{99.53}        &82.89 / \textbf{99.73}     &85.71 / \textbf{0.000}                \\ \hline
\end{tabular}
}
\label{tab:comparison_more_datasets_ent}
\end{table}

\begin{table}[h!]
\centering
\caption{{ Comparisons on more OOD datasets between our method and other U-OOD detection
methods with VAEs trained on CIFAR-100 (ID). \textbf{Bold} numbers are superior results.}}
\setlength\tabcolsep{14pt}
\renewcommand{\arraystretch}{1.5}
\scalebox{0.65}{
\begin{tabular}{c|ccc}
\hline
ID                    & \multicolumn{3}{c}{CIFAR-100}                            \\ \hline
OOD                   & AUROC $\uparrow$ & AUPRC $\uparrow$ & FPR80 $\downarrow$ \\ \hline
\multicolumn{1}{l|}{} & \multicolumn{3}{c}{{Likelihood / DEC(ours)}}            \\ \hline
CelebA                & 58.2 / \textbf{72.4} & 56.0 / \textbf{67.0} & 65.8 / \textbf{45.0}                   \\
Places365             & 56.5 / \textbf{71.7}            & 55.5 / \textbf{66.5}            & 74.6 / \textbf{45.8}                  \\
LFWPeople             & 63.9 / \textbf{69.3}            & 58.4 / \textbf{64.1}            & 61.3 / \textbf{50.7}                  \\
SUN                   & 58.3 / \textbf{69.3}            & 55.6 / \textbf{64.1}            & 61.0 / \textbf{50.7}                  \\
Texture               & 52.7 / \textbf{69.3}            & 48.4 / \textbf{66.0}            & 66.1 / \textbf{47.4}                   \\
Flowers102            & 80.5 / \textbf{87.3}            & 80.1 / \textbf{81.0}            & 23.9 / \textbf{19.1}                  \\
GTSRB                 & 58.7 / \textbf{73.9}            & 51.8 / \textbf{68.1}            & 49.4 / \textbf{41.0}                 \\
Const                 & 0.00 / \textbf{76.5}            & 0.00 / \textbf{85.0}            & 100. / \textbf{21.9}                  \\
Random                & 100. / \textbf{100.}            & 100. / \textbf{100.}            & 0.00 / \textbf{0.00}                 \\ \hline
\end{tabular}}
\label{tab_cifar100_dec}
\end{table}

\section{More Experimental Results on Resultant}
\label{sec:app_re_avoid}

We add more experimental results for comparing the Resultant in Table  \ref{tab_exp_celeba_more} and Table \ref{tab:comparison_more_datasets_fashion} with models trained on CelebA and FashonMNIST, respectively. Across all the datasets, our methods could generally outperform the likelihood's performance.
}

\begin{table}[h!]
\centering
\caption{Comparisons on more OOD datasets between our method and other U-OOD detection methods with VAEs trained on CelebA (ID). Bold numbers are superior results. }
\setlength\tabcolsep{7pt}
\renewcommand{\arraystretch}{1.3}
\scalebox{0.70}{
\begin{tabular}{lccccccccc}
\hline
\multicolumn{10}{c}{\textbf{AUROC}$\uparrow$ with models trained on \textbf{CelebA (ID)}}                                                              \\ \hline
\multicolumn{1}{l|}{OOD datasets}          & SVHN          & STL10         & Places365        & LFWPeople           & SUN           & GTSRB         & Texture           & Const         & Random       \\ \hline
\multicolumn{1}{l|}{Likelihood \cite{kingma2013auto}}                  & 27.2          & 56.9          & 50.2          & 52.2          & 27.1          & 67.9          & 54.5          & 1.24          & \textbf{100}          \\
\multicolumn{1}{l|}{HVK \cite{hvae_ood}}                   & 36.8          & 59.7          & 59.1          & \textbf{59.9} & 54.3          & 49.8          & 61.5          & 92.9          & 74.4         \\
\multicolumn{1}{l|}{$\mathcal{LLR}^{ada}$ \cite{informative-vae}} & 91.2          & 61.5          & 55.7          & 58.6          & 58.8          & 42.3          & 68.1          & 90.2          & 73.4         \\
\multicolumn{1}{l|}{\textit{\textbf{-Ours}}}    &                  &                  &                    &     &                  &                  &    &  &                \\
\multicolumn{1}{l|}{PHP}    &56.9                  &59.9                  &63.5                    &52.5     &67.2                  &72.0                  &63.2    &53.4  &100                \\
\multicolumn{1}{l|}{DEC}    &99.7                  &60.1                  &60.9                    &55.7     &66.1                  &67.8                  &68.5    &97.0  &100                \\
\multicolumn{1}{l|}{Resultant}           & \textbf{95.8} & \textbf{67.6} & \textbf{68.4} & 55.9          & \textbf{73.7} & \textbf{75.6} & \textbf{76.3} & \textbf{97.1} & \textbf{100} \\ \hline
\end{tabular}
}
\label{tab_exp_celeba_more}
\end{table}

\begin{table}[t]
\centering
\caption{The comparisons of our method ``Resultant'' and likelihood on more datasets. Bold numbers are superior performance.}
\setlength\tabcolsep{15pt}
\renewcommand{\arraystretch}{1.3}
\scalebox{0.65}{
\begin{tabular}{c|ccc}
\hline
ID          & \multicolumn{3}{c}{FashionMNIST}                                       \\ \hline
OOD         & AUROC $\uparrow$       & AUPRC $\uparrow$       & FPR80 $\downarrow$   \\ \hline
            & \multicolumn{3}{c}{Likelihood / Resultant \textbf{(ours)}}             \\ \hline
KMNIST      & 60.03 / \textbf{78.71} & 54.60 / \textbf{68.91} & 61.6 / \textbf{48.4} \\
Omniglot    & 99.86 / \textbf{100.0} & 99.89 / \textbf{100.0} & 0.00 / \textbf{0.00} \\
notMNIST    & 94.12 / \textbf{97.72} & 94.09 / \textbf{97.70} & 8.29 / \textbf{2.20} \\
CIFAR-10-G  & 98.01 / \textbf{99.01} & 98.24 / \textbf{99.04} & 1.20 / \textbf{0.40} \\
CIFAR-100-G & 98.49 / \textbf{98.59} & 97.49 / \textbf{97.87} & 1.00 / \textbf{1.00} \\
SVHN-G      & 95.61 / \textbf{96.20} & 96.20 / \textbf{97.41} & 3.00 / \textbf{0.40} \\
CelebA-G    & 97.33 / \textbf{97.87} & 94.71 / \textbf{95.82} & 3.00 / \textbf{0.40} \\
SUN-G       & 99.16 / \textbf{99.32} & 99.39 / \textbf{99.47} & 0.00 / \textbf{0.00} \\
Const       & 94.94 / \textbf{95.20} & 97.27 / \textbf{97.32} & 1.80 / \textbf{1.70} \\
Random      & 99.80 / \textbf{100.0} & 99.90 / \textbf{100.0} & 0.00 / \textbf{0.00} \\ \hline
\end{tabular}
}
\label{tab:comparison_more_datasets_fashion}
\end{table}

\section{Error Bar}
\label{app:sec:error_bar}
After a wide range of experiments, we find the our methods' error bar for this U-OOD detection task is relatively small and we present the error bar for the main results in Table \ref{tab:main_tab} and Table \ref{tab:comparison_more_datasets} as in Table \ref{tab:main_tab_error} and Table \ref{tab:comparison_more_datasets_error}.

\begin{table}[h!]
\centering
\caption{Error bar for the Comparison of U-OOD detection methods on \textbf{``Hard'' Benchmarks} and their corresponding reverse verification under the metric of AUROC $\uparrow$.}
\setlength\tabcolsep{7pt}
\renewcommand{\arraystretch}{1.2}
\resizebox{\linewidth}{!}{
\begin{tabular}{lccccc|ccccc}
\hline
                        & \multicolumn{5}{c|}{\textbf{``Hard'' Benchmarks}}                                                                                       & \multicolumn{5}{c}{\textbf{Reverse Verification}}                                                                                       \\ \hline
ID Dataset              & CIFAR-10             & FashionMNIST         & CelebA               & CelebA               & CIFAR-100             & SVHN                 & MNIST                & CIFAR-10             & CIFAT-100            & SVHN                 \\
OOD Dataset             & SVHN                 & MNIST                & CIFAR-10             & CIFAT-100            & SVHN                  & CIFAR-10             & FashionMNIST         & CelebA               & CelebA               & CIFAR-10O            \\ \hline

 Likelihood \cite{kingma2013auto}              & $\pm$1.4                 &  $\pm$0.8               &   $\pm$1.3               & $\pm$1.3                 & $\pm$1.1                  &  $\pm$0.0               &  $\pm$0.0                & $\pm$1.4                  & $\pm$ 1.5                 & $\pm$ 0.0             \\
HVK \cite{hvae_ood}                    & $\pm$2.3                &  $\pm$0.7               &  $\pm$2.0                &  $\pm$2.0               & $\pm$1.6                 & $\pm$2.3                     & $\pm$1.4                    &  $\pm$2.1                     & $\pm$2.1                     & $\pm$2.2                      \\
$\mathcal{LLR}^{ada}$ \cite{informative-vae}  & $\pm$  0.4               &$\pm$ 0.5                 & $\pm$ 0.6                 & $\pm$ 0.5                 & $\pm$  0.5                & $\pm$2.7                    & $\pm$1.0                      &  $\pm$1.2                     & $\pm$1.2                  & $\pm$2.5                     \\
Likelihood Ratio \cite{likelihood-ratio-jie}       &  $\pm$2.5                   &   $\pm$0.1                    & $\pm$2.7                      &   $\pm$1.8                  &  $\pm$2.0                      &  $\pm$0.1                     &  $\pm$0.0                  & $\pm$1.3                     & $\pm$1.4                   & $\pm$0.1                    \\
Likelihood Regret \cite{xiao}      & $\pm$0.3                    &  $\pm$0.1                    &  $\pm$1.0                     & $\pm$1.3                      &  $\pm$2.0                      & $\pm$0.2                      & $\pm$0.0                     &  $\pm$1.0                & $\pm$1.1                  &  $\pm$0.1                   \\
Complexity (PNG) \cite{input_complexity}       & $\pm$0.3                     & $\pm$0.1                   & $\pm$0.2                 & $\pm$0.2                  & $\pm$0.2                      & $\pm$0.1                  &$\pm$0.0                      &  $\pm$0.4                   &$\pm$0.4                    &$\pm$0.0                     \\
WAIC (5VAE) \cite{choi2018waic}              & $\pm$0.3                      &$\pm$0.9                     &$\pm$1.9                     &  $\pm$1.6                & $\pm$2.2               &  $\pm$0.2                   & $\pm$0.0             &  $\pm$2.2                    &  $\pm$2.0                   &  $\pm$ 0.4                    \\
\textit{\textbf{-ours}} &                    &                     &                   &                  &                       &                   &                      &                      &                     &                     \\
PHP                     &  $\pm$1.3                &  $\pm$0.5               & $\pm$1.2                 & $\pm$1.3                  & $\pm$1.1                   & $\pm$0.0                  &  $\pm$0.0               &  $\pm$0.8               &  $\pm$1.0                  &   $\pm$0.0             \\
DEC                     &  $\pm$0.3               &  $\pm$0.4           &  $\pm$0.6               & $\pm$0.7                  & $\pm$0.5                  & $\pm$0.0                 & $\pm$0.0                 &  $\pm$0.8                & $\pm$ 0.7              & $\pm$ 0.0                \\
 Resultant                   & $\pm$1.4         &  $\pm$0.5        &  $\pm$1.3        & $\pm$1.4        & $\pm$1.2        &  $\pm$0.0      &  $\pm$0.0        &  $\pm$0.8        & $\pm$ 1.0        & $\pm$ 0.0        \\ \hline
\end{tabular}
}
\label{tab:main_tab_error}
\end{table}

\begin{table}[t]
\centering
\caption{Error bar for the comparisons between ``Likelihood'' and our method ``Resultant'' on more datasets {(``Likelihood'' / ``Resultant'')}.}
\setlength\tabcolsep{20pt}
\renewcommand{\arraystretch}{1.3}
\scalebox{0.95}{
\resizebox{\linewidth}{!}{
\begin{tabular}{c|ccc|ccc}
\hline
\multirow{2}{*}{OOD} & \multicolumn{3}{c|}{CIFAR-10 (ID)} & \multicolumn{3}{c}{CIFAR-100 (ID)} \\ \cline{2-7} 
 & AUROC $\uparrow$ & AUPRC $\uparrow$ & FPR80 $\downarrow$ & AUROC $\uparrow$ & AUPRC $\uparrow$ & FPR80 $\downarrow$ \\ \hline
CelebA &   $\pm$1.4 / $\pm$0.8   &   $\pm$1.3 / $\pm$0.7   &   $\pm$0.7 / $\pm$0.6   &   $\pm$1.5 / $\pm$1.0  &   $\pm$0.8 / $\pm$0.7   &   $\pm$0.6 / $\pm$0.3   \\
SUN &   $\pm$0.9 / $\pm$0.4   &   $\pm$1.1 / $\pm$0.7  &   $\pm$1.5 / $\pm$0.9  &  $\pm$1.3 / $\pm$1.1   &   $\pm$1.7 / $\pm$0.8   &   $\pm$1.2 / $\pm$0.4   \\
Places365 &   $\pm$1.2 / $\pm$0.7   &  $\pm$1.3 / $\pm$0.7   &   $\pm$1.9 / $\pm$1.6  &   $\pm$1.9 / $\pm$0.6  &   $\pm$1.9 / $\pm$0.8   &   $\pm$1.7 / $\pm$0.5   \\
LFWPeople &   $\pm$1.5 / $\pm$1.4   &  $\pm$1.6 / $\pm$1.6   &   $\pm$1.8 / $\pm$1.7   &   $\pm$1.8 / $\pm$0.4   &   $\pm$1.7 / $\pm$0.7   &   $\pm$1.9 / $\pm$0.5   \\
Texture &   $\pm$2.1 / $\pm$0.4  &   $\pm$1.9 / $\pm$1.0   &  $\pm$2.3 / $\pm$1.6   &   $\pm$1.7 / $\pm$1.0   &   $\pm$1.9 / $\pm$1.1   &  $\pm$1.5 / $\pm$0.8  \\
Flowers102 &   $\pm$1.3 / $\pm$0.8   &   $\pm$1.2 / $\pm$1.0   &   $\pm$1.8 / $\pm$1.0   &   $\pm$0.9 / $\pm$0.3   &   $\pm$0.5 / $\pm$0.4   &   $\pm$0.6 / $\pm$0.2   \\
GTSRB &   $\pm$2.1 / $\pm$1.6   &   $\pm$2.0 / $\pm$1.7   &   $\pm$2.0 / $\pm$1.6   &  $\pm$1.9 / $\pm$0.8   &   $\pm$1.8 / $\pm$0.9   &   $\pm$1.4 / $\pm$1.0   \\
Const &   $\pm$0.0 / $\pm$1.0  &   $\pm$0.1 / $\pm$0.4   &   $\pm$0.0 / $\pm$0.3   &   $\pm$0.0 / $\pm$0.7   &   $\pm$0.0 / $\pm$1.0   &   $\pm$0.0 / $\pm$0.3  \\
Random &   $\pm$1.7 / $\pm$0.1   &   $\pm$1.2 / $\pm$0.1  &   $\pm$1.0 / $\pm$0.0   &   $\pm$0.0 / $\pm$0.0   &   $\pm$ 0.0 / $\pm$0.0   & $\pm$0.0 / $\pm$0.0  \\ \hline
\end{tabular}%
}
}
\label{tab:comparison_more_datasets_error}
\end{table}

\section{Broader Impacts} \label{app:impact}
In terms of the ethical aspects and future societal consequences, our U-OOD detection method could be utilized to enhance the safety of machine learning systems such as identifying aberrant or potentially harmful online content for public safety.
These applications support the goal of fostering a safe and socially responsible use of artificial intelligence and machine learning technologies.

\end{document}